\documentclass[letterpaper, 10 pt, conference]{ieeeconf}
\IEEEoverridecommandlockouts
\usepackage{comment}
\usepackage{mathtools}
\usepackage{ifluatex,ifxetex}
\ifluatex
  \usepackage{fontspec}
\else\ifxetex
  \usepackage{fontspec}
\else
  \usepackage[T1]{fontenc}
  \usepackage[utf8]{inputenc}
\fi\fi

\usepackage{tikz}
\usepackage{placeins}

\newlength\savewidth

\newcommand{\eg}{\emph{e.g}.}

\newcommand{\vs}{\emph{vs}.\ }

\usepackage{cite}
\usepackage{times}
\usepackage{epsfig}
\usepackage{graphicx}
\usepackage{caption}
\usepackage{amsmath}
\usepackage{subcaption}
\usepackage{tcolorbox}
\usepackage{multicol}
\usepackage{changepage}
\usepackage{colortbl}
\usepackage{multirow}
\usepackage{makecell}
\usepackage{amssymb}
\usepackage{comment}
\usepackage{wrapfig}
\usepackage{algorithm} 
\usepackage{float}
\usepackage{algorithmic}
\usepackage{mathtools}
\usepackage{xcolor}
\usepackage{pythonhighlight}
\usepackage{tabularx}
\definecolor{gray}{gray}{0.95}
\definecolor{gray3}{gray}{0.90}
\usepackage{footnote}
\usepackage{footmisc}
\usepackage{url}

\makesavenoteenv{tabular}

\makesavenoteenv{table}

\usepackage{bm}

\newcommand{\astfootnote}[1]{%
\let\oldthefootnote=\thefootnote%
\setcounter{footnote}{0}%
\renewcommand{\thefootnote}{\fnsymbol{footnote}}%
\footnotetext[1]{#1}%
\let\thefootnote=\oldthefootnote%
}

\usepackage{hhline}
\usepackage{booktabs}

\newcommand{\tablestyle}[2]{\setlength{\tabcolsep}{#1}\renewcommand{\arraystretch}{#2}\centering\small}
\definecolor{green}{HTML}{009000} 
\definecolor{red}{HTML}{ea4335}  
\definecolor{orange}{HTML}{E48253}
\definecolor{purple}{HTML}{68349A}
\definecolor{teaser_gray}{HTML}{EEEEEE}
\definecolor{gray2}{HTML}{9e9e9e}
\definecolor{blue}{HTML}{0071BC}
\definecolor{red2}{HTML}{ff0000}
\definecolor{train}{HTML}{CAE6CD}
\definecolor{test}{HTML}{F4D4D1}
\usepackage{overpic}

\def\Secref#1{Section~\ref{#1}}

\newcommand{\minisection}[1]{\noindent{\textbf{#1}.}}
\newcommand{\tabref}[1]{Table~\ref{#1}}
\newcommand{\eqrref}[1]{Equation~\ref{#1}}

\newcommand{\smodel}{RoboPrompt}
\definecolor{citecolor}{HTML}{0071BC}
\definecolor{orangeback}{HTML}{E8B65B}
\definecolor{linkcolor}{HTML}{ED1C24}
\usepackage{tikz}
\usepackage[pagebackref=false, breaklinks=true, colorlinks, citecolor=blue, linkcolor=linkcolor, bookmarks=false]{hyperref}

\title{\LARGE \bf
In-Context Learning Enables Robot Action Prediction in LLMs
}
\author{Yida Yin*, Zekai Wang*, Yuvan Sharma, Dantong Niu, Trevor Darrell, Roei Herzig \\
University of California, Berkeley
}

\begin{document}

\twocolumn[{%
\renewcommand\twocolumn[1][]{#1}%
\maketitle
\centering
\includegraphics[width=\linewidth]{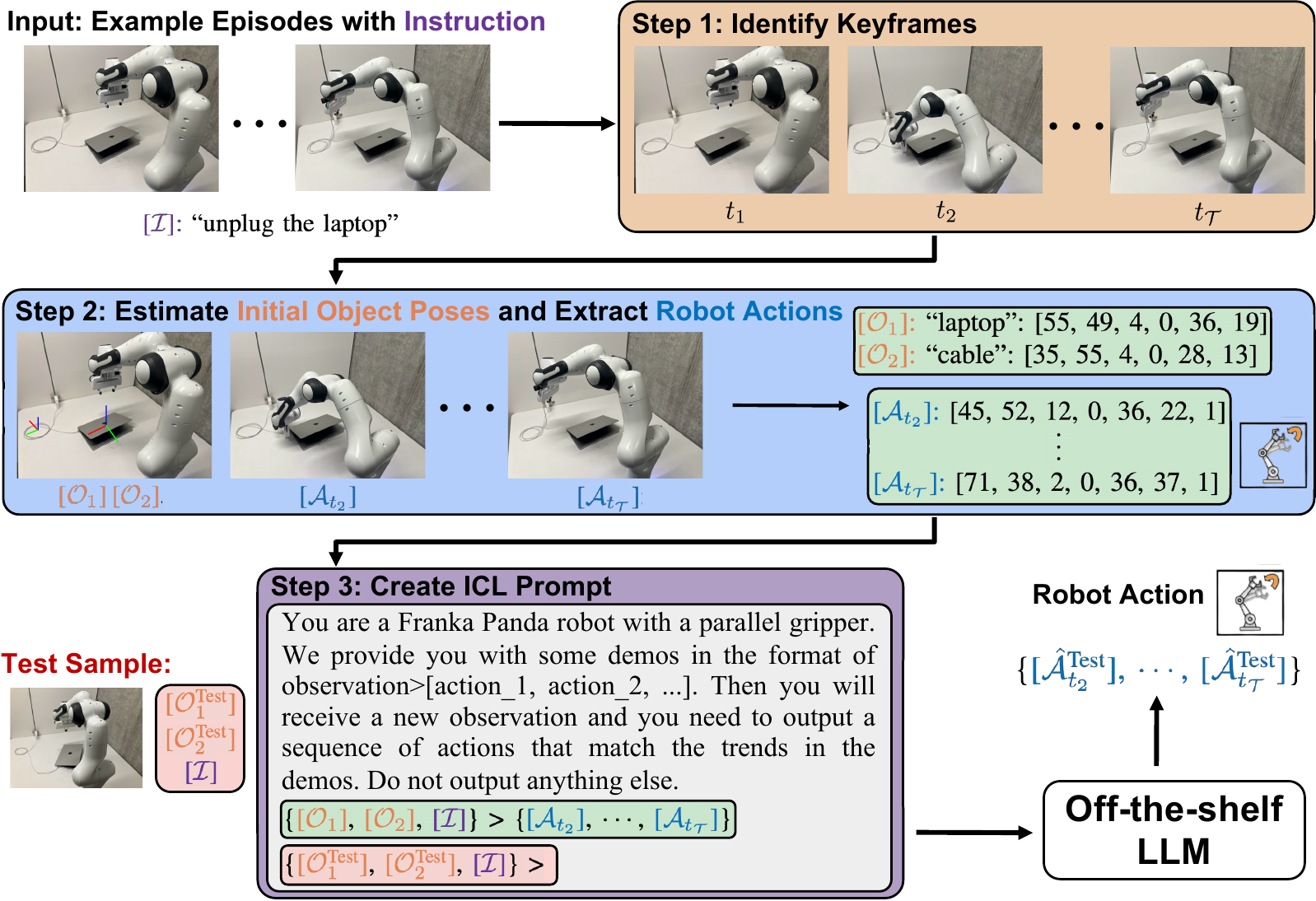}
\captionof{figure}{\textbf{Overview of {\smodel}.} We introduce a novel framework that enables an off-the-shelf text-only LLM to directly predict robot actions through in-context learning (ICL) examples without any additional training. Our method first identifies keyframes where critical robot actions occur. We next estimate \textcolor{orange}{initial object poses} and extract \textcolor{blue}{robot actions} from keyframes, and both are converted into textual descriptions. Using this textual information along with the given \textcolor{purple}{instruction}, we construct a structured prompt as ICL demonstrations, enabling the LLM to predict robot actions directly for an unseen test sample.}
\label{fig:teaser}
}]
\astfootnote{Equal contribution.}

\thispagestyle{empty}
\pagestyle{empty}

\begin{abstract}

Recently, Large Language Models (LLMs) have achieved remarkable success using in-context learning (ICL) in the language domain. However, leveraging the ICL capabilities within LLMs to directly predict robot actions remains largely unexplored. In this paper, we introduce {\smodel}, a framework that enables off-the-shelf text-only LLMs to directly predict robot actions through ICL without training. Our approach first heuristically identifies keyframes that capture important moments from an episode. Next, we extract end-effector actions from these keyframes as well as the estimated initial object poses, and both are converted into textual descriptions. Finally, we construct a structured template to form ICL demonstrations from these textual descriptions and a task instruction. This enables an LLM to directly predict robot actions at test time. Through extensive experiments and analysis, {\smodel} shows stronger performance over zero-shot and ICL baselines in simulated and real-world settings. Our project page is available at \href{https://davidyyd.github.io/roboprompt/}{https://davidyyd.github.io/roboprompt}.

\end{abstract}

\section{Introduction}

Recently, Large Language Models (LLMs), such as GPT-4~\cite{openai2024gpt4technicalreport}, Claude-3.5~\cite{claude35_model_card_addendum}, and Llama-3.1~\cite{dubey2024llama3herdmodels}, have demonstrated state-of-the-art performance on a variety of language tasks. Interestingly, LLMs exhibit a powerful emergent property --- in-context learning (ICL)~\cite{brown2020languagemodelsfewshotlearners}, where LLMs learn a new task during inference by conditioning on a few input-output demonstrations and making predictions for new, unseen inputs. Recent works~\cite{min2022rethinkingroledemonstrationsmakes, levy2023diversedemonstrationsimproveincontext, wei2023chainofthoughtpromptingelicitsreasoning} have demonstrated that ICL enhances model performance in various language tasks. A natural question arises: how to perform ICL in robotics using an off-the-shelf text-only LLM without training.

The first challenge in applying ICL in robotics is forming a compact and effective representation for ICL demonstrations. Recent research~\cite{liu2021makesgoodincontextexamples, zhao2021calibrateuseimprovingfewshot} has shown that ICL performance greatly depends on the quality of the provided demonstrations. Moreover, as shown in~\cite{li2024longcontextllmsstrugglelong, liu2023lostmiddlelanguagemodels}, simply adding a large number of inputs across a long context might lead to hallucinations and degraded performance. Therefore, it is critical to identify the essential parts within example episodes and use them to form ICL examples. Secondly, robot agents perform a specific task by mapping visual observation input, such as RGB and depth images, into robotic control output. However, the format of input and output is not generically compatible with text-only LLMs. To resolve this issue, the input and output need to be transformed into a text format that LLMs can process. Finally, LLMs are trained on a broad corpus of text by statistically predicting the next word from an input sequence. ICL further leverages this next token prediction capability for LLMs to learn a new task during inference by providing input-output demonstrations. Nevertheless, it is unclear what the input-output relationship should be when performing ICL for robotics.

In this work, we introduce \emph{\smodel}, a framework that enables pretrained LLMs to directly predict robot actions based on ICL demonstrations. As shown in Figure~\ref{fig:teaser}, our method consists of three steps. First, we identify keyframes from an example episode by finding when the joint velocities approach zero, or the gripper state transitions between open and closed. This keyframe extraction scheme captures important moments in an episode. Second, we estimate the object poses at the first timestep by leveraging an off-the-shelf pose estimate model~\cite{wen2024foundationposeunified6dpose} and extract the robot actions from all the keyframes.\footnote{Our action space consists of 6-DoF end-effector pose and gripper state.} We then convert the extracted robot actions and estimated object poses into textual descriptions. Third, we pair these textual descriptions together with a task instruction to form an ICL example using a structured template. This allows an LLM to predict robot actions directly based on new object poses from the test image and a test task instruction.

Through extensive empirical evaluations, we show that {\smodel} enables off-the-shelf LLMs to directly predict robot actions via ICL. We assess our method in 16 tasks from the RL-Bench simulation~\cite{james2019rlbench} and 6 real-world tasks on a Franka Emika Panda robot. Our results demonstrate {\smodel} outperforms several zero-shot and in-context baselines. Finally, our ablation analysis indicates {\smodel} can be applied to various LLMs, is robust to pose estimation errors, scales with the number of ICL examples, and performs competitively against supervised methods. 

\section{Related Work}

\minisection{Pretrained frontier models for robotics} Recent success in LLMs and VLMs~\cite{openai2024gpt4technicalreport, claude35_model_card_addendum, geminiteam2024gemini15unlockingmultimodal, llava} has driven various applications in robotics. To perceive the world, these models either rely on vision modality~\cite{liu2024moka,hu2023lookleapunveilingpower} to parse image inputs directly or employ separate perception modules~\cite{radford2021learning, minderer2022simpleopenvocabularyobjectdetection,liu2023grounding, kirillov2023segment} to extract scene representations and convert them into text format. Powered by the reasoning ability of LLMs, these models can break down high-level task descriptions into detailed, step-by-step plans~\cite{huang2022language, huang2023voxposer, singh2023progprompt}. These plans can be expressed in various formats, such as natural language~\cite{ahn2022can, zeng2022socratic, chen2023open, duan2024manipulate}, executable code~\cite{liang2023code, wang2023voyageropenendedembodiedagent, huang2023visuallanguagemapsrobot}, or value maps~\cite{lin2023text2motion, yu2023language, huang2024rekep}. Finally, robots sequentially execute the generated plans using predefined motion primitives or separate low-level policies~\cite{kalashnikov2021mt, jang2022bc}. While these methods demonstrate surprising zero-shot performance, they often require prompt engineering or hand-crafted design. KAT~\cite{dipalo2024kat} addresses this with ICL examples by predicting action tokens, which can be transformed back to standard 6-DoF action. LLMs have also demonstrated the ability to utilize ICL for predicting joint positions, allowing a robot to walk~\cite{wang2023promptrobotwalklarge}. In contrast, our approach performs ICL directly in the space of 6-DoF object poses and end-effector actions, enabling the robot to perform manipulation tasks. 

\minisection{Vision Language Action Models} Recent research has focused on extending VLMs to output motion control by training on pairs of images and robot actions. These models are often referred to as Vision Language Action models (VLAs). For instance, RT-2~\cite{brohan2023rt} finetunes both robotic trajectory data and large-scale vision language tasks to improve robot control. RT-2-X~\cite{openxembod} scales up the performance using the expansive Open X-Embodiment dataset. RFM-1~\cite{rfm-1} further extends the multimodal approach to facilitate interactive human-robot communication by pretraining across five modalities (text, image, video, sensor data, and robot actions). LLARVA~\cite{niu2024llarvavisionactioninstructiontuning} introduces an additional training objective of predicting intermediate 2D trajectories to align the vision and action spaces. OpenVLA~\cite{kim24openvla} incorporates an additional visual encoder DINOv2~\cite{oquab2024dinov2learningrobustvisual} to strengthen the visual grounding ability for robot learning. LLaRA~\cite{li2024llarasuperchargingrobotlearning} generates auxiliary spatial-temporal datasets from existing robot data to enhance policy performance. HPT~\cite{wang2024scalingproprioceptivevisuallearningheterogeneous} proposes to train a joint policy network across different embodiments and tasks to scale proprioceptive-visual learning. Compared with these methods, our approach does not require pretraining or finetuning on any data. By using off-the-self LLMs, we can acquire robotic skills through text-based ICL demonstrations.

\minisection{In-Context Learning} With the increasing size of model and data, LLMs exhibit striking emergent ability --- in-context learning (ICL)~\cite{brown2020languagemodelsfewshotlearners}. With a few input-output pairs as demonstrations, pretrained LLMs can generalize to new tasks \textit{without training} by identifying patterns from these examples. ICL has been successfully applied across various domains, including traditional NLP tasks~\cite{kim2022selfgeneratedincontextlearningleveraging, chen2021codex}, benchmarks that demand complex reasoning~\cite{wei2023chainofthoughtpromptingelicitsreasoning, zhou2022teachingalgorithmicreasoningincontext}, visual question answering~\cite{huang2024multimodaltaskvectorsenable}, autonomous driving~\cite{mao2023gpt}, and robotics~\cite{mirchandani2023largelanguagemodelsgeneral, zhu2024incoro}. In this paper, we focus on the potential of applying ICL for robotics rather than vision or language domains. By creating a textual prompt that contains robot actions from keyframes and initial object poses, we exploit LLM's ICL capabilities to generate robot actions directly based on new object poses.

\section{Problem Formulation}
\label{sec:form}
We address the problem of robot action prediction in the ICL setting using an off-the-shelf LLM without additional training. In ICL setting, an LLM, denoted as $f(\cdot)$, is provided with a set of $n$ input-output examples $\{(x^i, y^i)\}_{i=1}^n$. The model's task is to generate a response $\hat{y}^\text{Test}$ for an unseen test query $x^\text{Test}$ based on the provided examples:
\begin{equation}
    \hat{y}^\text{Test} = f(x^\text{Test}\ |\ (x^1, y^1),\cdots,(x^n, y^n)).
\label{eq:llm}
\end{equation}

Here, we use an LLM to perform ICL on robotic episodes. Each episode contains (i)  a task instruction $\mathcal{I}$; (ii) an RGB-D image $\mathcal{V}$ capturing the environment setup at the first timestep from a calibrated camera; (iii) a sequence of 7-DoF joint velocities $\{\mathcal{S}_t\}_{t=1}^T$; and (iv) a sequence of end-effector actions $\{\mathcal{A}_t\}_{t=1}^T$, where each $\mathcal{A}_t$ consists of a 6-DoF pose (\eg, translation and rotation in Euler angles) in the world frame $\mathcal{W}$ and a gripper state: $\mathcal{A}_{t} = [\mathcal{A}^{\text{translation}}_{t}, \mathcal{A}^{\text{rotation}}_{t}, \mathcal{A}^{\text{gripper}}_{t}]$. 

In the next section, we describe the three-step process.

\section{{\smodel} Framework}

To address the challenge of performing ICL in robotics, we propose \emph{\smodel}, a framework that enables an off-the-shelf LLM to directly predict robot actions through ICL without additional training. Figure~\ref{fig:teaser} illustrates our three-step approach. First, we identify keyframes where critical robot actions occur within each example episode (Section~\ref{sec:method:keyframe}). Then, we extract robot actions from these keyframes and estimate object poses in the environment at the first timestep (Section~\ref{sec:method:extract_actions_and_pose}). Last, we construct ICL demonstrations and feed them into the LLM to predict actions (Section~\ref{sec:method:ICL_prompt}).

\subsection{Identifying Keyframes}
\label{sec:method:keyframe}

As mentioned above, the ICL performance of LLMs greatly depends on the quality of selected demonstrations~\cite{liu2021makesgoodincontextexamples, zhao2021calibrateuseimprovingfewshot}. Hence, we would like to select the frames that contain the critical information. Inspired by~\cite{james2022qattentionenablingefficientlearning}, we identify when important actions of the robot happen based on two criteria: (i) the joint velocities $\mathcal{S}_{t}$ are near zero; or (ii) the gripper state $\mathcal{A}_{t}^{\text{gripper}}$ has changed. We refer to each selected frame as keyframe. Each keyframe is denoted as $t_k$ and it holds:
\begin{equation}
     \lVert \mathcal{S}_{t_k} \rVert_2 < \delta\ \text{or}\ \mathcal{A}_{t_k}^{\text{gripper}} \neq \mathcal{A}_{t_k+1}^{\text{gripper}},
\end{equation}
where $\delta$ is a small velocity threshold (See~\Secref{sec:exp:imple}).

We note that the near-zero joint velocities indicate a change in the robot's direction, and any shift between gripper states implies interactions with objects through the gripper. This approach significantly reduces the length of each episode from more than 200 frames down to 5-15 frames. Nevertheless, our keyframe extraction scheme ensures that important moments are captured and preserved. Next, we describe what information is extracted from these keyframes.

\subsection{Estimating Object Poses and Extracting Robot Actions}
\label{sec:method:extract_actions_and_pose}

After finding the keyframes, we extract two main elements that will be used to construct the ICL prompt.

\minisection{Estimating object poses} Most existing works leveraging LLMs for robotic planning~\cite{huang2023voxposer, liang2023code, huang2022language} obtain the center position for each object using segmentation models, such as GroundingDINO~\cite{liu2023grounding} and SAM~\cite{kirillov2023segment}. However, relying on object locations results in poor performance~\cite{dipalo2024kat} as most robotic tasks require precise and dexterous manipulation. 

To address this, we add the orientation of objects in addition to the center position. Specifically, we determine the location and orientation of each object within the environment at the first timestep $t_1$ using an external off-the-shelf pose estimate model~\cite{wen2024foundationposeunified6dpose}. We assume access to a set of $m$ object names $\{\mathcal{M}_j\}_{j=1}^m$ (\eg, ``laptop'', ``cable'') in the environment, as done in previous works~\cite{huang2023voxposer}\footnote{Object names are usually available in language instructions when LLMs are employed in robotic planning.}. This set of object names in the same task remains consistent across all example episodes and during test time.

We denote the pose estimation process as $g$, and the pose for the $j$-th object is then defined as:
\begin{equation}
    \mathcal{P}_j=g(\mathcal{V}, \mathcal{M}_j),
\label{eq:pose}
\end{equation}

\noindent where $\mathcal{V}$ is the RGB-D image at the first timestep. Finally, we transform each pose into the world frame $\mathcal{W}$.

\minisection{Robot actions} With the keyframes identified in Section~\ref{sec:method:keyframe}, we extract a sequence of robot actions: $\{\mathcal{A}_{t_k}\}_{k=2}^\mathcal{T}$. Note we ignore the first keyframe because the initial robot action is always the same. These actions occur within a continuous space, where it is very challenging for the LLMs to perform ICL. To address this, we follow approaches in the fully supervised methods~\cite{shridhar2022perceiveractormultitasktransformerrobotic, kim24openvla, octo_2023, niu2025pretrainingautoregressiveroboticmodels} to discretize the continuous 6-DoF pose space into bins and binarize the gripper state. Each translation component $\mathcal{A}^{\text{translation}}_{t}$ is discretized into 100 bins, where the total span of these bins is defined by the robot's maximum possible range of motion in that dimension. Similarly, each rotation component $\mathcal{A}^{\text{rotation}}_{t}$ (in Euler angles) is discretized into 72 bins, where each bin represents a 5-degree increment. To maintain consistency, we apply the same discretization technique to each object pose. 

To simplify the notation, we form the observation $\mathcal{O}_j$ for the $j$-th object by combining its label $\mathcal{M}_j$ with its corresponding discretized pose $\mathcal{P}_j$ as textual descriptions:
\begin{figure}[h]
\centering
\vspace{-0.25cm}
\begin{tcolorbox}[colback=gray, colframe=black, width=0.4\columnwidth, boxrule=0.5mm, left=0pt, right=0pt, top=1pt, bottom=1pt]
$\mathcal{O}_j=$ ``[$\mathcal{M}_j$] : [$\mathcal{P}_j$]''
\end{tcolorbox}
\vspace{-0.5cm}
\end{figure}

Next, we discuss how we form an ICL prompt from the textual descriptions of object poses and extracted robot actions, and then feed it into an LLM.

\subsection{Constructing the ICL Prompt}
\label{sec:method:ICL_prompt}

We construct a structured template to form ICL examples from the textual descriptions and then predict robot actions with LLMs based on these ICL examples. 

\minisection{Creating inputs and outputs for an ICL example} Our aim here is to generate the input $x_i$ and output $y_i$ and this input-output pair serves as an ICL example. The input $x_i$ is formulated using the observations for all objects $\{\mathcal{O}_j\}_{j=1}^m$ and the language instruction $\mathcal{I}$. The output $y_i$ is constructed with robot actions at keyframes $\{\mathcal{A}_{t_k}\}_{k=2}^\mathcal{T}$. The complete constructed prompt template is shown below:

\begin{table*}[t]
\tablestyle{14pt}{1.1}
\centering
\small
\begin{tabular}{lcccccccc}
& \makecell{close \\ jar} & \makecell{slide \\ block} & \makecell{sweep to \\ dustpan} &  \makecell{open \\ drawer} & \makecell{turn \\ tap} & \makecell{stack \\ blocks} &\makecell{push \\ button} & \makecell{place \\ wine} \\

\Xhline{1.0pt}
\multicolumn{9}{c}{\textcolor{gray2}{Supervised methods}} \\
\textcolor{gray2}{
RVT-2~\cite{goyal2024rvt2learningprecisemanipulation}} & \textcolor{gray2}{100} & \textcolor{gray2}{92} & \textcolor{gray2}{100} & \textcolor{gray2}{74} & \textcolor{gray2}{99} & \textcolor{gray2}{80} & \textcolor{gray2}{100} & \textcolor{gray2}{95} \\ 
\textcolor{gray2}{Act3D~\cite{gervet2023act3d3dfeaturefield}} & \textcolor{gray2}{92} & \textcolor{gray2}{93} & \textcolor{gray2}{92} & \textcolor{gray2}{93} & \textcolor{gray2}{94} & \textcolor{gray2}{12} & \textcolor{gray2}{99} & \textcolor{gray2}{80}\\
\Xhline{0.3pt}
\multicolumn{9}{c}{Zero-shot and ICL methods} \\

VoxPoser~\cite{huang2023voxposer} & 44 & 76 & 0 & 0 & 0 & 68 & 60 & 20\\
KAT~\cite{dipalo2024kat}  & 20 & 52 & 0 & 32 & 48 & 0 & 100 & 28 
\\
\rowcolor{gray}
\smodel & \textbf{100} & \textbf{80} & \textbf{100} & \textbf{72} & \textbf{100} & \textbf{84} & \textbf{100} & \textbf{52} \\
\\
& \makecell{screw \\ bulb} & \makecell{put in \\ drawer} & \makecell{meat off\\ grill} & \makecell{stack \\ cups} & \makecell{put in \\ safe} & \makecell{put in \\ cupboard} & \makecell{sort \\ shape} & \makecell{place \\ cups} \\

\Xhline{1.0pt}
\multicolumn{9}{c}{\textcolor{gray2}{Supervised methods}} \\

\textcolor{gray2}{
RVT-2~\cite{goyal2024rvt2learningprecisemanipulation}} & \textcolor{gray2}{88} & \textcolor{gray2}{92} & \textcolor{gray2}{99} & \textcolor{gray2}{80} & \textcolor{gray2}{96} & \textcolor{gray2}{66} & \textcolor{gray2}{95} & \textcolor{gray2}{38}\\ 
\textcolor{gray2}{Act3D~\cite{gervet2023act3d3dfeaturefield}} & \textcolor{gray2}{47} & \textcolor{gray2}{90} & \textcolor{gray2}{94} & \textcolor{gray2}{9} & \textcolor{gray2}{95} & \textcolor{gray2}{51} & \textcolor{gray2}{8} & \textcolor{gray2}{3}\\
\Xhline{0.3pt}
\multicolumn{9}{c}{Zero-shot and ICL methods} \\ 
VoxPoser~\cite{huang2023voxposer} &  32 & 0 & 4 &  0 & 0 & \textbf{32} & 0 & 0

 \\
KAT~\cite{dipalo2024kat} & 0 & 0 & 16 & 0 & \textbf{36} & 0 & 0 & 0
\\

\rowcolor{gray}
\smodel &  \textbf{40} & \textbf{20} & \textbf{16} & \textbf{16} & 24 & 16 & \textbf{8} & 0

\end{tabular}

\caption{\textbf{Simulation results on RLBench environment.} We evaluate each method across 16 RLBench tasks. For each task, we report the average success rate (\%) over 25 episodes. {\smodel} significantly outperforms various zero-shot and in-context learning (ICL) methods across a wide range of tasks. The supervised methods are \textcolor{gray2}{gray out}.}
\vspace{-.5cm}
\label{tab:sim}
\end{table*}

\begin{figure}[h]
\centering
\vspace{0.25cm}
\begin{tcolorbox}[colback=gray, colframe=black, width=.65\columnwidth, boxrule=0.5mm, left=0pt, right=0pt, top=1pt, bottom=1pt]
$x^i=$ ``\{{\textcolor{orange}{[$\mathcal{O}_1$]},  \textcolor{orange}{[$\mathcal{O}_2$]}, $\cdots$, \textcolor{orange}{[$\mathcal{O}_m$]}}, \textcolor{purple}{[$\mathcal{I}$]}\}''

$y^i=$ ``\{\textcolor{blue}{[$\mathcal{A}_{t_2}$]}, \textcolor{blue}{[$\mathcal{A}_{t_3}$]}, $\cdots$, \textcolor{blue}{[$\mathcal{A}_{t_\mathcal{T}}$]}\}''
\end{tcolorbox}
\vspace{-0.75cm}
\end{figure}
To construct the test input $x^\text{Test}$, we can apply this prompt template based on a test RGB-D image $\mathcal{V}^\text{Test}$ and a test instruction $\mathcal{I}$. Specifically, we first compute the pose for each object in the test image, denoted as $\mathcal{P}^{\text{Test}}_j$, using \eqrref{eq:pose}. We then use the same prompt to generate the test input $x^\text{Test}$. 

\minisection{Forming the ICL prompt} Since multiple episodes can serve as ICL demonstrations, we insert a symbol ``>'' between the input-output pair of each ICL example and separate consecutive pairs with a comma. After listing all ICL examples, we append the test input $x^\text{Test}$ at the end. The ICL prompt structure is illustrated as follows:

\begin{figure}[h]
\centering

\begin{tcolorbox}[colback=gray, colframe=black, width=.7\columnwidth, boxrule=0.5mm, left=0pt, right=0pt, top=1pt, bottom=1pt]
$x^1$ > $y^1$, $x^2$ > $y^2$, $\cdots$, $x^n$ > $y^n$, $x^\text{Test}$ > 
\end{tcolorbox}
\vspace{-0.5cm}
\end{figure}

This ICL prompt is then fed into an LLM using~\eqrref{eq:llm} to generate a response $\hat{y}^\text{Test}$ that contains a sequence of predicted robot actions $\{\hat{\mathcal{A}}_{t_k}^\text{Test}\}_{k=2}^\mathcal{T}$ in an autoregressive manner. These actions can be parsed and executed by a robot.

\section{Experiments}
\label{sec:exp}

We evaluate {\smodel} on 16 tasks from RLBench~\cite{james2019rlbench} and 6 tasks with a real Franka Emika Panda robot. We compare our method to both zero-shot (ZS) and ICL methods.

\subsection{Implementation Details}
\label{sec:exp:imple}
We employ GPT-4 Turbo as the base LLM and use \emph{10 ICL examples}. In simulation, the velocity threshold for keyframe extraction is set to $\delta=0.1$, and we use the ground-truth center position of each object as the object poses. In a real robot, the velocity threshold is set to $\delta=0.01$, and we estimate a 6-DoF pose for each object by leveraging FoundationPose~\cite{wen2024foundationposeunified6dpose} with GroundingDino~\cite{liu2023grounding}. More implementation details are in Section~\ref{supp:impl} of Supplementary. 

\begin{figure*}

\tablestyle{.8pt}{0.9}
\centering
  \begin{tabular}{@{}cccc@{\hskip .3em}ccc@{}}
  \begin{overpic}[width=.16\linewidth]{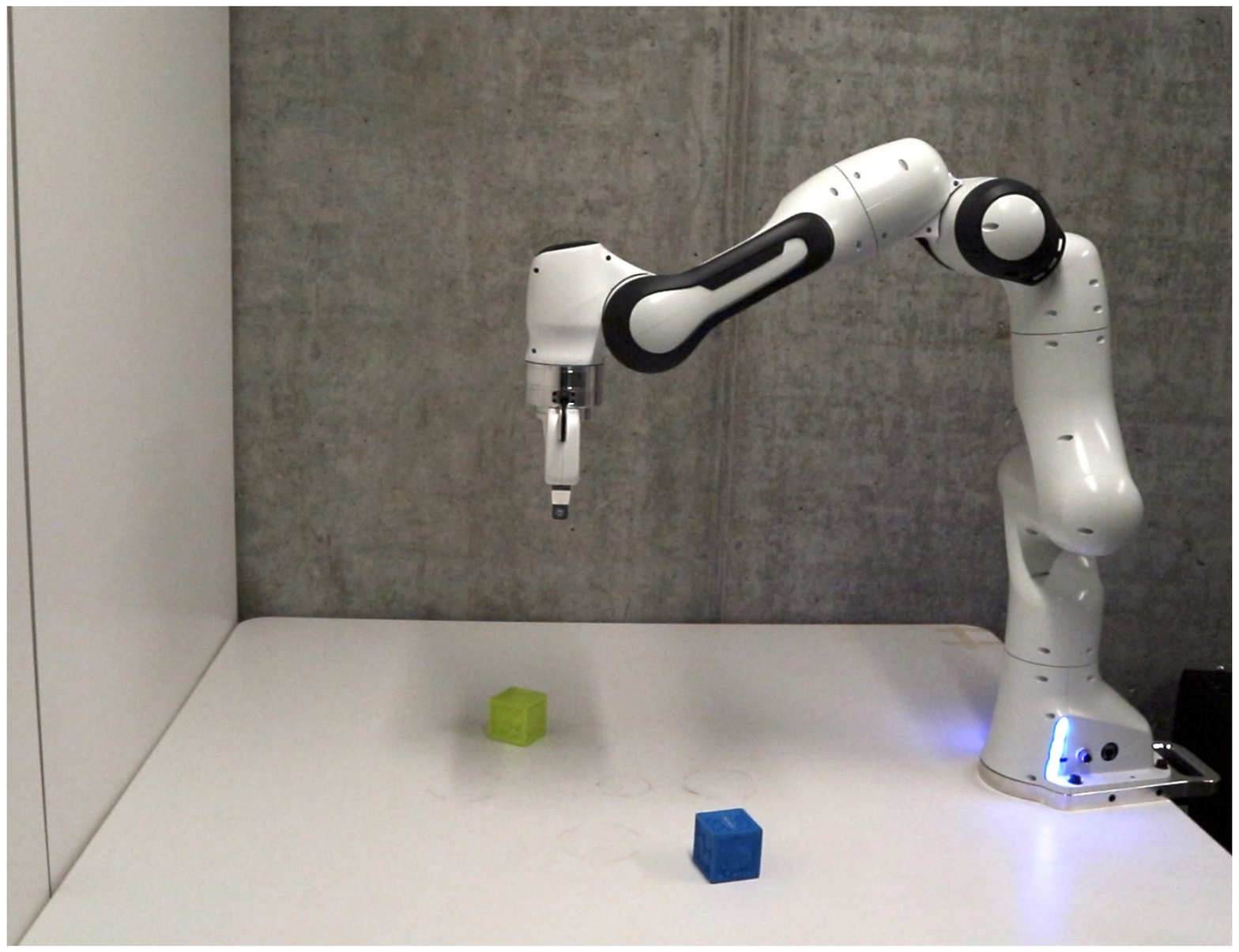}
        \put(88, 66){\textcolor{white}{$t_1$}}
    \end{overpic} &
    
    \begin{overpic}[width=.16\linewidth]
    {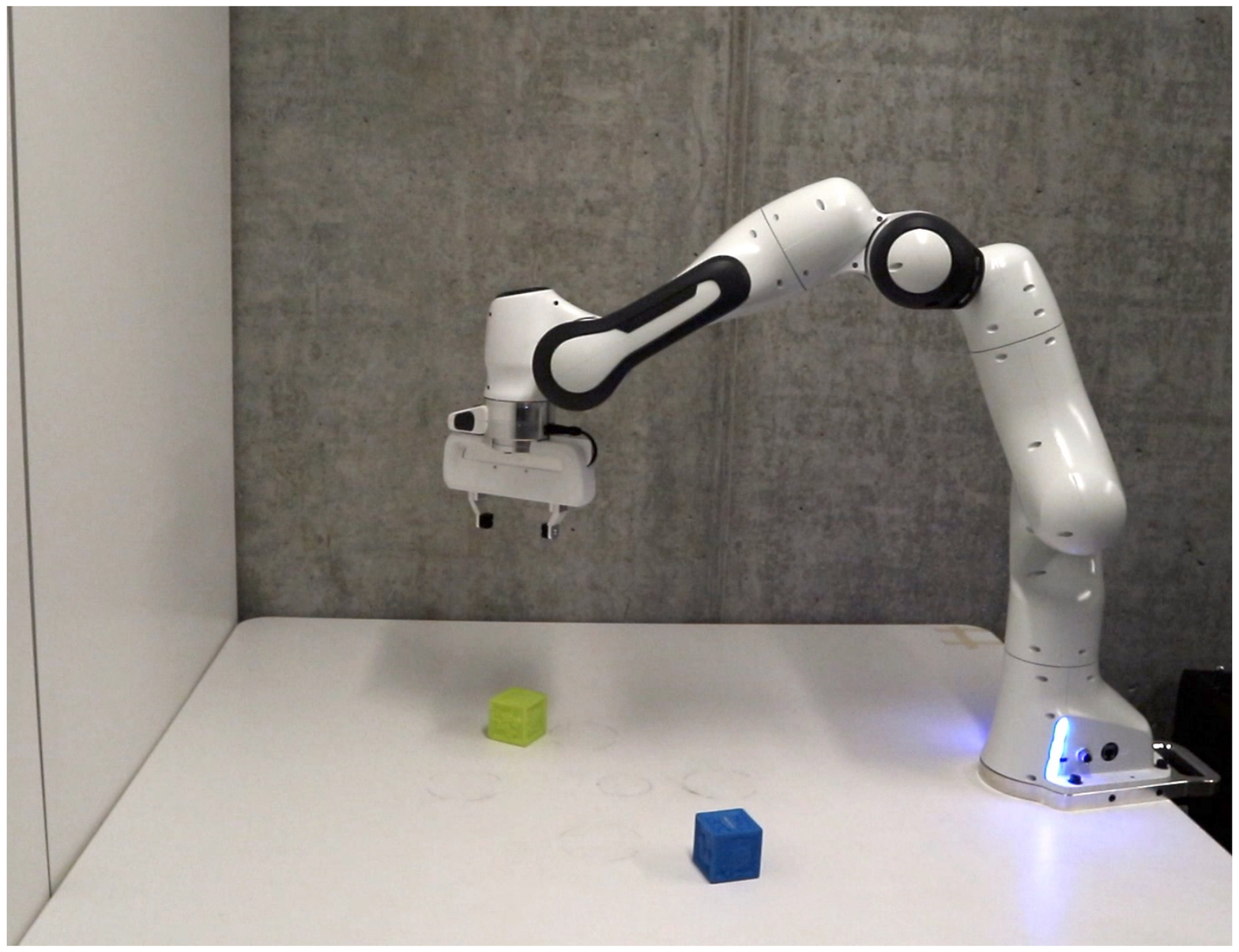}
        \put(88, 66){\textcolor{white}{$t_2$}}
    \end{overpic} & 

    \begin{overpic}[width=.16\linewidth]{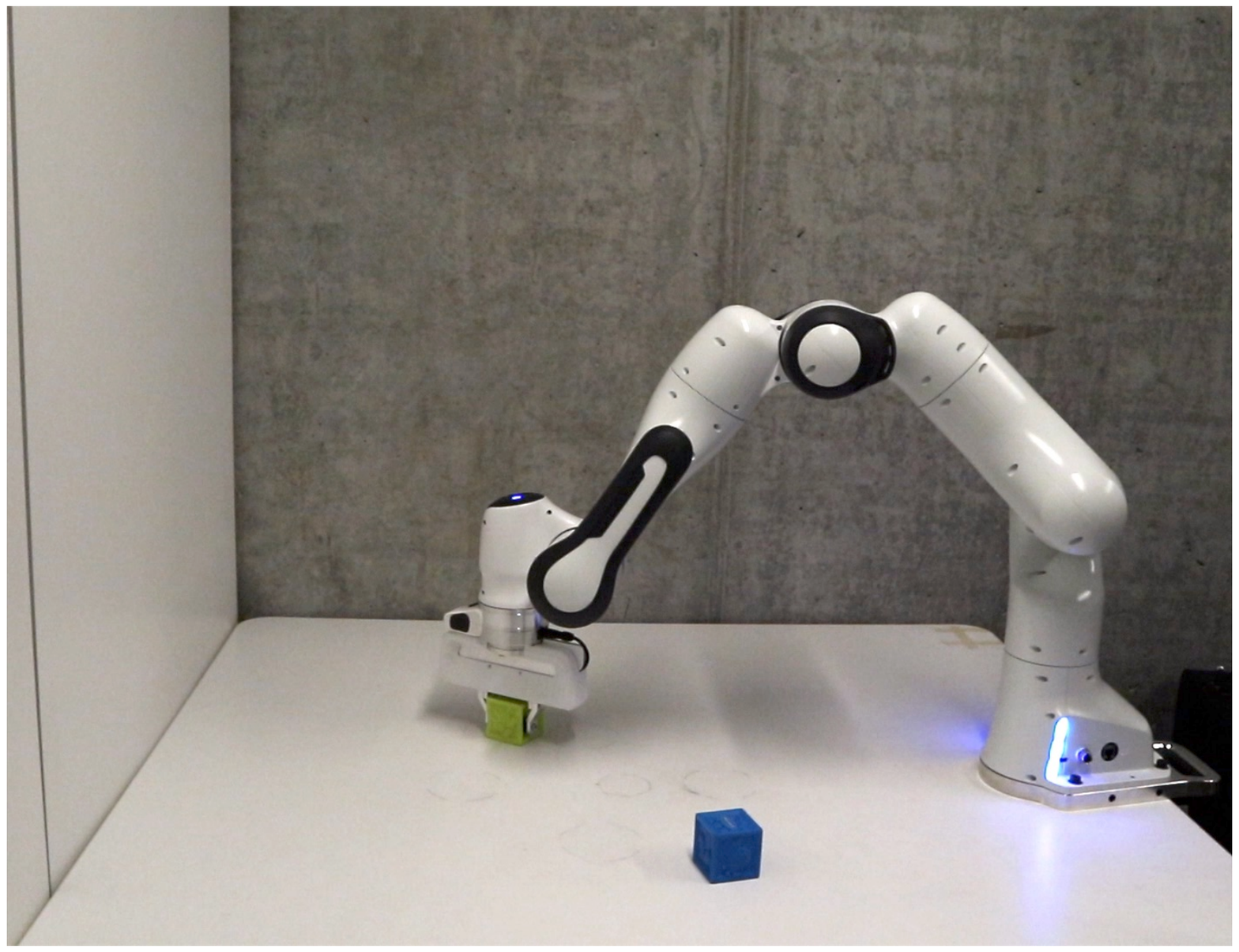}
        \put(88, 66){\textcolor{white}{$t_3$}}
    \end{overpic} & 

    &
    
     \begin{overpic}[width=.16\linewidth]{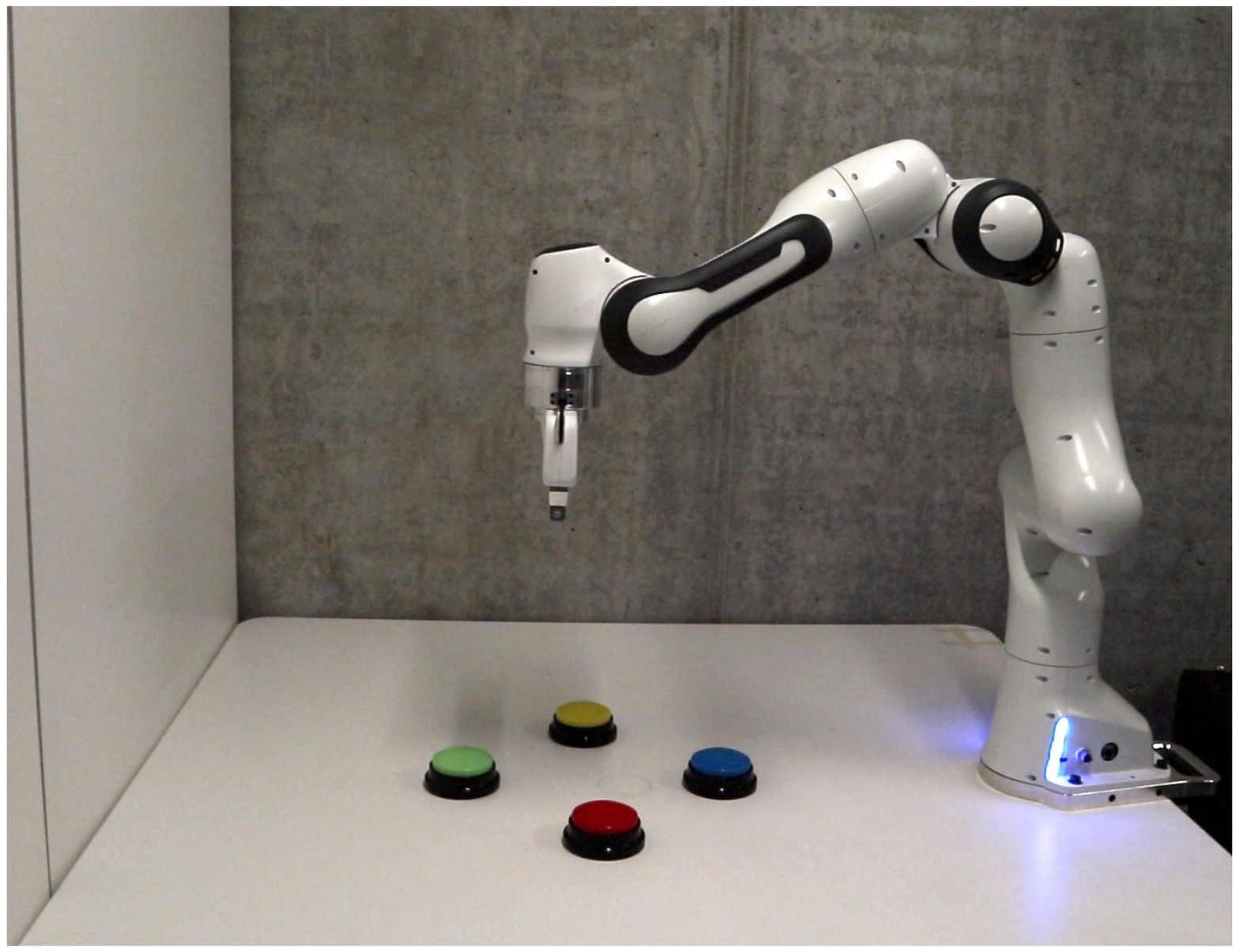}
        \put(88, 66){\textcolor{white}{$t_1$}}
    \end{overpic} &
    
    \begin{overpic}[width=.16\linewidth]
    {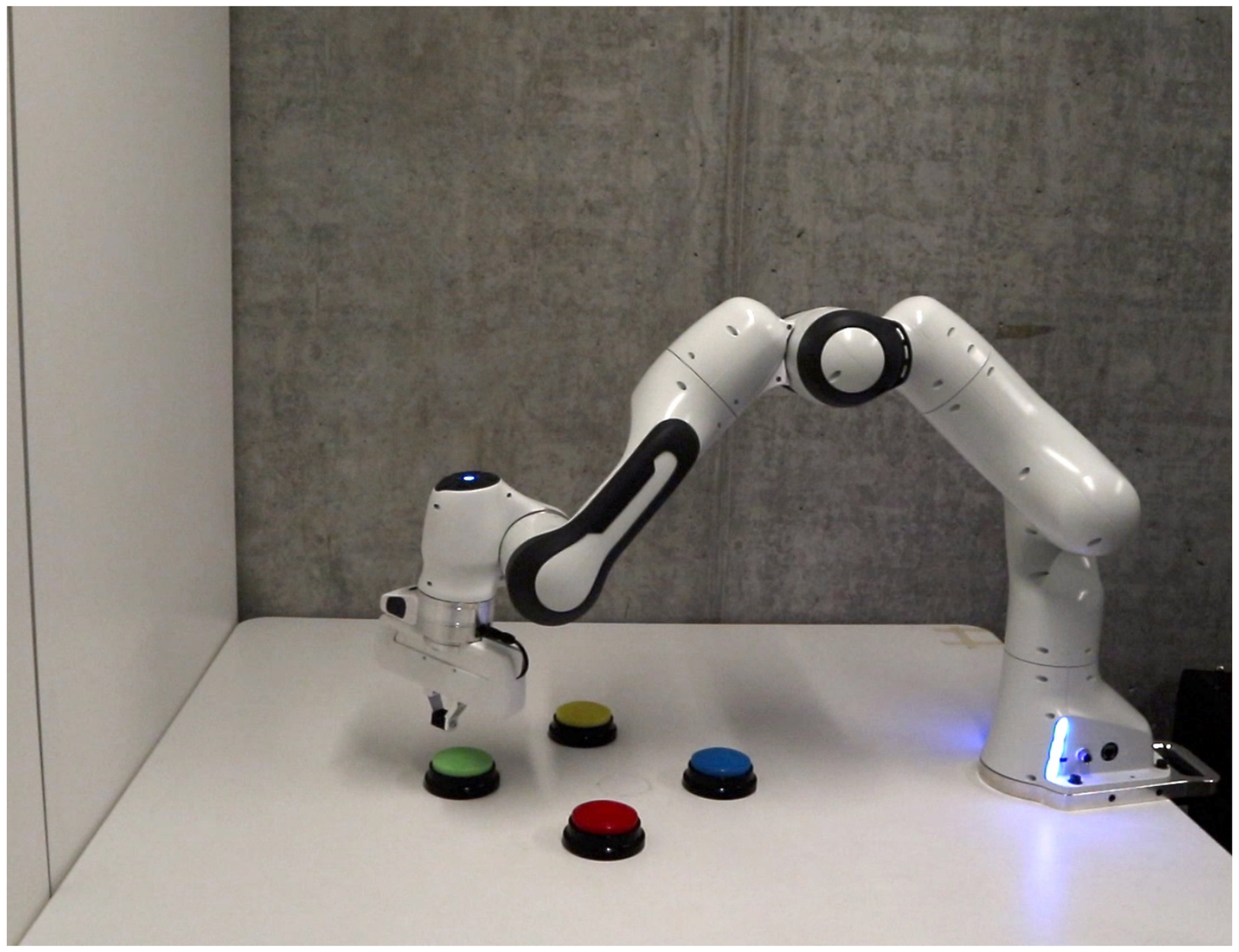}
        \put(88, 66){\textcolor{white}{$t_2$}}
    \end{overpic} & 

    \begin{overpic}[width=.16\linewidth]{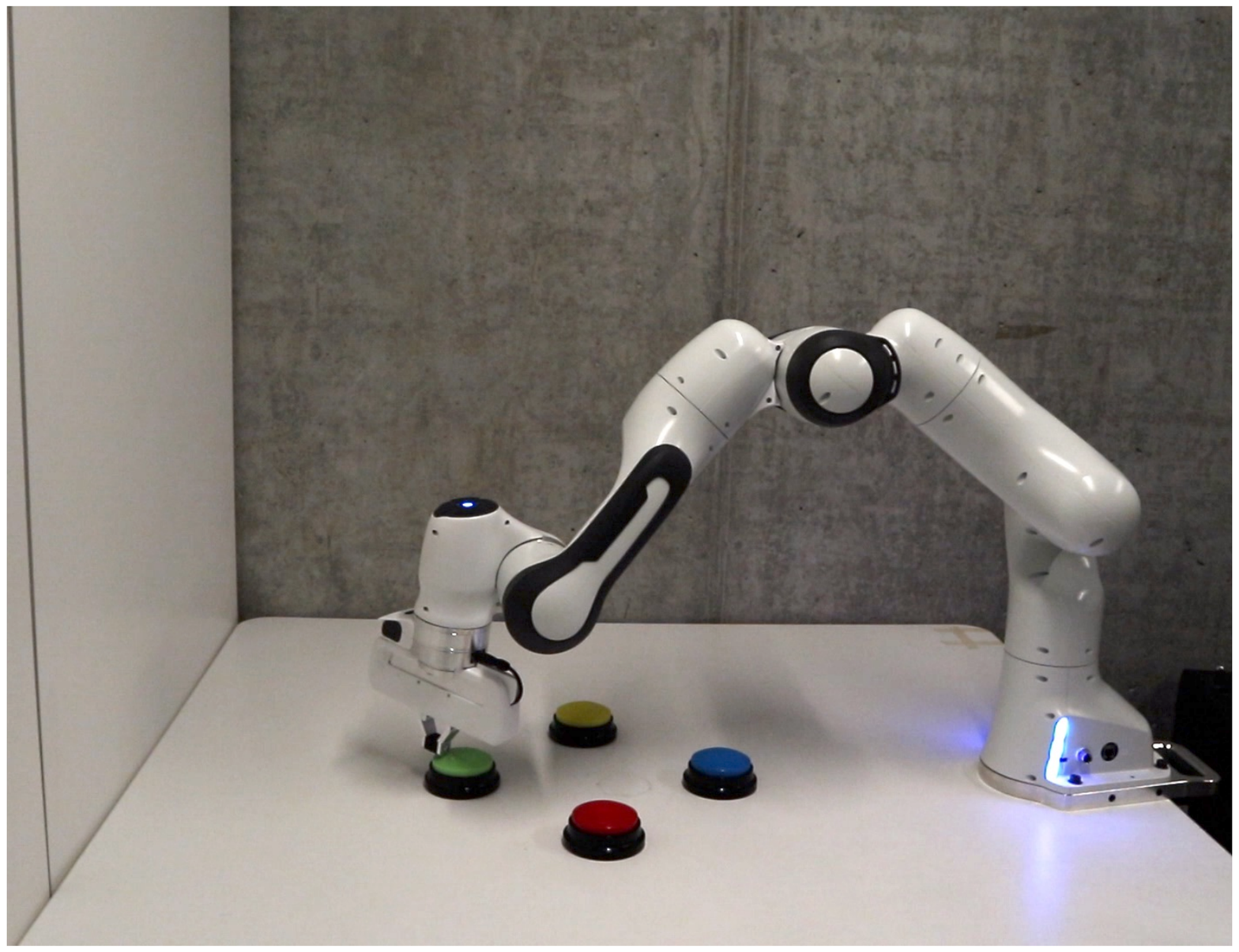}
        \put(88, 66){\textcolor{white}{$t_3$}}
    \end{overpic} \\
    
    \multicolumn{3}{c}{(a) stack cube} &
    \multicolumn{3}{c}{(b) push button} \\

     \begin{overpic}[width=.16\linewidth]{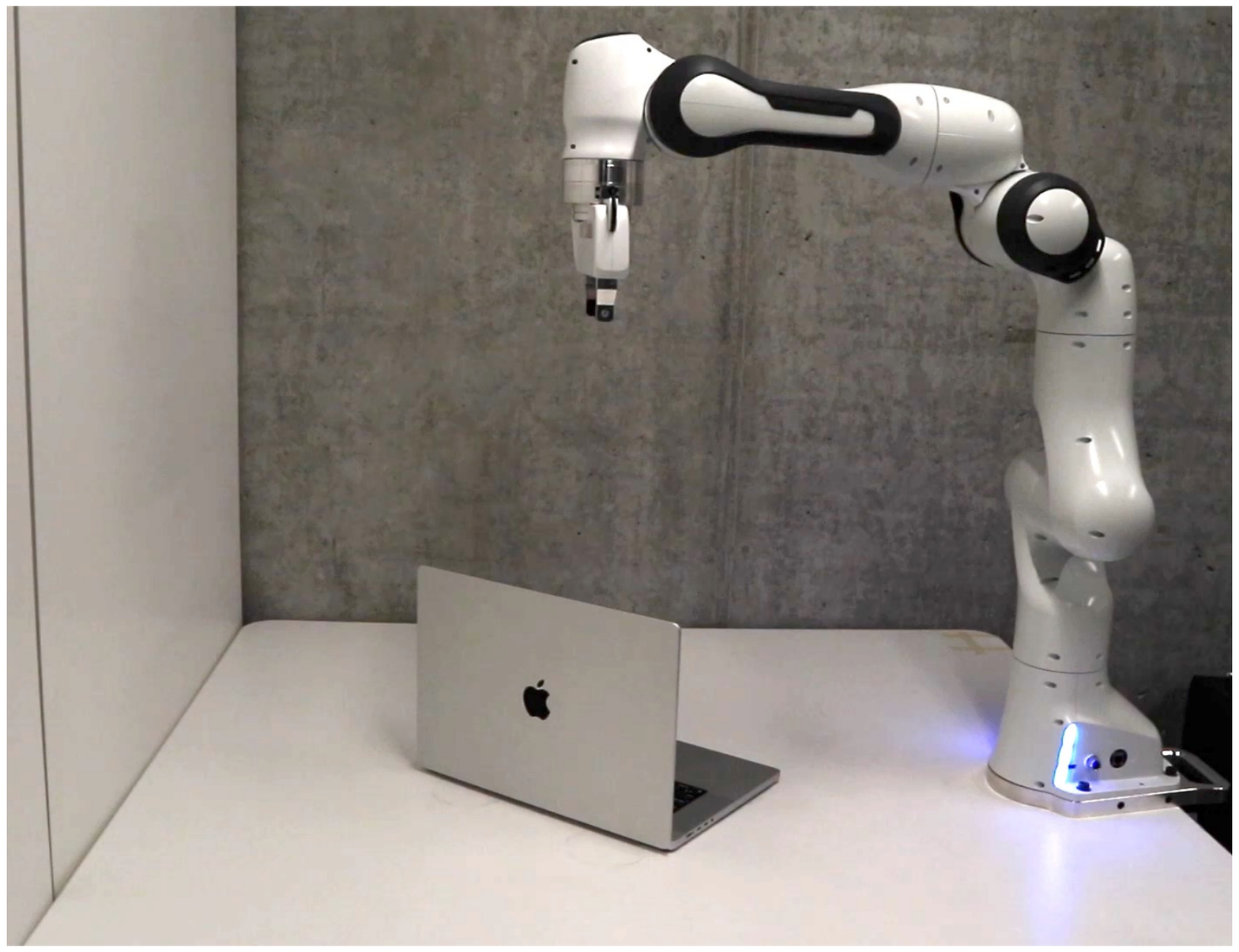}
        \put(88, 66){\textcolor{white}{$t_1$}}
    \end{overpic} & 

     \begin{overpic}[width=.16\linewidth]{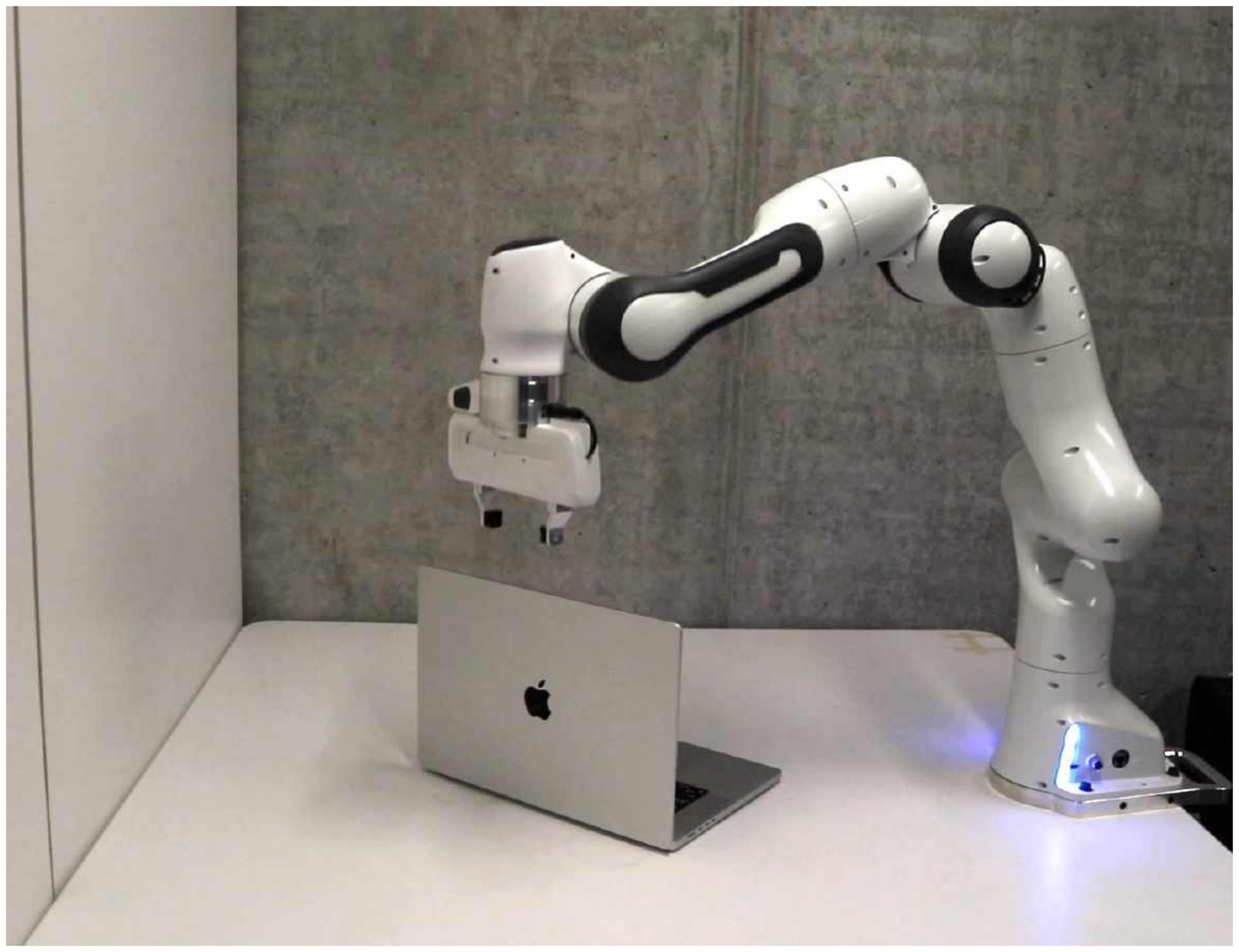} 
        \put(88, 66){\textcolor{white}{$t_2$}}
    \end{overpic} & 

    \begin{overpic}[width=.16\linewidth]{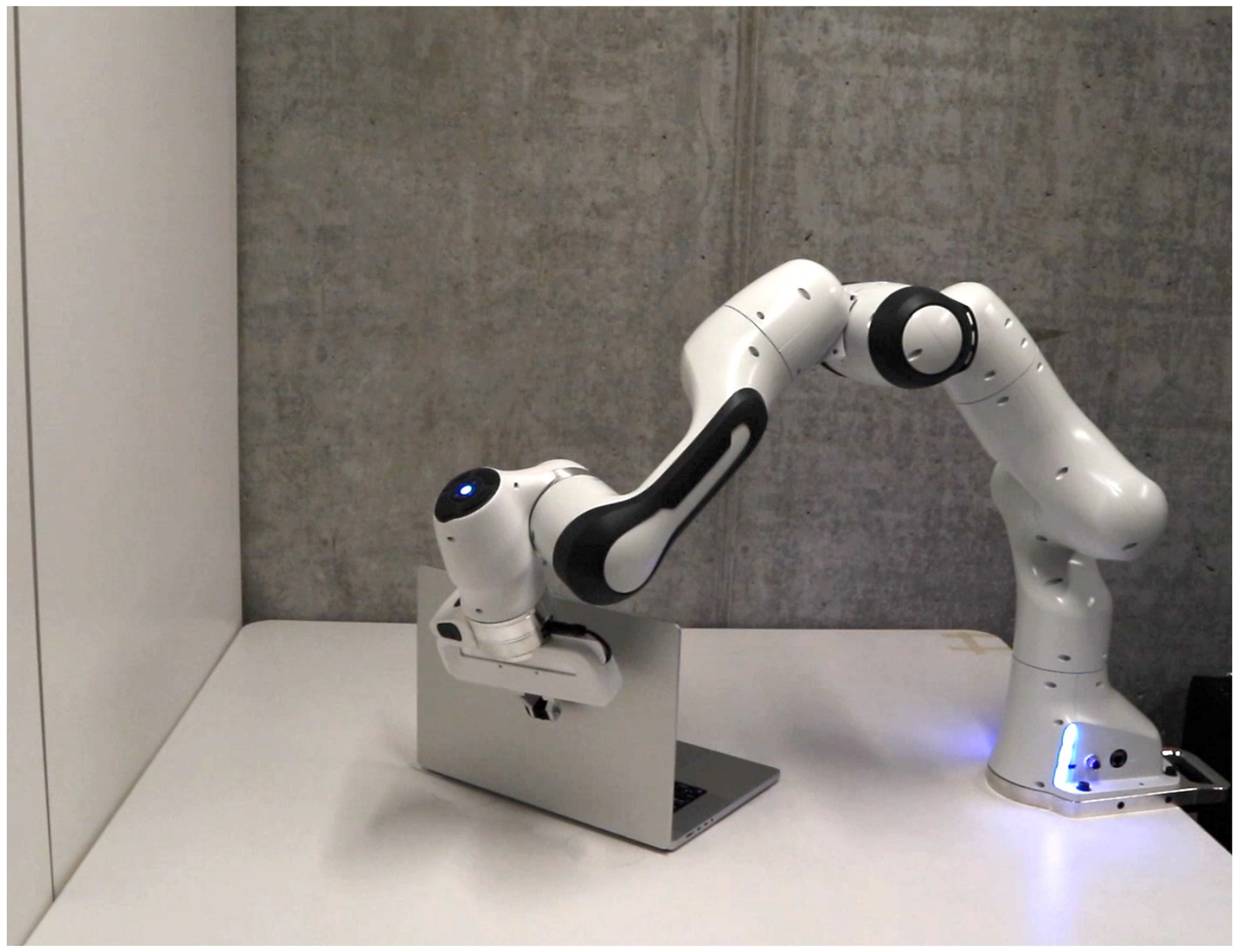}
        \put(88, 66){\textcolor{white}{$t_3$}}
    \end{overpic} & 
    
    &

     \begin{overpic}[width=.16\linewidth]{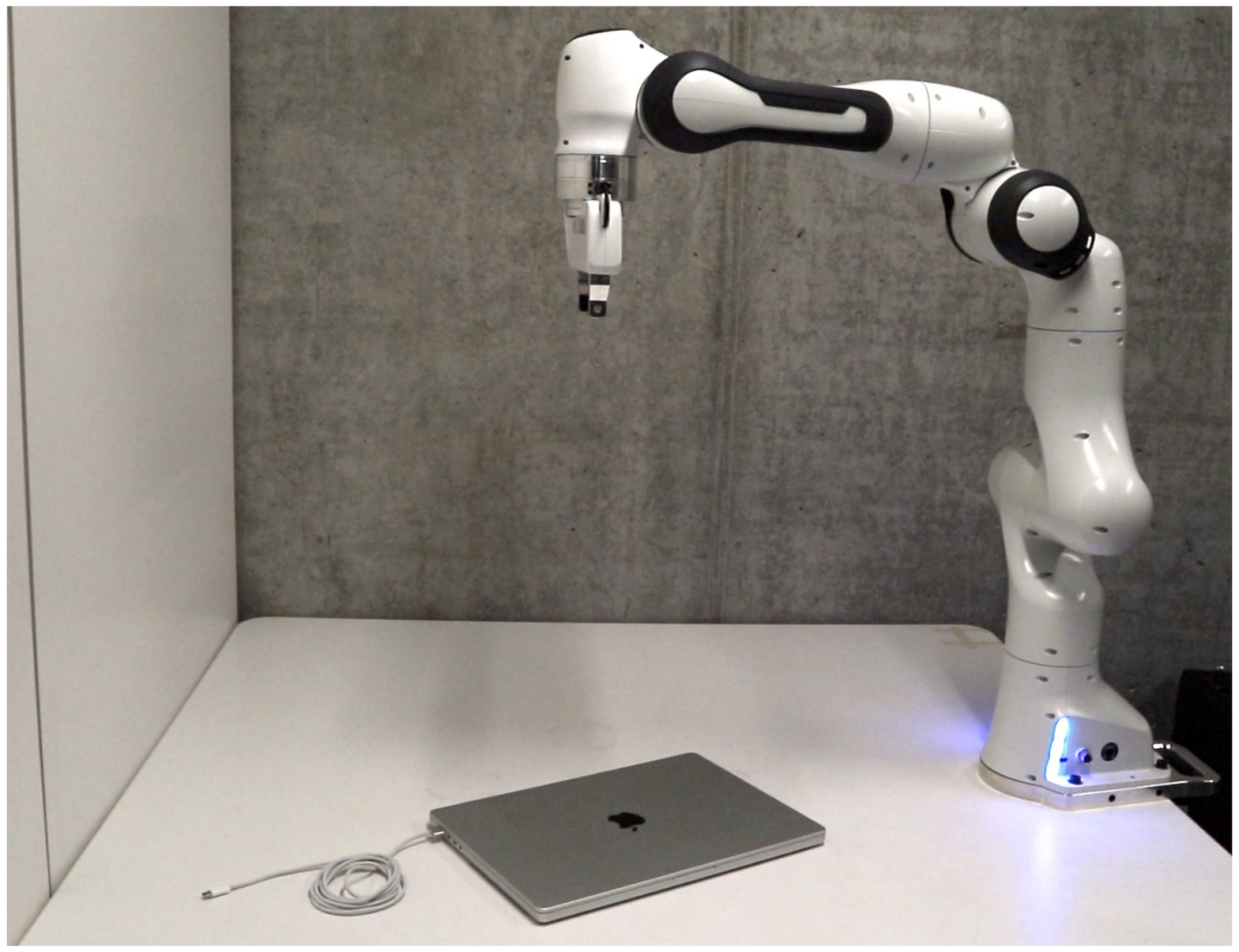}
        \put(88, 66){\textcolor{white}{$t_1$}}
    \end{overpic} &

     \begin{overpic}[width=.16\linewidth]{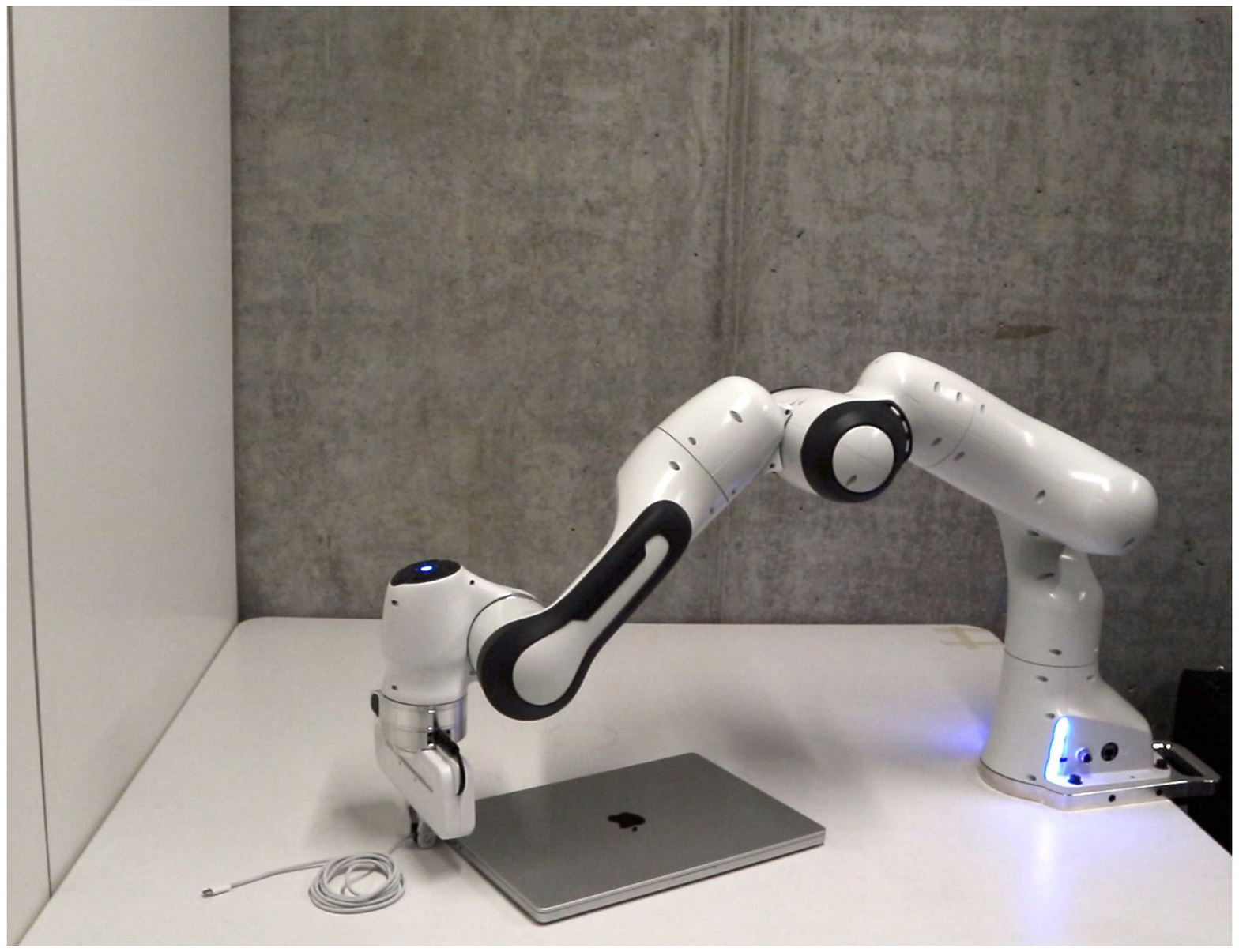}
        \put(88, 66){\textcolor{white}{$t_2$}}
    \end{overpic} &

     \begin{overpic}[width=.16\linewidth]{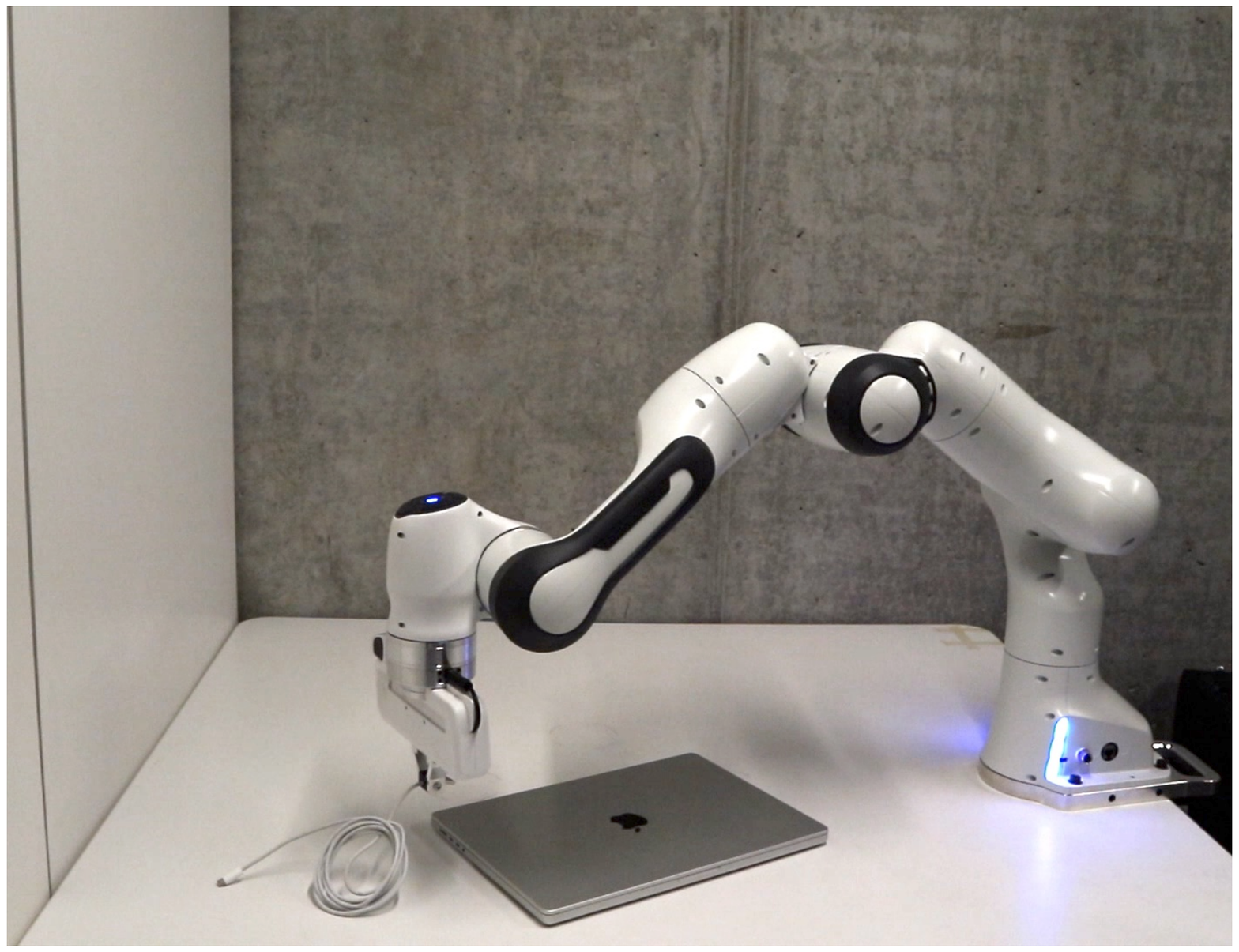}
        \put(88, 66){\textcolor{white}{$t_3$}}
    \end{overpic} \\
    \multicolumn{3}{c}{(c) close laptop} &
    \multicolumn{3}{c}{(d) unplug cable}\\
    
  \end{tabular}
  \captionof{figure}{\textbf{Visualization of the first few predicted actions from {\smodel}.} Each predicted action captures an important moment in a task. With the estimated object poses, the gripper's orientation closely aligns with that of each object.}
  \label{fig:real_task}
  \vspace{-0.1cm}
\end{figure*}

\begin{table*}[t]
\tablestyle{19pt}{1.1}
\centering
\small
\begin{tabular}{lcccccc}
 & \makecell{stack \\ cube} & \makecell{destack \\ cube} & \makecell{push \\ button} & \makecell{close \\ laptop} & \makecell{unplug \\ cable} & \makecell{push \\ multiple buttons} \\ 

\Xhline{1.0pt}
VoxPoser~\cite{huang2023voxposer} & 80 & 50 & 60 & 30 & 20 &  50 \\
KAT~\cite{dipalo2024kat} & 60 & 70 & 30 & 20 & 20 & 10 \\
\rowcolor{gray}
\smodel & \textbf{90} & \textbf{80} & \textbf{90} & \textbf{50} & \textbf{50} & \textbf{80} \\
\\
\end{tabular}
\vspace{-.1cm} 
\caption{\textbf{Real-robot results.} We evaluate each method across 6 real-world tasks. For each task, we calculate the average success rate (\%) over 10 episodes. {\smodel} achieves a better performance than several zero-shot and ICL methods.}
\label{tab:real}
\vspace{-.3cm}
\end{table*}

\subsection{Baselines} We compare {\smodel} to a few ZS and ICL baselines that leverage LLMs or VLMs in robotics. Voxposer~\cite{huang2023voxposer} builds a 3D voxel map of value functions for predicting waypoints. KAT~\cite{dipalo2024kat} transforms an object into keypoint tokens and an action into action tokens to perform ICL. 

In simulation, for a fair comparison with ZS and ICL baselines, we use the ground-truth center position for each object. In the real world, each baseline uses its own vision module. We also include end-to-end supervised methods, such as RVT-2~\cite{goyal2024rvt2learningprecisemanipulation} and Act3D~\cite{gervet2023act3d3dfeaturefield}, in RLBench for reference. 

\subsection{Simulation Results}
\label{section:exp:simu}
\minisection{Experiment Setup} We evaluate 16 tasks on a Franka robot with a parallel gripper from the RLBench simulation. The robot is allowed up to 25 steps to complete the task.

\minisection{Results} \tabref{tab:sim} shows the results of 16 RLBench tasks. {\smodel} significantly outperforms other ZS and ICL methods with an average success rate of 51.8\%, whereas Voxposer and KAT both achieve 21.0\%. {\smodel} can perform better on most tasks under a simple prompt structure in contrast to Voxposer, which requires adjusting prompts for each task. Moreover, compared to KAT, our method does not require transformation from keypoints to actions; instead, our method predicts the actions directly. Note that {\smodel} underperforms other baselines on the ``put in safe'' and ``put in cupboard'' tasks. This is likely because our method does not incorporate the detailed geometry of each object, as our approach uses a single pose to represent the object. We leave the use of shape information for each object for future work.

In addition, our method demonstrates competitive performance against fully supervised approaches, such as RVT-2 and Act3D, which achieve average success rates of 81.4\% and 65.0\%, respectively. Unlike these methods, which are trained on hundreds of example episodes using a supervised imitation learning objective, {\smodel} requires no training on any data and instead leverages the ICL capability of an off-the-shelf LLM to learn robotic tasks with inference only. Nevertheless, it should be noted that {\smodel} still has a performance gap compared with these supervised method on more challenging tasks (the second row in Table~\ref{tab:sim}) that require fine-grained or multi-stage object interactions.

\subsection{Real-Robot Results}

\minisection{Experiment setup} Following the setup~\cite{radosavovic2023robotlearningsensorimotorpretraining, niu2024llarvavisionactioninstructiontuning}, we use a 7-DoF Franka Emika Panda robot arm with a parallel jaw gripper, and a low-level Polymetis controller~\cite{Polymetis2021}. We record each example episode at 6 Hz. The RGB-D image at the first timestep is captured by the Intel RealSense D435 camera. We evaluate both ZS and ICL methods on the following 6 tasks: (i) ``stack cube'': stack blue/yellow cube on top of the other one, (ii) ``destack cube'': destack blue/yellow cube from top of the other, (iii) ``push button'': press a red/yellow/green/blue button, (iv) ``close laptop'': close the laptop screen, (v) ``unplug cable'': disconnect the laptop cable, (vi) ``push multiple buttons'': press multiple buttons in a random order. For the ``push multiple buttons'' task, we provide ICL demonstrations for \emph{only pressing a single button}. During the test time, each method is given a task instruction to \emph{press a sequence of buttons}. 

\minisection{Results} Table~\ref{tab:real} shows the results of real-world tasks. Similar to simulation results, {\smodel} can achieve an average success rate greater than 80\% on simple manipulation tasks (\eg, ``stack cube'' and ``push buttons''). For more complex tasks requiring precise object contact (e.g., ``unplug cable'', ``close laptop''), {\smodel} demonstrates reasonable performance, even with only 10 ICL examples. 

We visualize the first few actions predicted by {\smodel} for some of the tasks in Figure~\ref{fig:real_task}. The robot interacts with the object precisely and the orientation of the gripper closely aligns with that of the relevant objects. More qualitative visualizations are in Section~\ref{supp:qual} of Supplementary.

\minisection{ICL Emergent property} For the ``push multiple buttons'' task, we only provide ICL examples of pressing just one single button (e.g., ``push the red/yellow/green button''), while during the evaluation, we use a \textit{sequence of buttons} (e.g., ``push the red button, then push the yellow button, and then push the green button''). Remarkably, {\smodel} can learn during the evaluation to \emph{press multiple buttons} given a test instruction specifying the order of button pressing with an 80\% success rate. This behavior shows that {\smodel} can learn to perform a new robotic skill by composing a series of single tasks that are available in the ICL demonstrations.

\begin{figure*}
\centering
\vspace{-1.25em}
\begin{subfigure}[t]{0.32\textwidth}
    \centering
    \includegraphics[width=\textwidth]{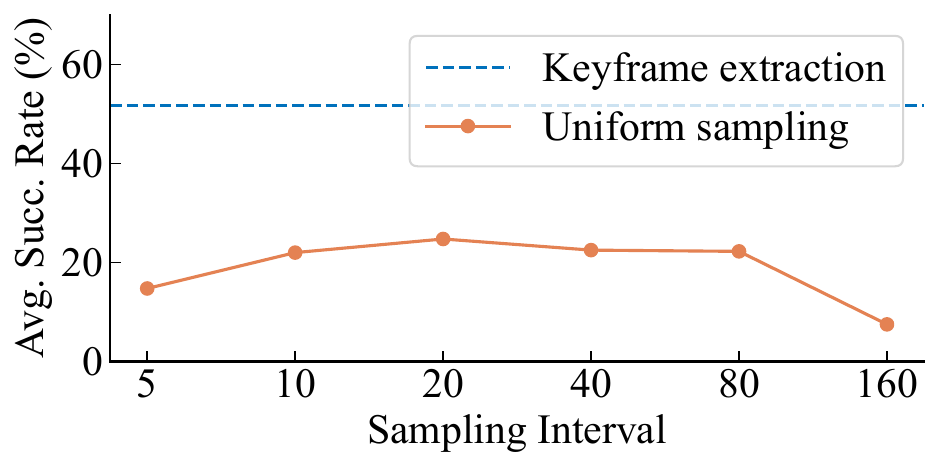}
    \vspace{-.6cm}
    \caption{{Keyframe extract. vs. Uniform sampling}}
    \label{fig:uniform}
\end{subfigure}
\hfill
\begin{subfigure}[t]{0.32\textwidth}
    \centering
    \includegraphics[width=\textwidth]{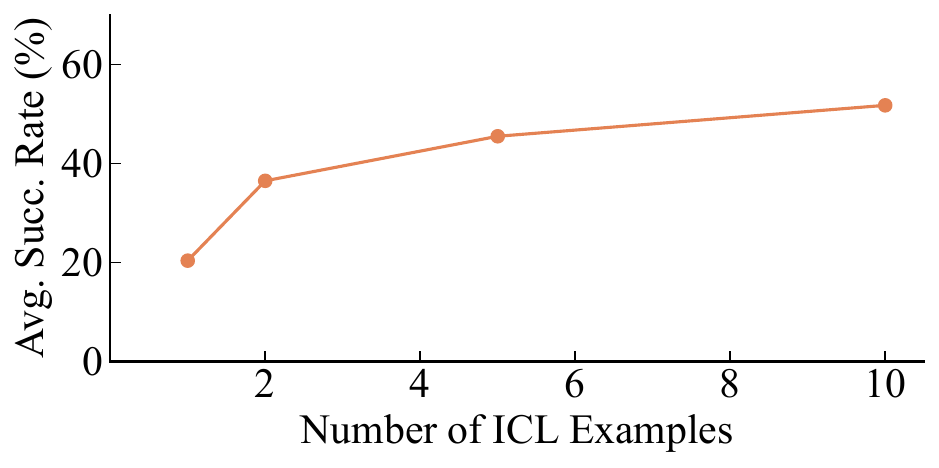}
    \vspace{-.6cm}
    \caption{{{\smodel} scaling}}
    \label{fig:k_shot}
\end{subfigure}
\hfill
\begin{subfigure}[t]{0.32\textwidth}
    \centering
    \includegraphics[width=\textwidth]{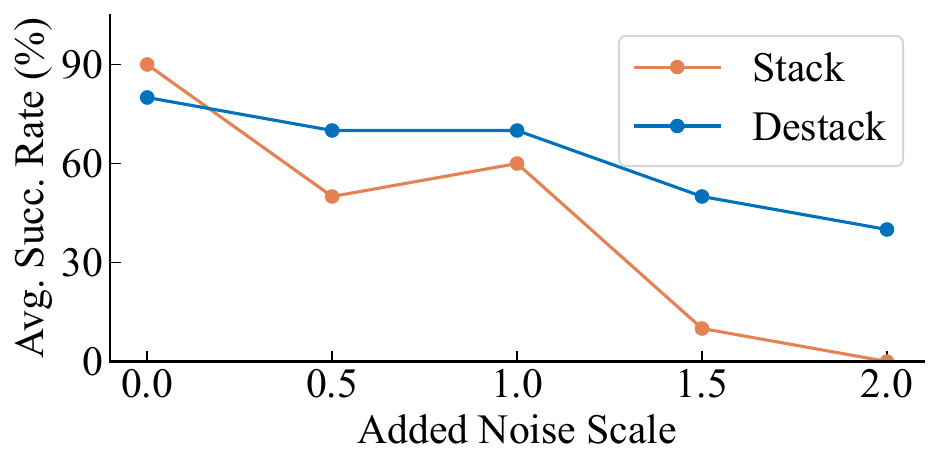}
    \vspace{-.6cm}
    \caption{{Pose estimation robustness}}
    \label{fig:pose}
\end{subfigure}
\caption{\textbf{Ablations on {\smodel}.} We demonstrate (a) {\smodel} with keyframes extraction outperforms uniform sampling with different intervals; (b) {\smodel}'s performance improves as the number of ICL examples increases; and (c) {\smodel} can achieve high success rates under moderate levels of pose estimation noise.}
\vspace{-1em}
\label{fig:abl}
\end{figure*}

\section{Analysis}

In the following, we analyze various components in {\smodel} and use 10 ICL examples by default. Additional experiments results are in Section~\ref{supp:add_res} of Supplementary.

\minisection{Keyframe extraction} We ablate our keyframe extraction scheme proposed in Section~\ref{sec:method:keyframe}. Specifically, we replace keyframe extraction with uniform action sampling, varying the sampling interval $k$ across $\{5, 10, 20, 40, 80\}$ frames. The uniformly sampled actions are then used to construct the output in ICL examples. Figure~\ref{fig:uniform} shows the comparison with the average success rate on the 16 RLBench tasks.

The results show that {\smodel} with keyframe extraction consistently outperforms the uniform sampling approach, with an average performance increase of nearly 20\%.  This trend holds across different sampling intervals. This is because smaller sampling intervals lead to longer and less effective ICL examples, which can confuse and mislead LLMs, while longer intervals risk missing crucial actions. Our keyframe extraction method mitigates these issues by only adding those critical actions to ICL examples.

\minisection{Number of ICL examples} We explore how the number of ICL examples influences the performance of {\smodel}. The default setup uses 10 ICL examples. To understand the impact of this number, we plot the average success rate of {\smodel} across 16 RLBench tasks while varying the number of examples. The results are shown in Figure~\ref{fig:k_shot}.

{\smodel}'s performance increases as ICL examples scale up. This is similar to the findings in~\cite{brown2020languagemodelsfewshotlearners} that apply ICL to evaluate LLMs on standard language benchmarks, where more ICL examples lead to better LLMs' performance. Moreover, {\smodel} achieves an average success rate of approximately 40\% even with only 2 ICL examples.
\begin{wraptable}{r}{0.24\textwidth}
\vspace{-10pt}
\begin{adjustwidth}{-10pt}{0pt}
\tablestyle{5pt}{1}
\begin{tabular}{lc}
 & \makecell{Average \\ Succ. Rate} \\
\Xhline{1.pt}
Llama3-8B-Inst.~\cite{dubey2024llama3herdmodels} &  28.3\\
GPT-4o mini & 44.8\\
Qwen2-7B-Inst.~\cite{qwen} & 48.8 \\
GPT-4 Turbo~\cite{openai2024gpt4technicalreport} & 51.8 \\
GPT-4o & 56.3
\end{tabular}
\vspace{-2pt}
  \captionsetup{width=.95\linewidth}
\caption{\textbf{Various LLMs with {\smodel}.}}
\label{tab:model}
\vspace{-10pt}
\end{adjustwidth}
\end{wraptable}
\minisection{Different LLMs} In our setting, {\smodel} employs GPT-4 Turbo as the default LLM. To evaluate the generality of our approach, we replace GPT-4 Turbo with other popular LLMs and assess {\smodel}'s performance on 16 RLBench tasks. Our evaluation ranges from open-source models to those with only API access. As shown in Table~\ref{tab:model}, {\smodel} consistently achieves high success rates with a variety of LLMs. Moreover, we observe that stronger LLM leads to better performance of {\smodel}.

\minisection{Robustness to pose estimation error} To assess the impact of pose estimation accuracy on {\smodel}'s performance, we add Gaussian noise into the estimated object poses for both the ICL examples and the test samples. The added noise is scaled by a factor of $k \in \{0.5, 1, 1.5, 2\}$, relative to the original pose estimation error. We assess {\smodel} on two real-world tasks: ``stack cube'' and ``destack cube''. The average pose estimation errors for the cubes are 1.68 cm in translation and 4.61 degrees in rotation, corresponding to roughly 2\% of the environment's spatial range. Figure~\ref{fig:pose} shows the results. It can be seen that {\smodel} is still robust to moderate levels of pose estimation error.

\minisection{Open-loop \vs closed-loop} As described in Section~\ref{sec:method:extract_actions_and_pose}, {\smodel} only takes one observation of all the objects at the first timestep, which is an open loop method. Here, we explore the possibility of closed-loop planning by adding more observations $\mathcal{O}$ from a greater number of keyframes.  For each ICL example, we add another observation at each keyframe, resulting in a new prompt that contains multiple pairs of observations and corresponding actions. We observe that this only leads to a 0.7\% increase in the average success rate. This is likely because the objects and the environment in our tasks are static and do not change during evaluation.

    \begin{wraptable}{r}{0.24\textwidth}
    
\begin{adjustwidth}{-10pt}{0pt}
\vspace{-12pt}
\tablestyle{3pt}{1}
\begin{tabular}{lccc}
 &   destack & \makecell{push \\ buttons} \\
\Xhline{1.0pt}
{\smodel} & 80 & 100 \\
Octo~\cite{octo_2023} & 40 & 20 \\
LLARVA~\cite{niu2024llarvavisionactioninstructiontuning} & 100 & 80 \\
\end{tabular}
\vspace{-2pt}
  \captionsetup{width=.95\linewidth}
\caption{\textbf{Comparison to supervised methods.}}
\label{tab:supervised}
\vspace{-10pt}\end{adjustwidth}
\end{wraptable}
\minisection{Supervised methods comparison} We also compare our method {\smodel} to the latest robotics frontier models Octo~\cite{octo_2023} and LLARVA~\cite{niu2024llarvavisionactioninstructiontuning} on two of the real-world tasks. The results are shown in Table~\ref{tab:supervised}. Interestingly, {\smodel} can achieve competitive performance when compared to frontier models that are trained on thousands of robotic episodes~\cite{openxembod}.

\section{Conclusion}

Building upon the recent success of LLMs, our proposed framework {\smodel} represents a significant advancement in applying ICL for robotics. In particular, our framework enables off-the-shelf text-only LLMs to directly predict robot actions through ICL demonstrations without training. We have demonstrated strong performance over several zero-shot and ICL baselines in both simulated and real-world settings. Our work marks a meaningful step forward and encourages research in applying ICL to various robotics applications. 

While {\smodel} offers substantial benefit, it is important to recognize certain limitations in our approach. First, LLMs can only perform ICL to predict robotic actions every few seconds, while some robots (\eg, humanoids) often require high-frequency controls (more than 10 Hz). Second, while we have shown {\smodel} can complete various manipulation tasks on a single hand robot, how to apply it to robotic tasks involving dynamic environments or in more complex settings (\eg, bimanual manipulation and whole-body control) remains unclear. We hope our work can encourage future research to explore these directions.

\section*{ACKNOWLEDGMENT}
We would like to thank Max Fu, Baifeng Shi, Brandon Huang, and Chancharik Mitra, Boya Zeng, and Zhuang Liu for helpful feedback, and Ilija Radosavovic for stimulating discussions. This project was partly supported by the BAIR's industrial alliance programs and the BDD program. 

\clearpage
\newpage
{\small
\bibliographystyle{IEEEtran}
\bibliography{main}
}

\clearpage
\newpage
\section*{Supplementary Material for ``{\smodel}''}
\renewcommand{\thesubsection}{\arabic{subsection}}
\setcounter{subsection}{0}
Here, we provide additional information about experiment results, implementation details, and qualitative examples. Specifically, \Secref{supp:add_res} provides more experiment results, \Secref{supp:impl} provides additional implementation details for both simulation and real-world experiments, and \Secref{supp:qual} provides qualitative visualizations of real-robot experiments.

\section{Additional Results}
\label{supp:add_res}

We present several additional experiments that further demonstrate the benefits of our {\smodel} framework. 

\minisection{Direct action prediction} Our method {\smodel} predicts robot actions directly through ICL examples. Instead, KAT~\cite{dipalo2024kat} recently has shown it is also possible to first transform each robot action into action tokens (triplets of 3D points) and then predict action tokens via ICL.

\begin{figure*}[t]
\centering
\begin{subfigure}[t]{0.32\textwidth}
    \centering
    \includegraphics[width=\textwidth]{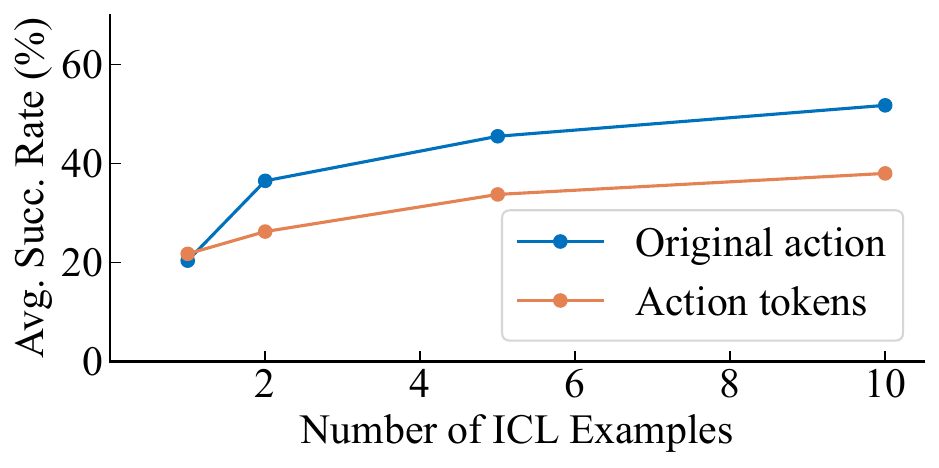}
    \caption{{action prediction \vs action tokens}}
    \label{fig:action}
\end{subfigure}
\hfill
\begin{subfigure}[t]{0.32\textwidth}
    \centering
    \includegraphics[width=\textwidth]{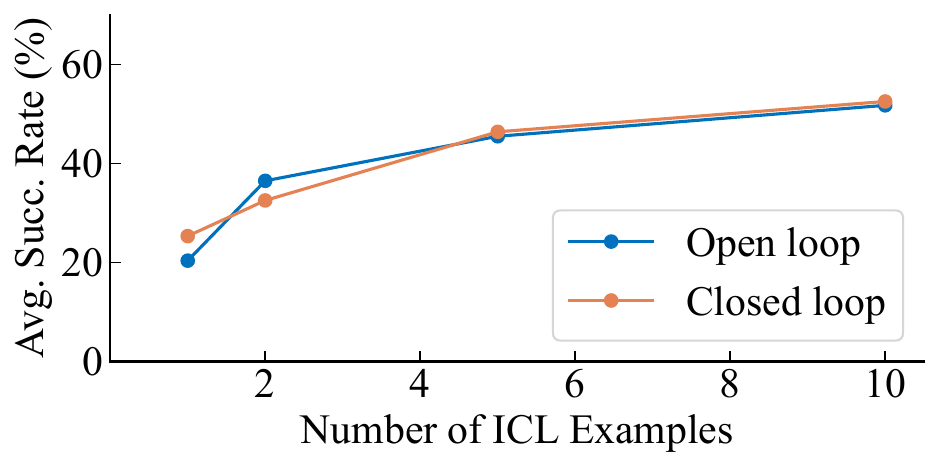}
    \caption{{Open loop \vs closed loop}}
    \label{fig:obs}
\end{subfigure}
\hfill
\begin{subfigure}[t]{0.32\textwidth}
    \centering
    \includegraphics[width=\textwidth]{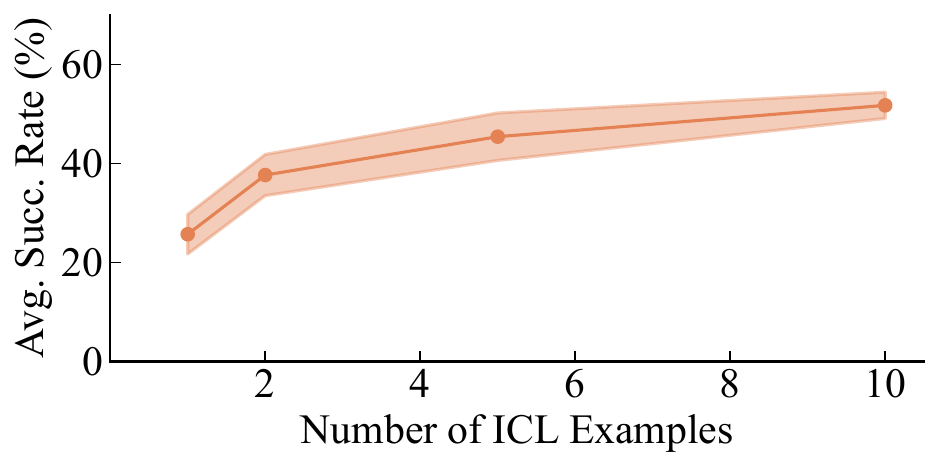}
    \caption{{Different system prompts}}
    \label{fig:prompt}
\end{subfigure}
\caption{\textbf{Additional experiments on {\smodel}.} We demonstrate (a) {\smodel} with origin action performs better than that with action tokens; (b) open-loop {\smodel} does not boost the performance by a large margin; and (c) {\smodel} can achieve a consistent high accuracy with different system prompts (light orange is standard deviation across prompts).}
\label{fig:add}
\end{figure*}

To evaluate our design choice for {\smodel}, we transform each robot action from the ICL examples into action tokens, and convert the predicted action tokens from LLMs during the test time back to standard 6-DoF actions. Figure~\ref{fig:action} shows the average performance of {\smodel} with action tokens across 16 RLBench tasks. The results indicate our method does not benefit from performing ICL on action tokens. Thus, our default setting in {\smodel} is to directly predict robot actions.

\minisection{Open-loop Vs. closed-loop} By default, {\smodel} is an open-loop method that only takes one observation at the first timestep. Here, we test a closed-loop approach by adding more observations from keyframes. Specifically, we form an ICL example by combining multiple pairs of observations and actions at each keyframe. Figure~\ref{fig:obs} shows the comparison across 16 RLBench tasks. The results indicate the closed-loop approach has minimal performance improvement. Thus, we opt for the open-loop approach for {\smodel}: taking a single observation at the first timestep.

\minisection{Different system prompts} Throughout the paper, {\smodel} employs the same system prompt (Figure~\ref{fig:teaser}) to form the ICL prompt. However, recent studies~\cite{min2022rethinkingroledemonstrationsmakes, holtzman2022surfaceformcompetitionhighest, liu2021makesgoodincontextexamples} have shown the performances of LLMs using ICL are highly sensitive to the prompt design. Here we would like to understand how sensitive {\smodel} is to our designed system prompt. Specifically, we form two new system prompts by asking GPT-4o to paraphrase the original one. We illustrate the original one as well as the two new counterparts below:

\begin{figure}[h]
\centering
\begin{tcolorbox}[colback=gray, colframe=black, width=.95\columnwidth, boxrule=0.5mm, left=0pt, right=0pt, top=1pt, bottom=1pt]
``You are a Franka Panda robot with a parallel gripper. We provide you with some demos in the format of observation>[action\_1, action\_2, ...]. Then you will receive a new observation and you need to output a sequence of actions that match the trends in the demos. Do not output anything else.''
\end{tcolorbox}
\vspace{-.25cm}
\caption*{(a) original prompt}
\vspace{-.25cm}
\end{figure}
\begin{figure}[h]
\centering
\begin{tcolorbox}[colback=gray, colframe=black, width=.95\columnwidth, boxrule=0.5mm, left=0pt, right=0pt, top=1pt, bottom=1pt]
``You are an end-effector Franka Panda robot equipped with a parallel gripper. We will give you a series of demonstrations in the format observation>[action\_1, action\_2, ...]. Afterward, you will receive a new observation, and your task is to generate a sequence of actions that align with the patterns shown in the demos. Make sure to only output the actions and nothing else.''
\end{tcolorbox}
\vspace{-.25cm}
\caption*{(b) first paraphrased prompt}
\vspace{-.5cm}
\end{figure}%
\begin{figure}[H]
\centering
\begin{tcolorbox}[colback=gray, colframe=black, width=.95\columnwidth, boxrule=0.5mm, left=0pt, right=0pt, top=1pt, bottom=1pt]
``You are a Franka Panda robot equipped with a parallel gripper. We will provide you with demonstrations in the format: observation>[action\_1, action\_2, ...]. Afterward, you will receive a new observation and must generate a sequence of actions that align with the patterns shown in the demos. Ensure that nothing else is included in your output.''
\end{tcolorbox}
\vspace{-.25cm}
\caption*{(c) second paraphrased prompt}
\vspace{-.25cm}
\end{figure}

Figure~\ref{fig:prompt} shows the results of {\smodel} with above system prompts. Overall, we can observe {\smodel}'s performance varies little across different system prompts. 
\section{Additional Implementation Details}
\label{supp:impl}

Here we provide additional details for both simulated RLBench and real experiments.

\subsection{RLBench Experiments}
\label{supp:rlbench}

{\smodel} is evaluated on 16 tasks from RLBench simulation. For each task, we only provide 10 ICL demonstrations and evaluate 25 times. During each evaluation, the objects' positions and orientations in the scene are randomized. Below, we describe the setup for each task along with its corresponding success criteria.

\minisection{Close jar} The task is to put a lid on a target jar. The success criteria is the lid being on top of the target jar and the robot gripper not grasping any object. 

\minisection{Slide block} The task is to slide a block onto a target square. The success criteria is some part of the block being on the specified target square. 

\minisection{Sweep to dustpan} The task is to sweep dirt particles into a target dustpan. The success criteria is all five dirt particles being inside the target dustpan. 

\minisection{Open drawer} The task is to open the bottom of a drawer. The success criteria is the joint of the drawer fully extended. 

\minisection{Turn tap} The task is to turn the left handle of a tap. The success criteria is the joint of the left handle being at least $90^\circ$ away from the starting position. 

\minisection{Stack blocks } The task is to stack any 2 of 4 total blocks on the green platform. The success criteria is 2 blocks being inside the area of the green platform. 

\minisection{Push button} The task is to push a single button. The success criteria is the target button being pressed. 

\minisection{Place wine} The task is to pick up a wine bottle and place it at the middle of a wooden rack. The success criteria is the placement of the bottle at the middle of the rack. 

\minisection{Screw bulb} The task is to pick up a light bulb from the stand and and screw it into the bulb stand. The success criteria is the bulb being screwed inside the bulb stand.

\minisection{Put in drawer} The task is to place a block into the bottom drawer. The success criteria is the placement of the block inside the bottom drawer. 

\minisection{Meat off grill} The task is to take a piece of chicken off the grill and put it on the side. The success criteria is the placement of the chicken on the side, away from the grill. 
 
\minisection{Stack cups} The task is to stack two cups inside the target one. The success criteria for this task is two cups being inside the target one. 

\minisection{Put in safe} The task is to pick up a stack of money and place it at the bottom shelf of a safe. The success criteria is the stack of money being at the bottom shelf of the safe. 

\minisection{Put in cupboard} The task is to place a target grocery inside a cupboard. The success criteria is the placement of the target grocery inside the cupboard. 

\minisection{Sort shape} The task is to pick up a cube and place it in the correct hole in the sorter. The success criteria is the cube being inside the corresponding hole.  

\minisection{Place cups} The task is to place a cup on the cup holder. The success criteria is the alignment of the cup's handle with any spoke on the cup holder.

\subsection{Real Robots Experiments} 
\label{supp:real_robot}

\begin{figure}
    \centering
    \includegraphics[width=.8\linewidth]{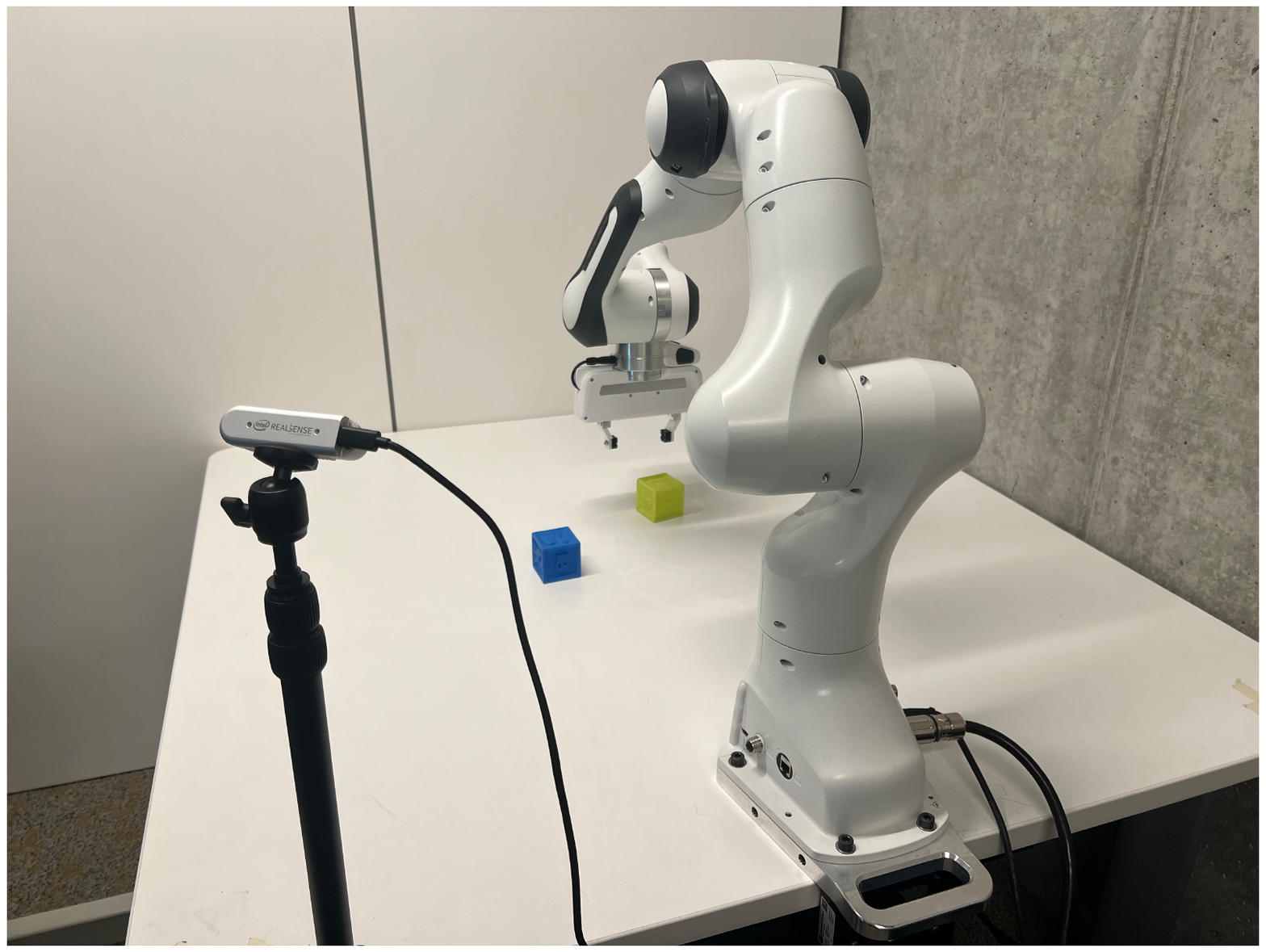}
    \caption{\textbf{Setup.} The real-robot setup with a Franka Emika Panda used for evaluating {\smodel}.}
    \label{fig:realsetup}
    \vspace{-1em}
\end{figure}

\minisection{Hardware Setup} We use a Franka Emika Panda robot with a parallel jaw gripper for real robot data collection and evaluations. A Intel RealSense D435 camera positioned on the left of the Franka robot provides a RGB-D visual observation, as shown in Figure \ref{fig:realsetup}. The RGB image is captured at 1920x1080 resolution, and the depth image is aligned to the resolution of the RGB image.

\minisection{Evaluation}
We evaluate {\smodel} on 6 real-world tasks. When evaluating {\smodel} on one task, we provide 10 ICL demonstrations for that task. Note for tasks that involve variations (\eg, ``stack cube'', ``destack cube'', ``push button'', and ``push multiple buttons''), 10 ICL demonstrations are formed from  different variations. {\smodel} is evaluated on each task for 10 times. Each time the position and orientation of the object in the scene is randomized. Below, we describe the setup for each task along with its corresponding success criteria as well as the names of objects and the task instruction we provide in each task.

\minisection{Stack cube} The task is to stack a blue/yellow cube on top of the other. The success criteria is the correct cube being on the top of the other and the robot gripper not grasping anything. The names of objects we provide are ``blue cube'' and ``yellow cube''. The language instruction is ``stack the blue/yellow cube on the yellow/blue cube.''

\minisection{Destack cube} The task is to destack a blue/yellow cube from the top of the other. The success criteria is the cube landed on the table. The names of objects we provide are ``blue cube'' and ``yellow cube''. The language instruction is ``destack the blue/yellow cube that is on the yellow/blue cube.''

\minisection{Push button} The task is to push a red/yellow/green/blue button. The success criteria is the specified button being pushed. The names of objects we provide are ``red button'', ``yellow button'', ``green button'', and ``blue button''. The language instruction is ``push the red/yellow/green/blue button''.

\minisection{Close laptop} The task is to close the screen of the laptop. The success criteria is the laptop screen being closed completely. The names of objects we provide are ``laptop''. The language instruction is ``close the laptop''.

\minisection{Unplug cable} The task is to unplug the cable of the laptop. The success criteria is the cable being disconnected from the laptop. The names of objects we provide are ``laptop'' and ``cable''. The language instruction is ``unplug the laptop''.

\begin{figure*}[t]
    \tablestyle{.8pt}{0.9}
\centering
  \begin{tabular}{@{}ccccccc@{}}
  \includegraphics[width=.16\linewidth]{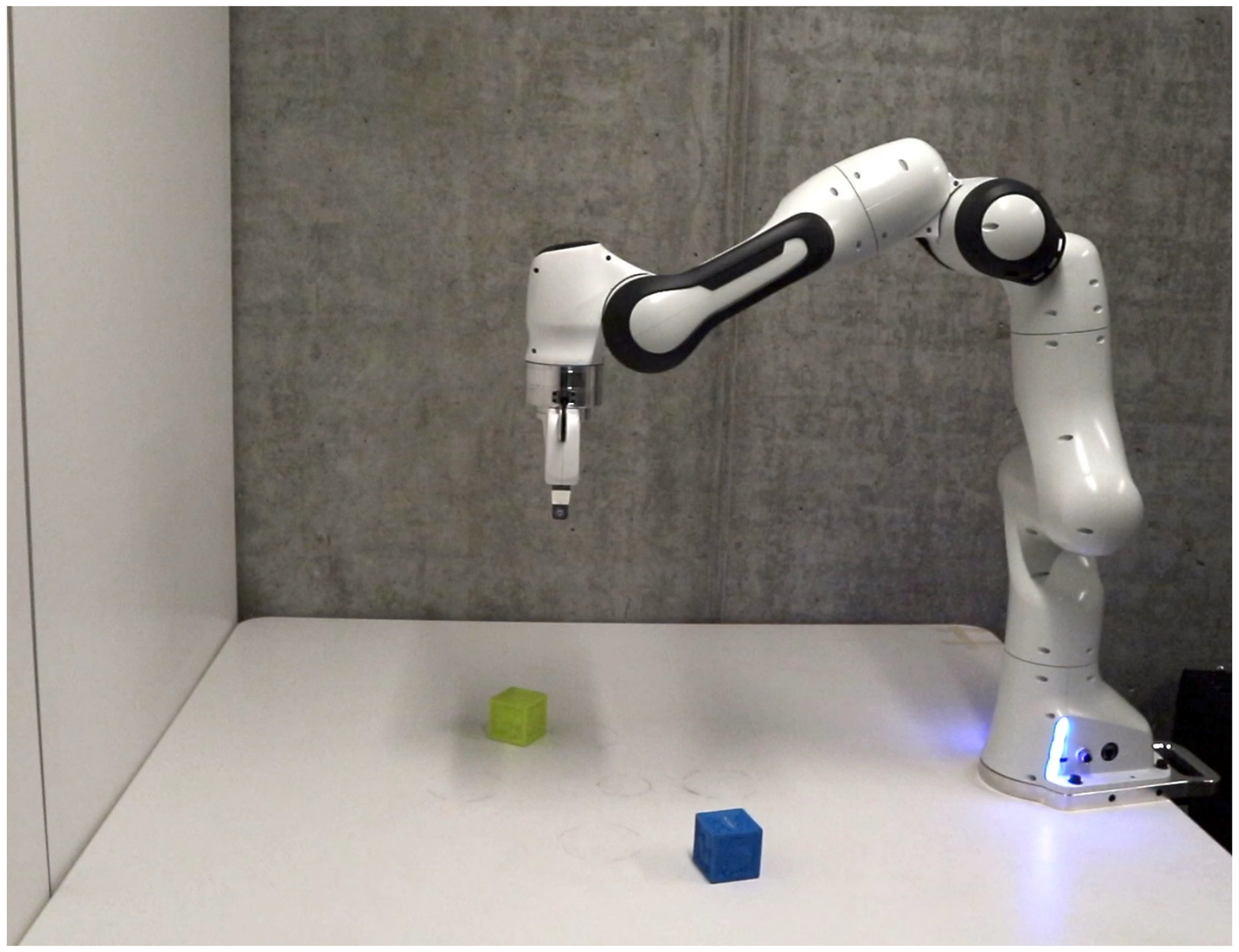} & 
  \includegraphics[width=.16\linewidth]{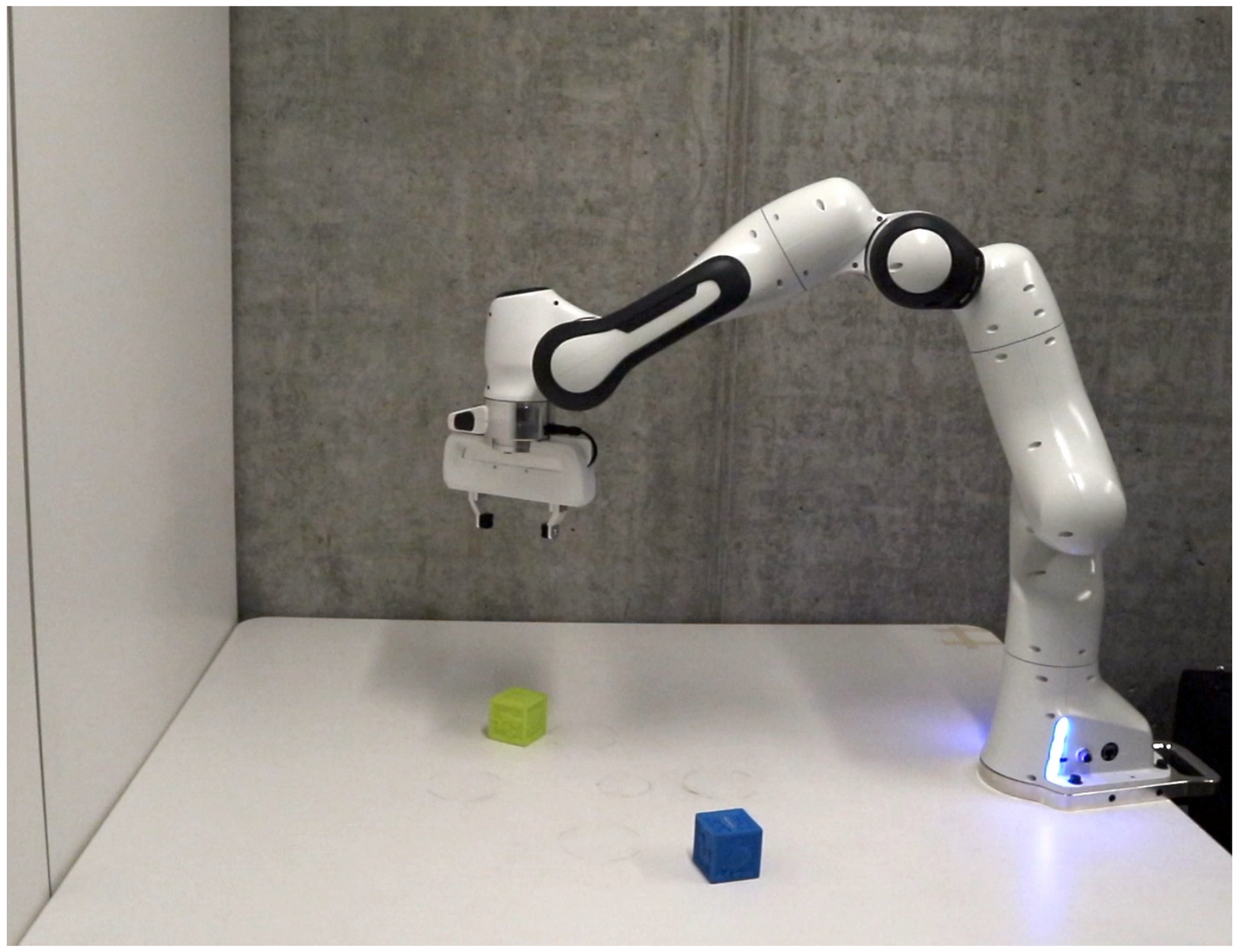} & 
  \includegraphics[width=.16\linewidth]{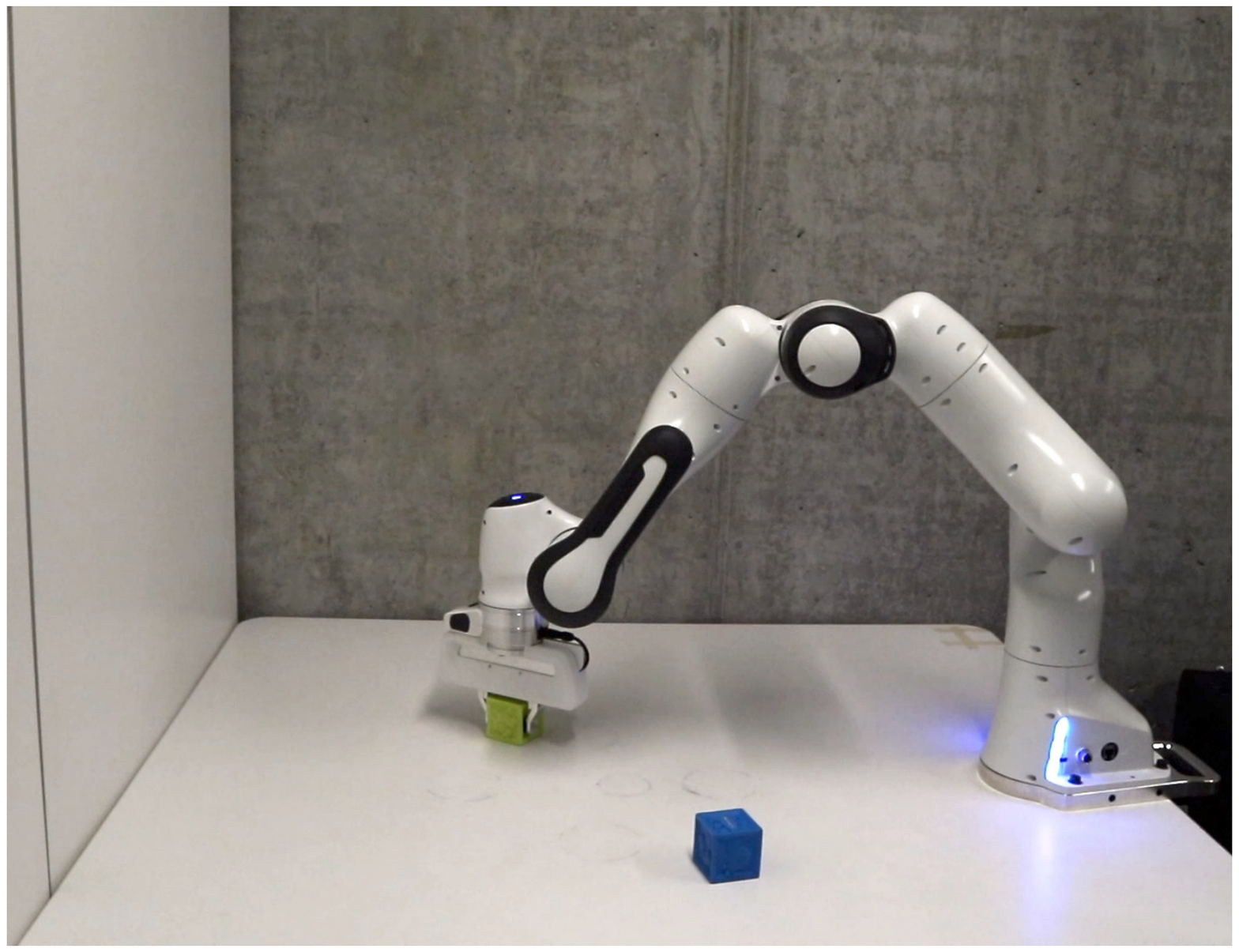} & 
  \includegraphics[width=.16\linewidth]{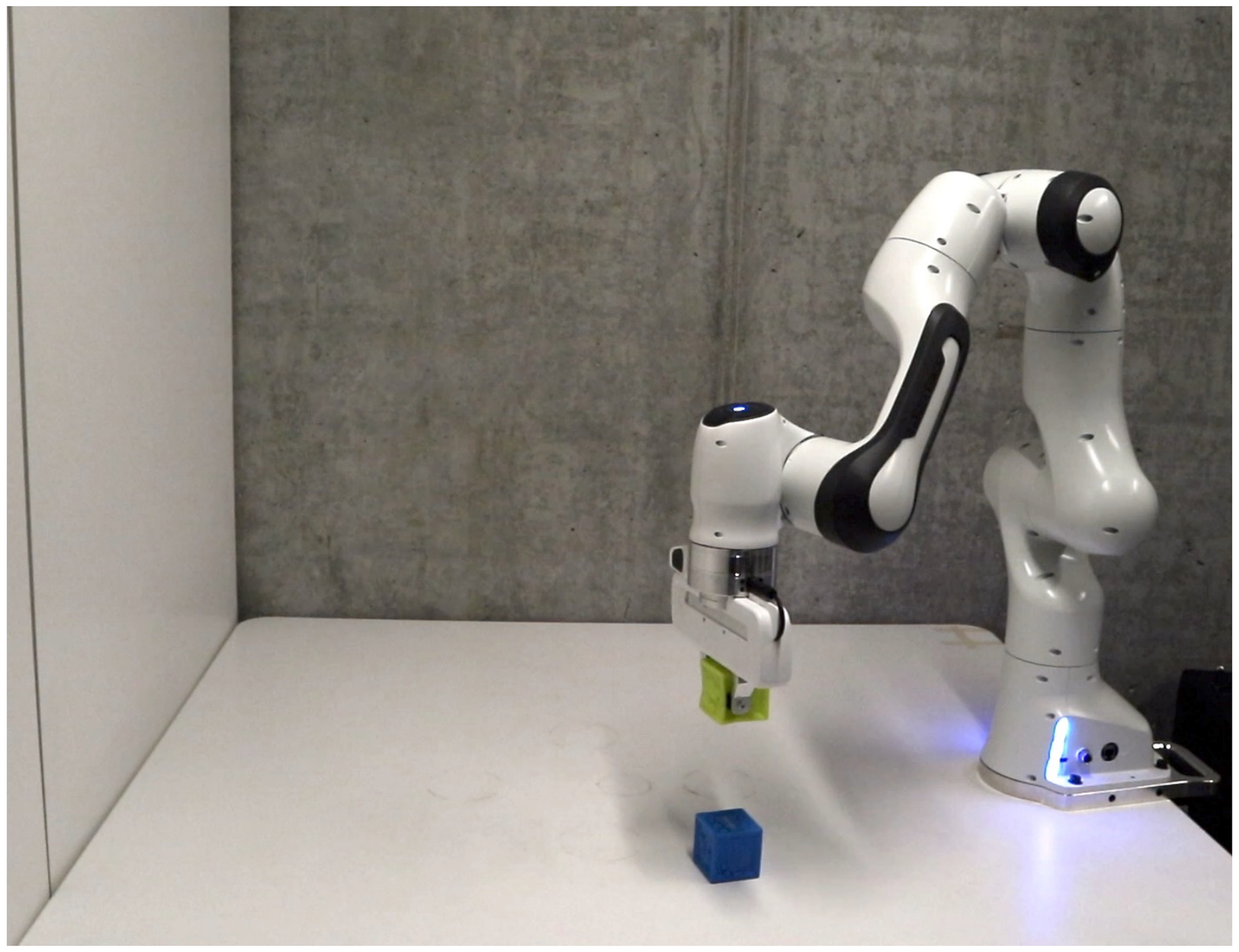} & 
  \includegraphics[width=.16\linewidth]{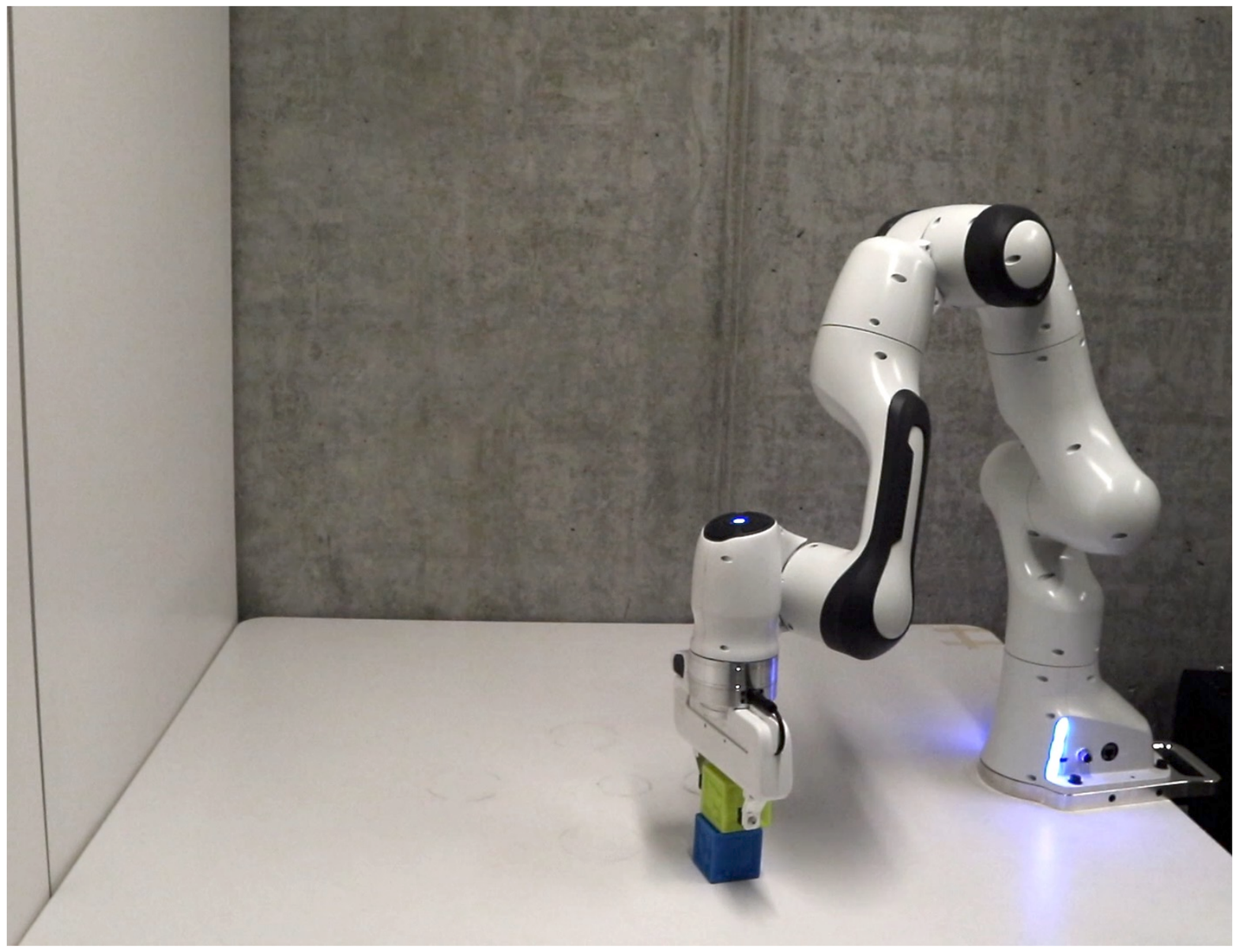} & 
  \includegraphics[width=.16\linewidth]{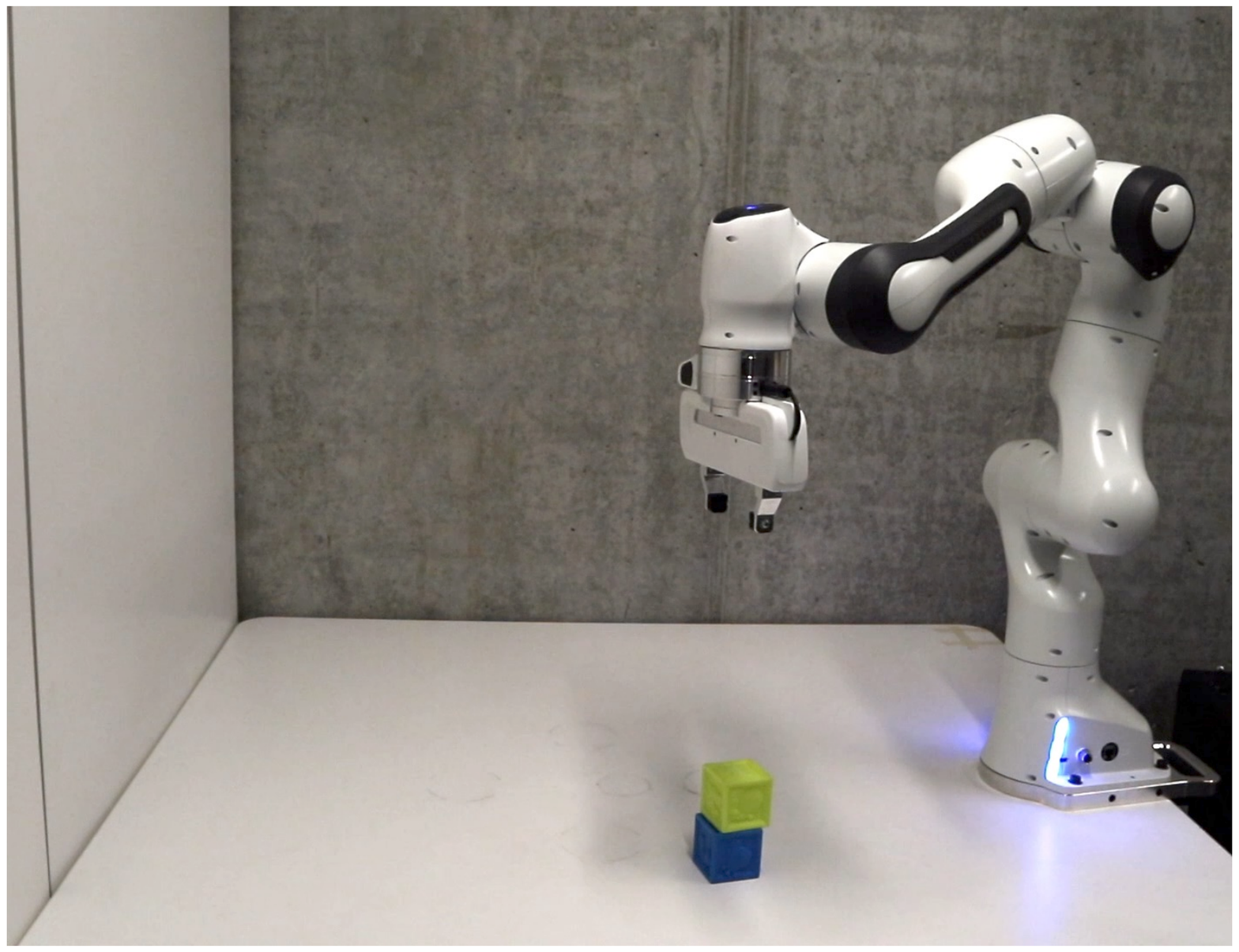} \\
  
  \multicolumn{6}{c}{(a) stack cube} \\
  
  \includegraphics[width=.16\linewidth]{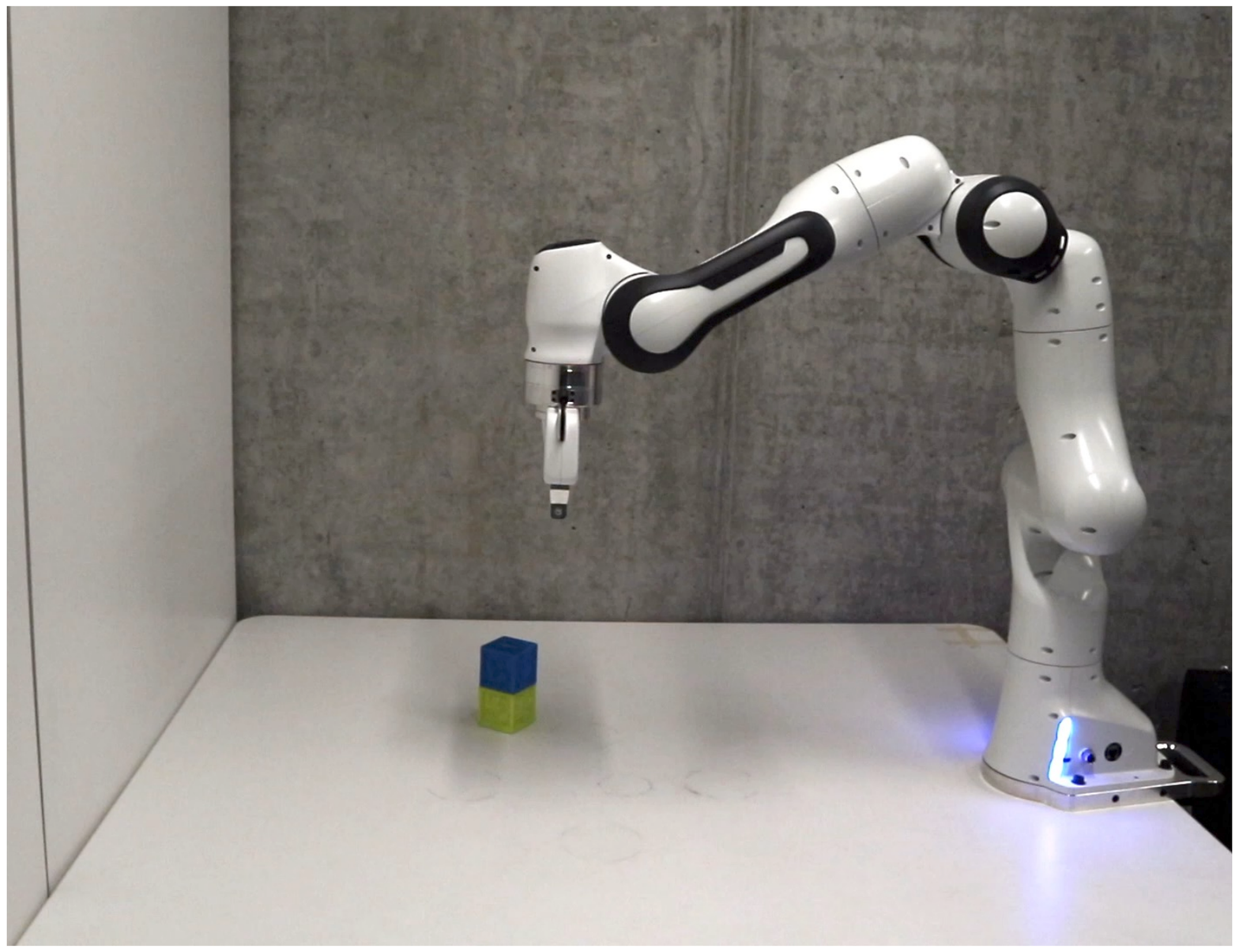} & 
  \includegraphics[width=.16\linewidth]{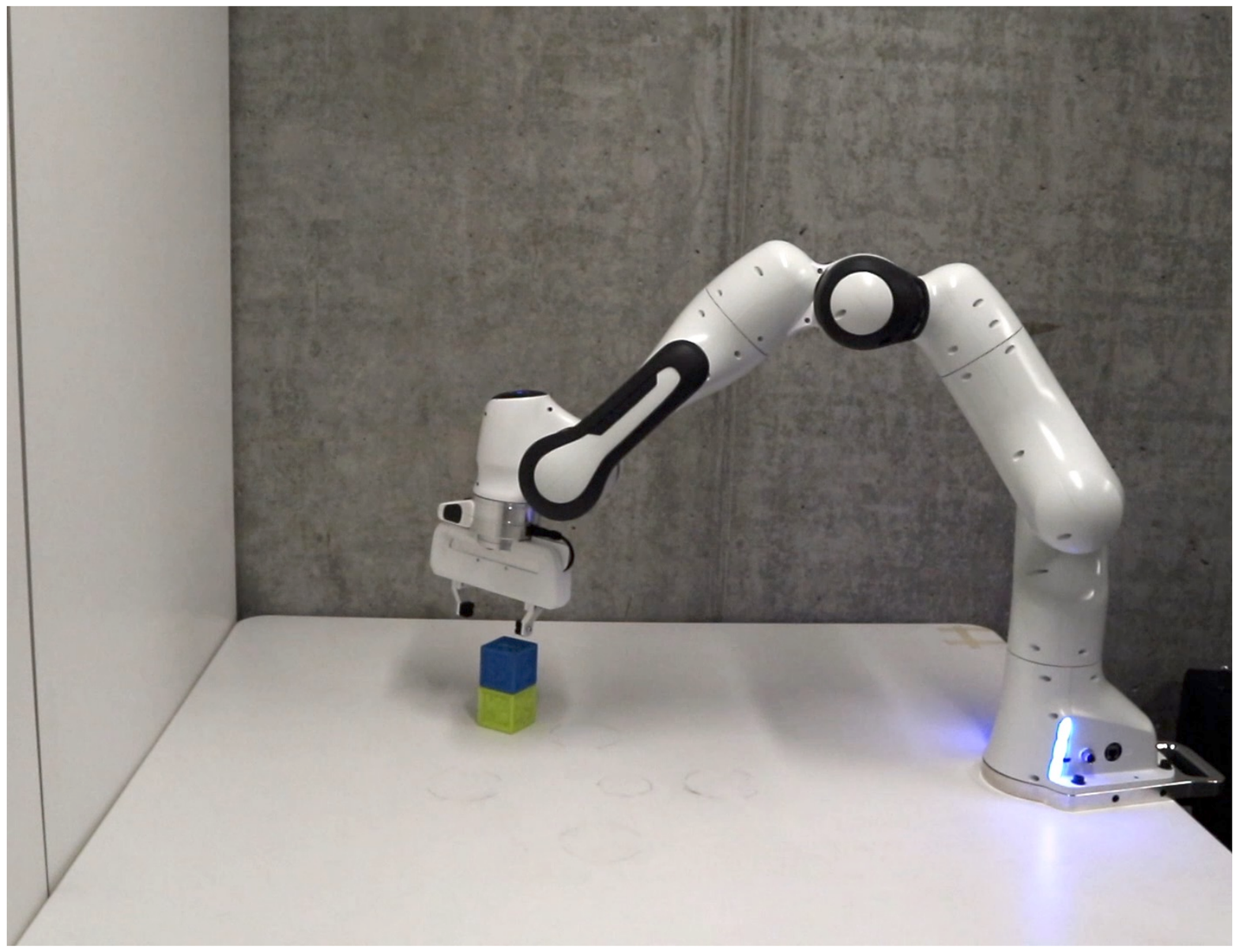} & 
  \includegraphics[width=.16\linewidth]{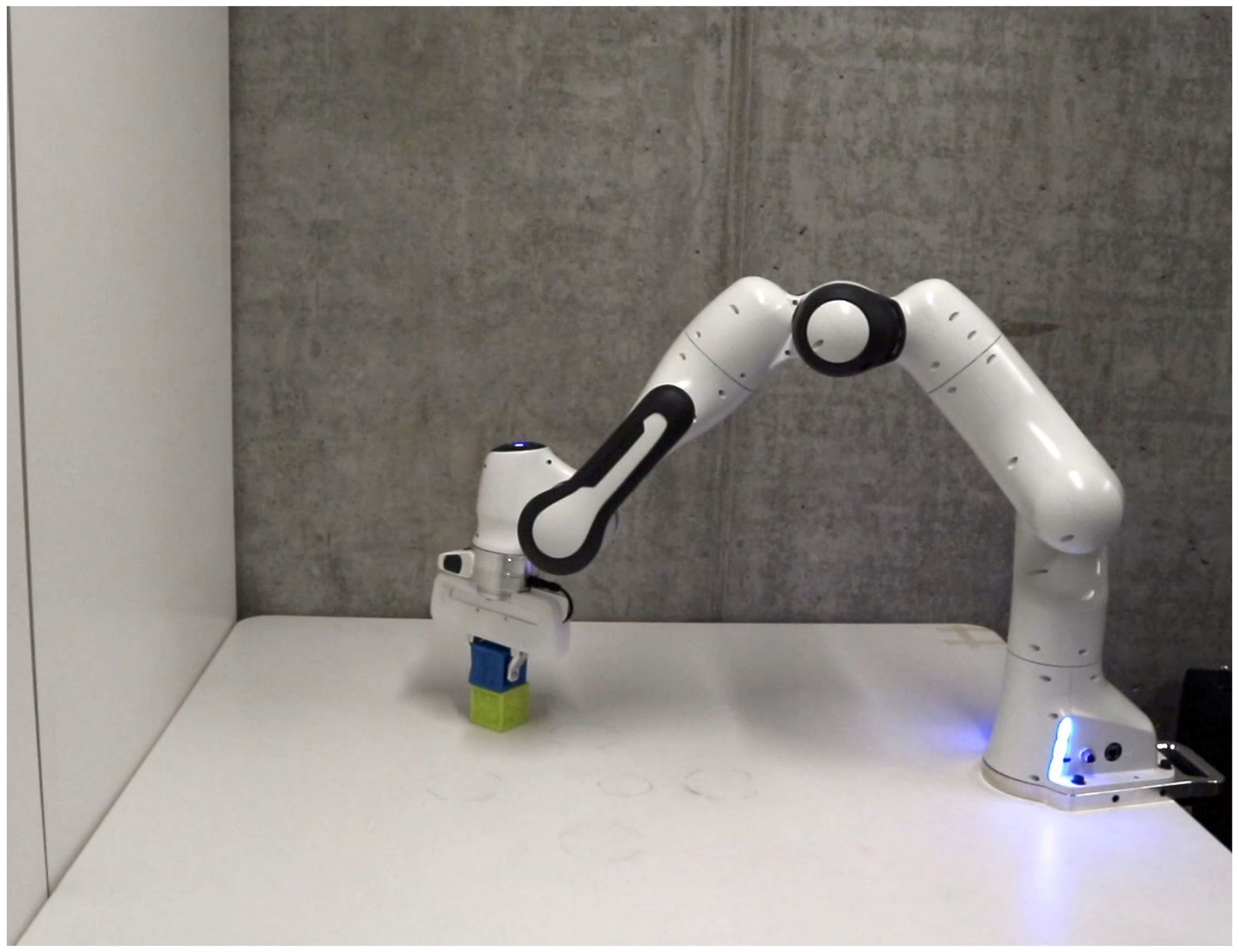} & 
  \includegraphics[width=.16\linewidth]{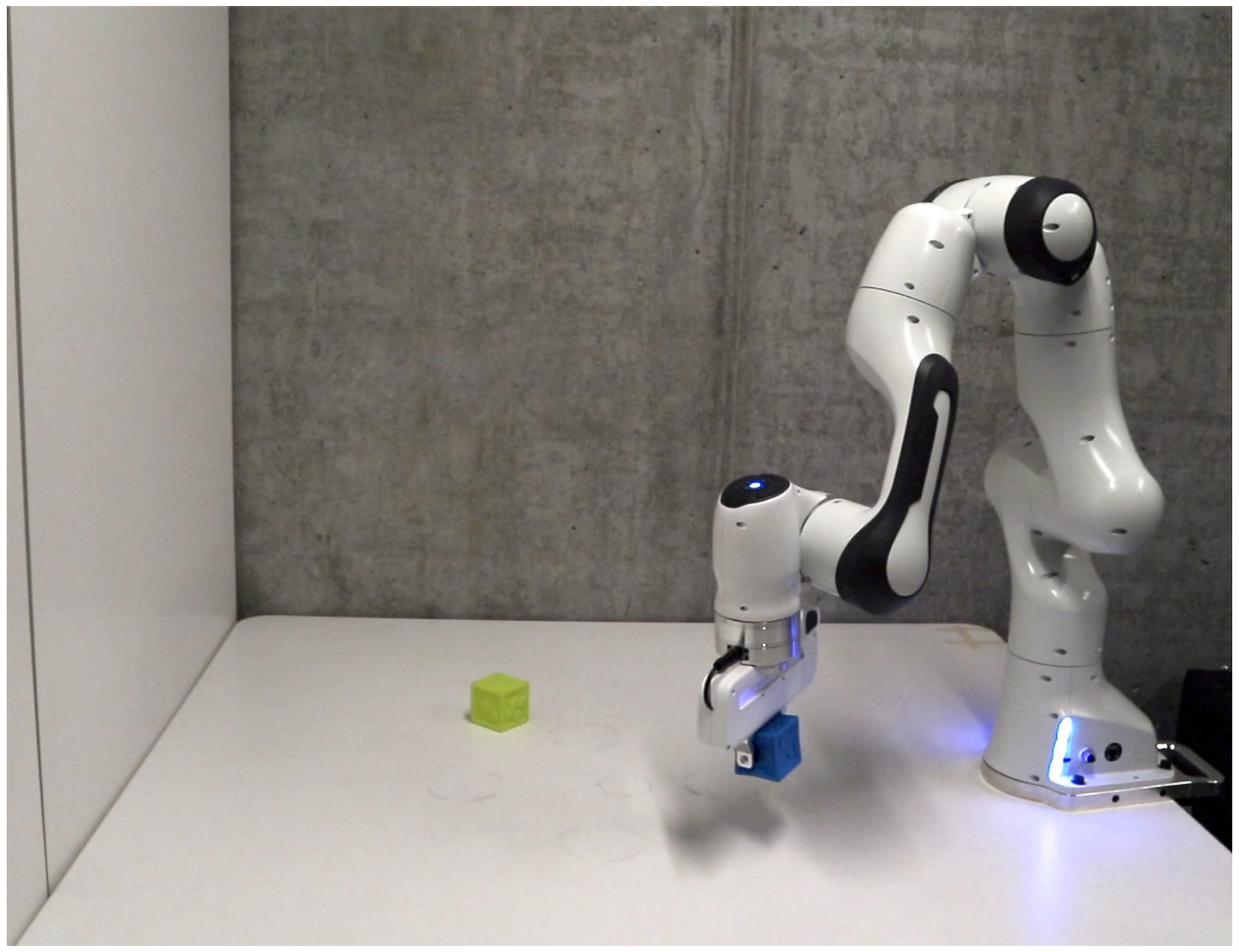} & 
  \includegraphics[width=.16\linewidth]{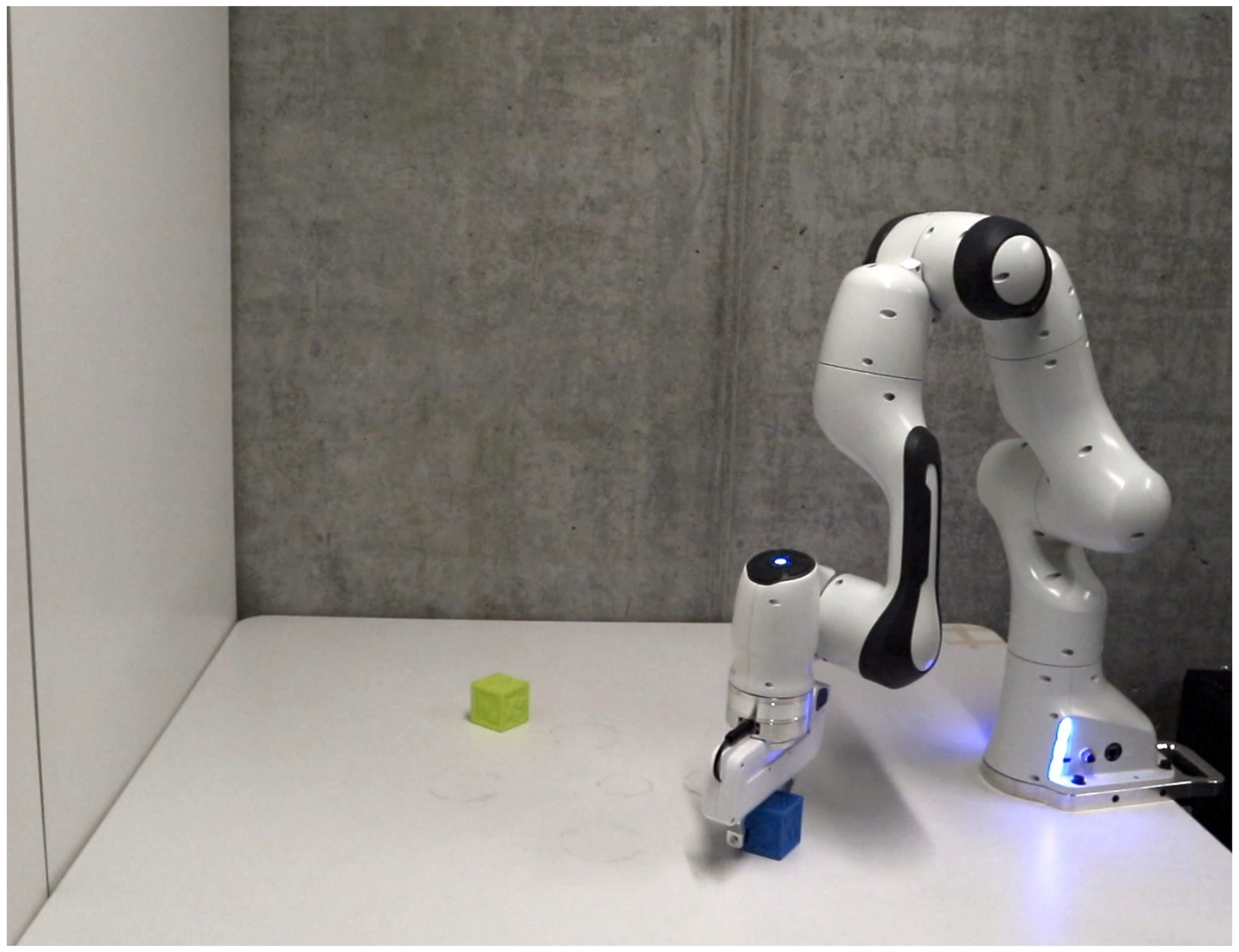} & 
  \includegraphics[width=.16\linewidth]{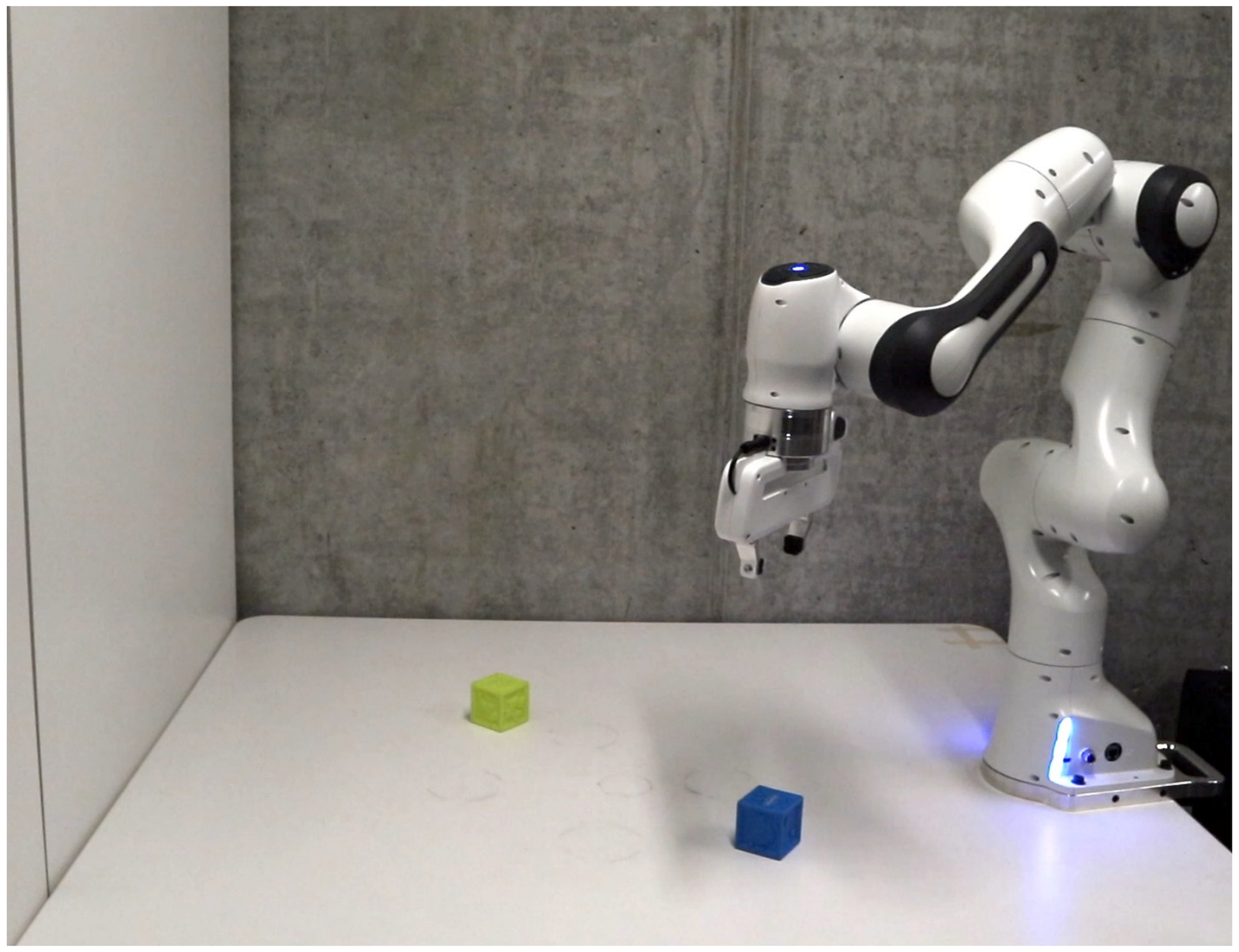} \\

  \multicolumn{6}{c}{(b) destack cube} \\

  \includegraphics[width=.16\linewidth]{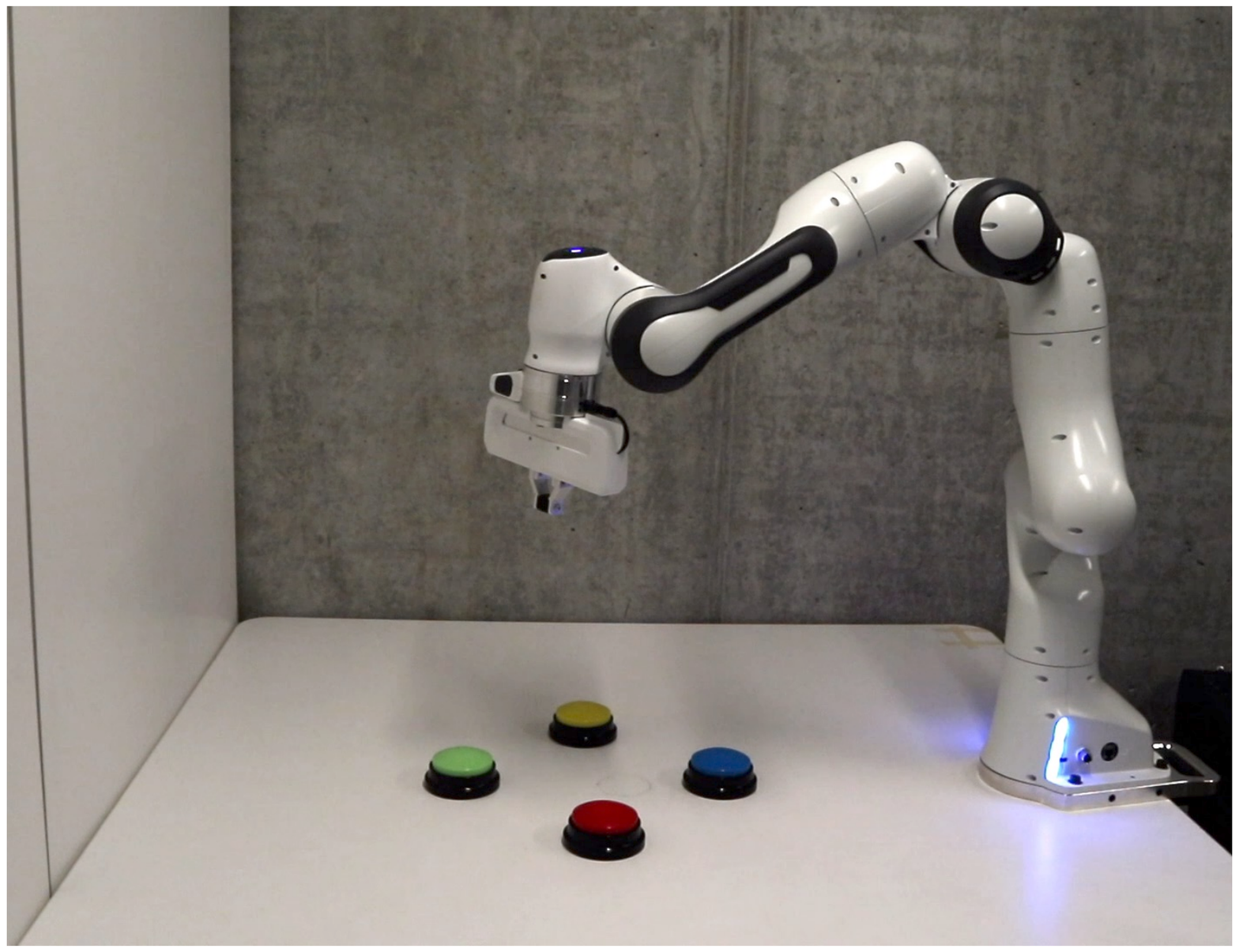} & 
  \includegraphics[width=.16\linewidth]{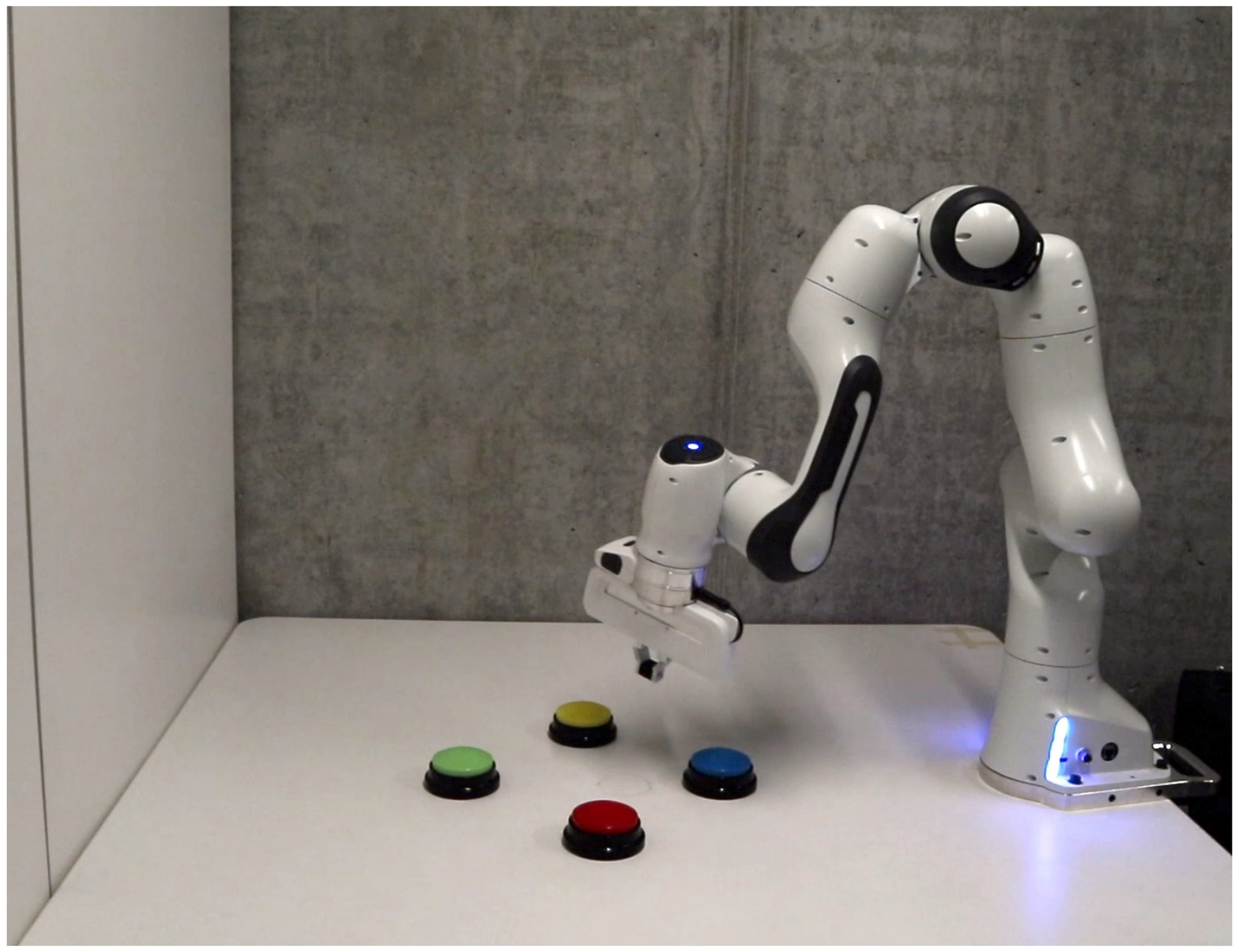} & 
  \includegraphics[width=.16\linewidth]{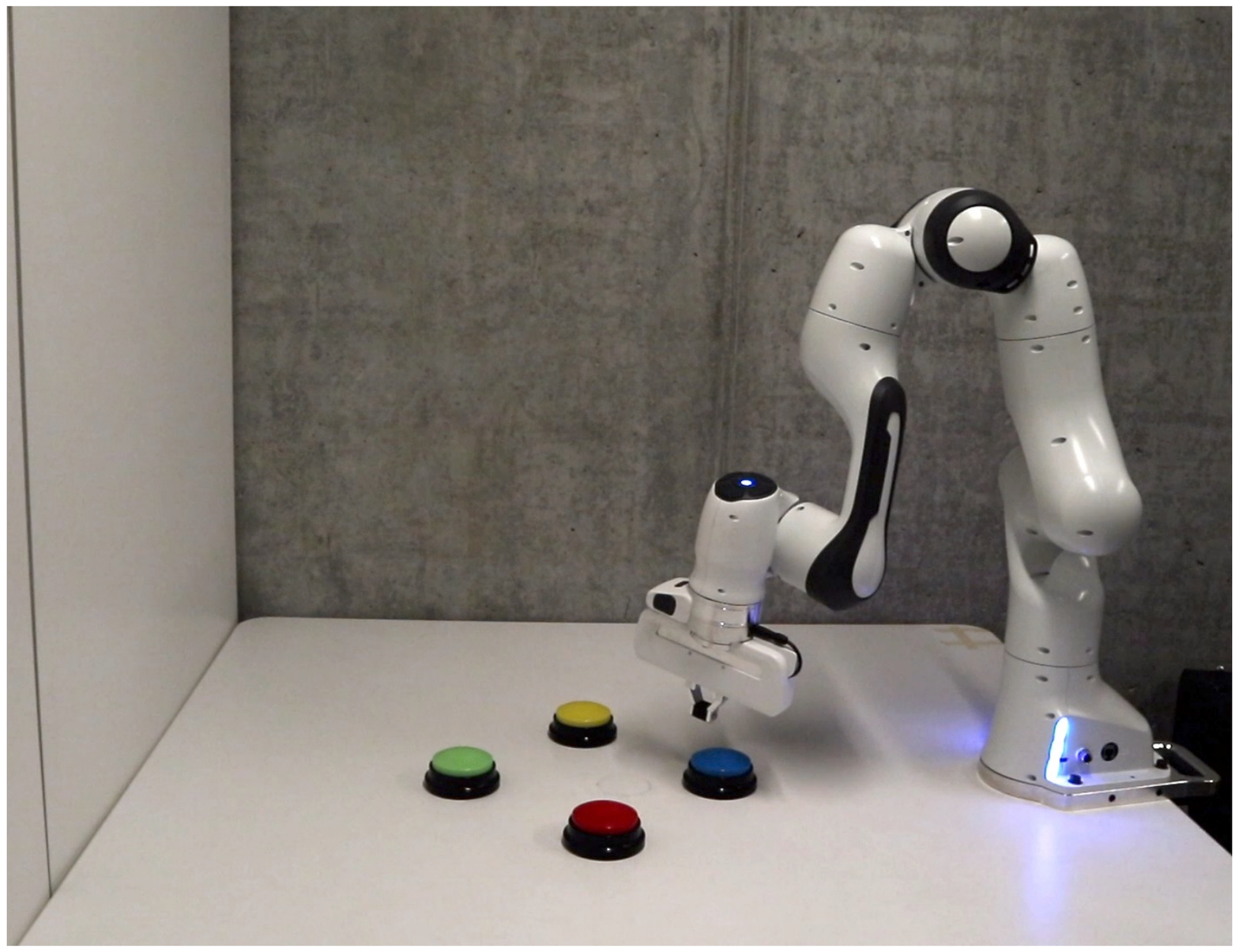} & 
  \includegraphics[width=.16\linewidth]{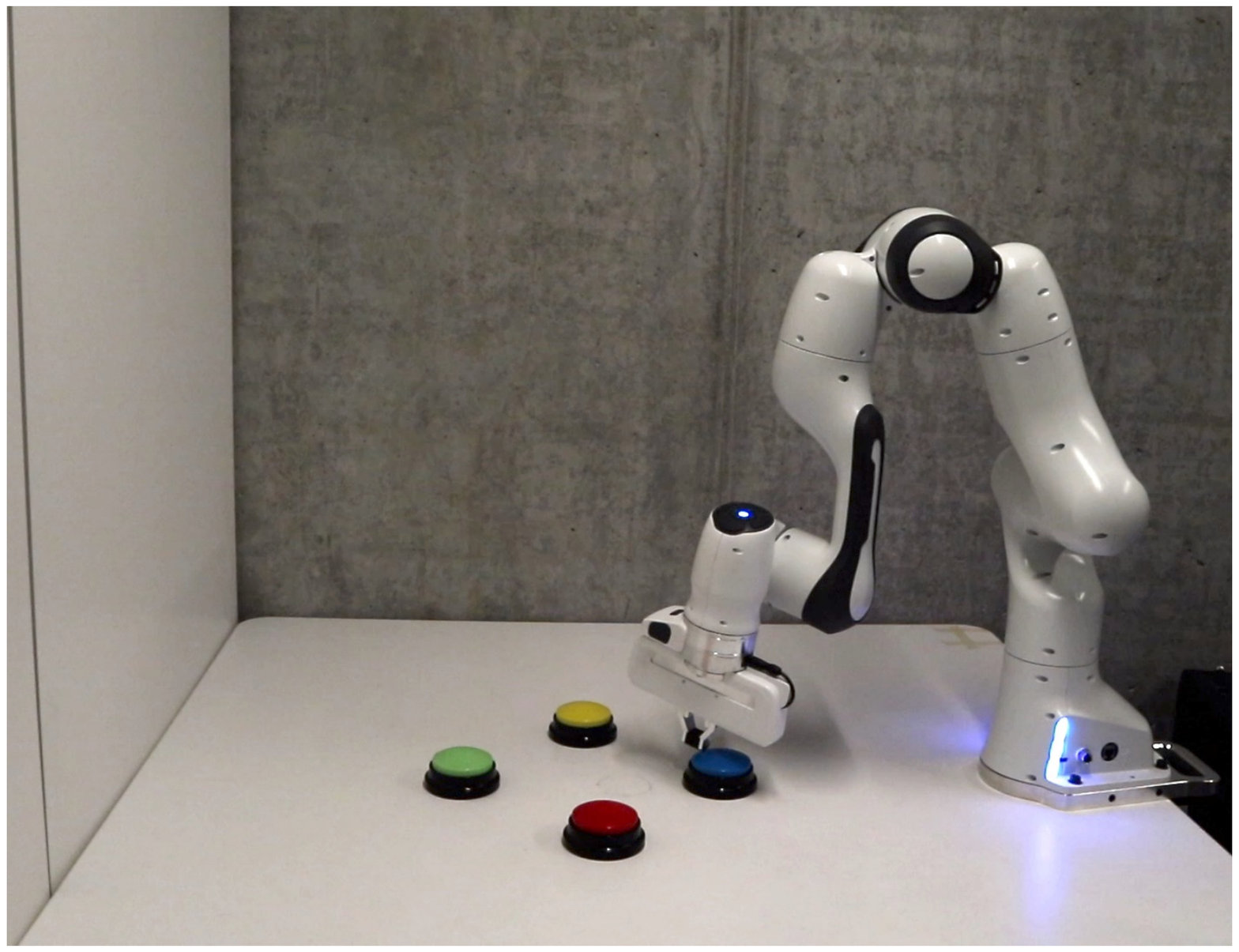} & 
  \includegraphics[width=.16\linewidth]{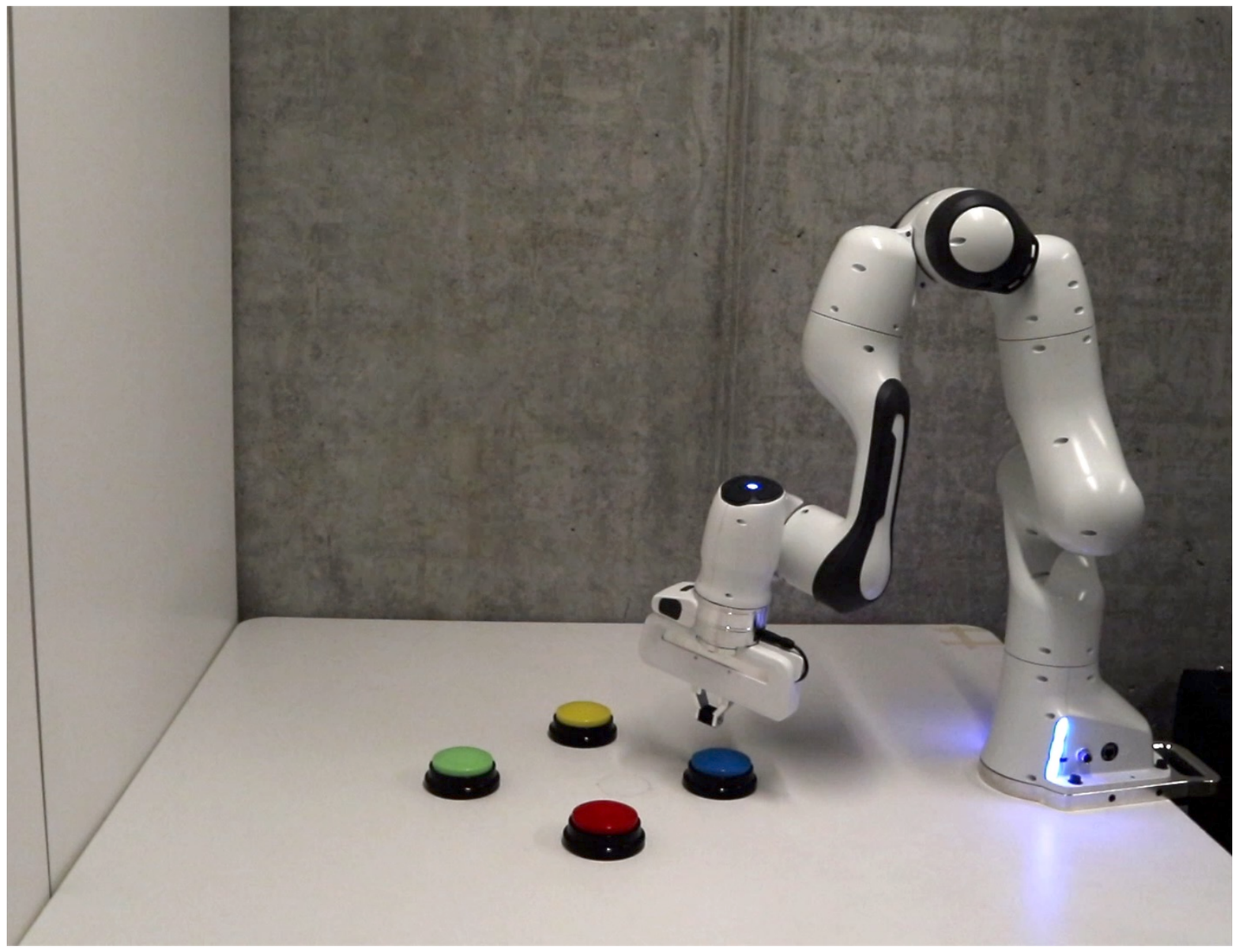} & 
  \includegraphics[width=.16\linewidth]{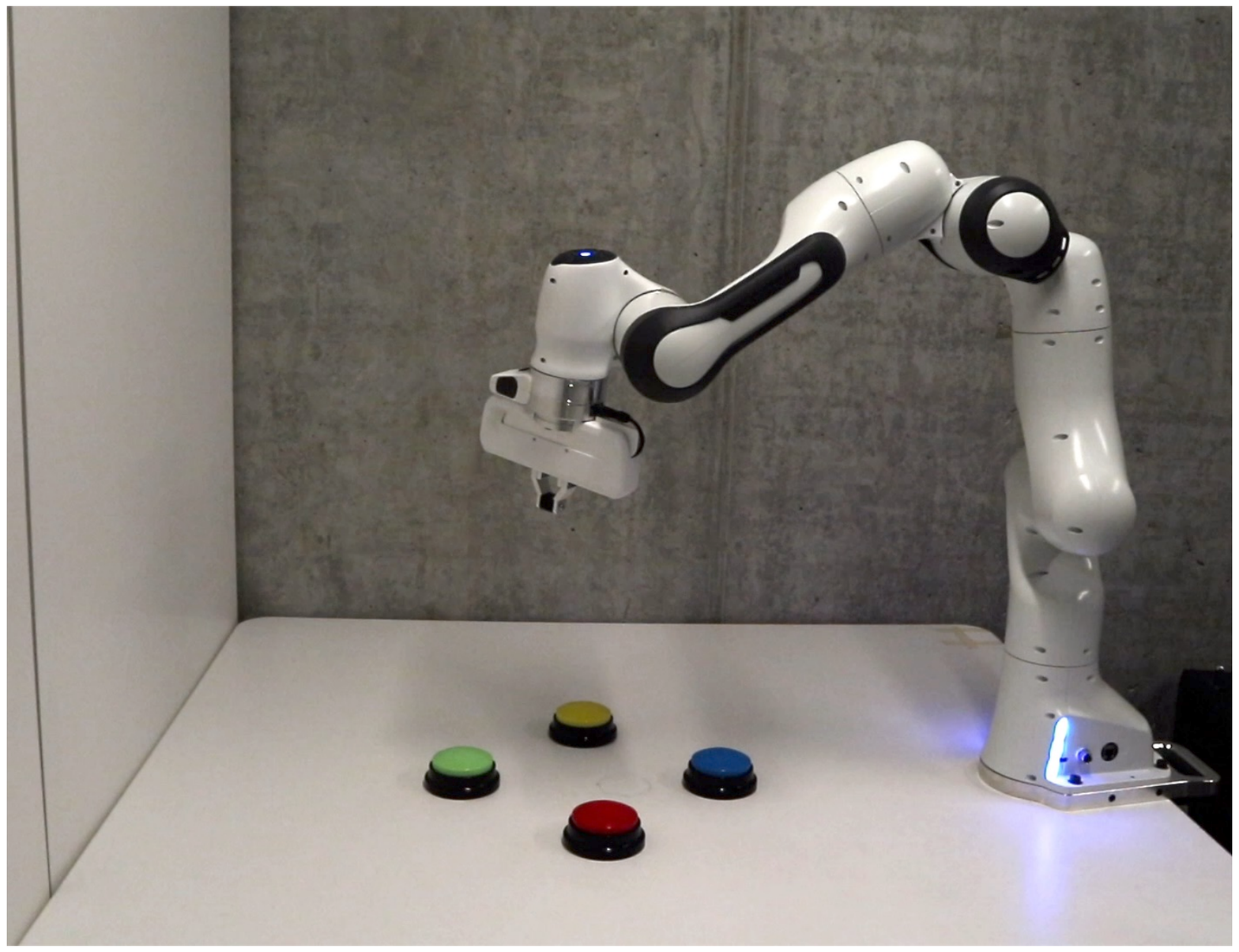} \\

  \multicolumn{6}{c}{(c) push button} \\

  \includegraphics[width=.16\linewidth]{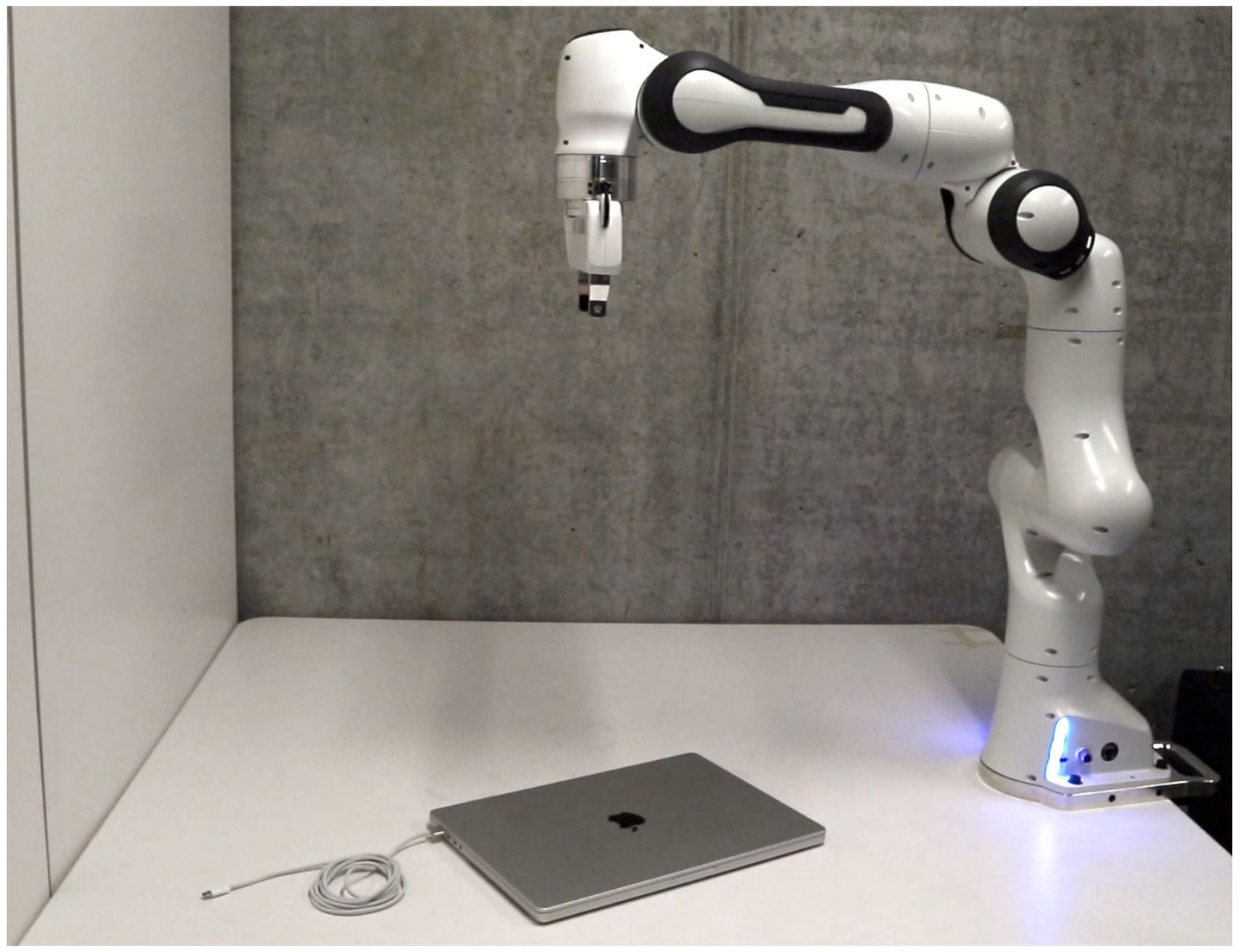} & 
  \includegraphics[width=.16\linewidth]{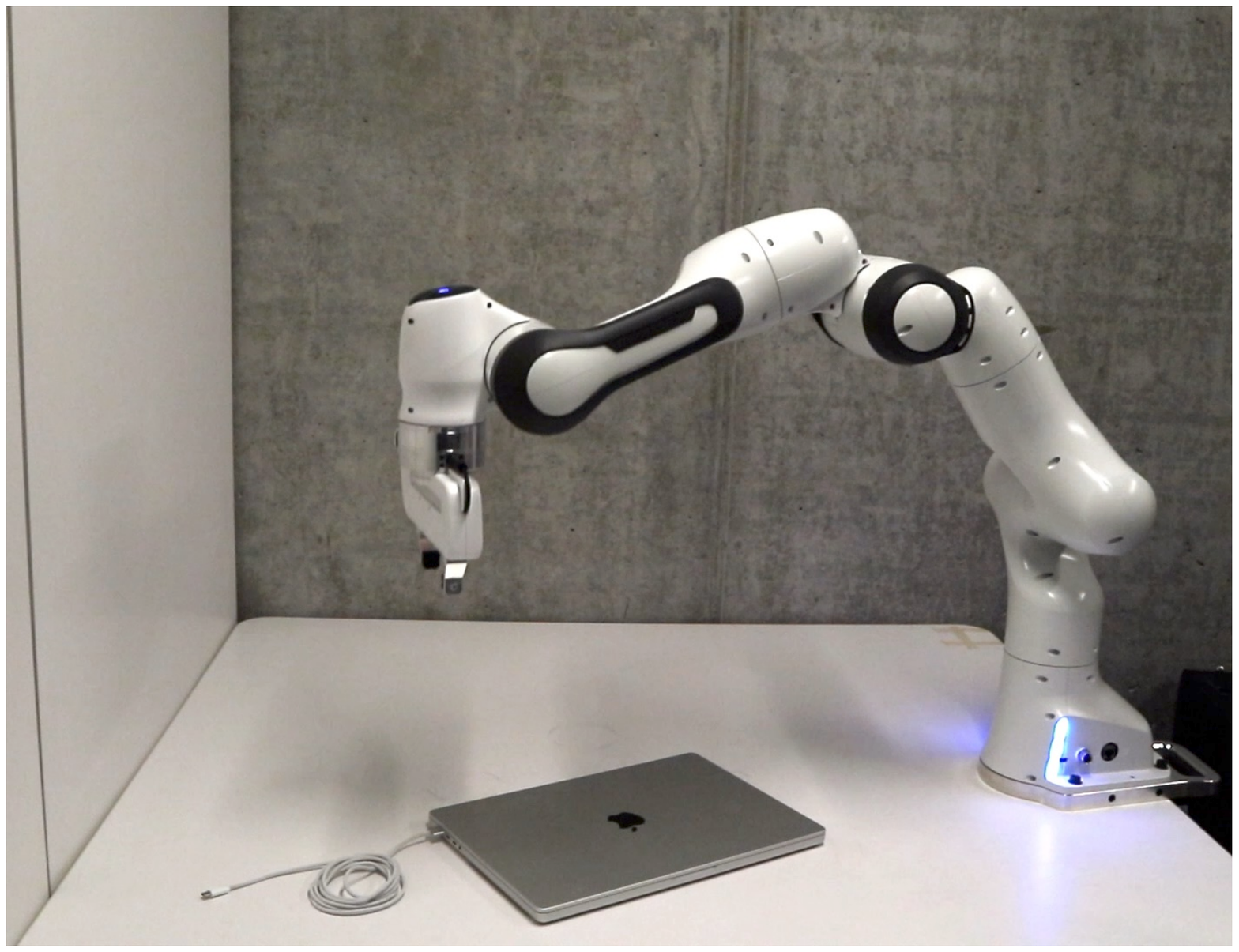} & 
  \includegraphics[width=.16\linewidth]{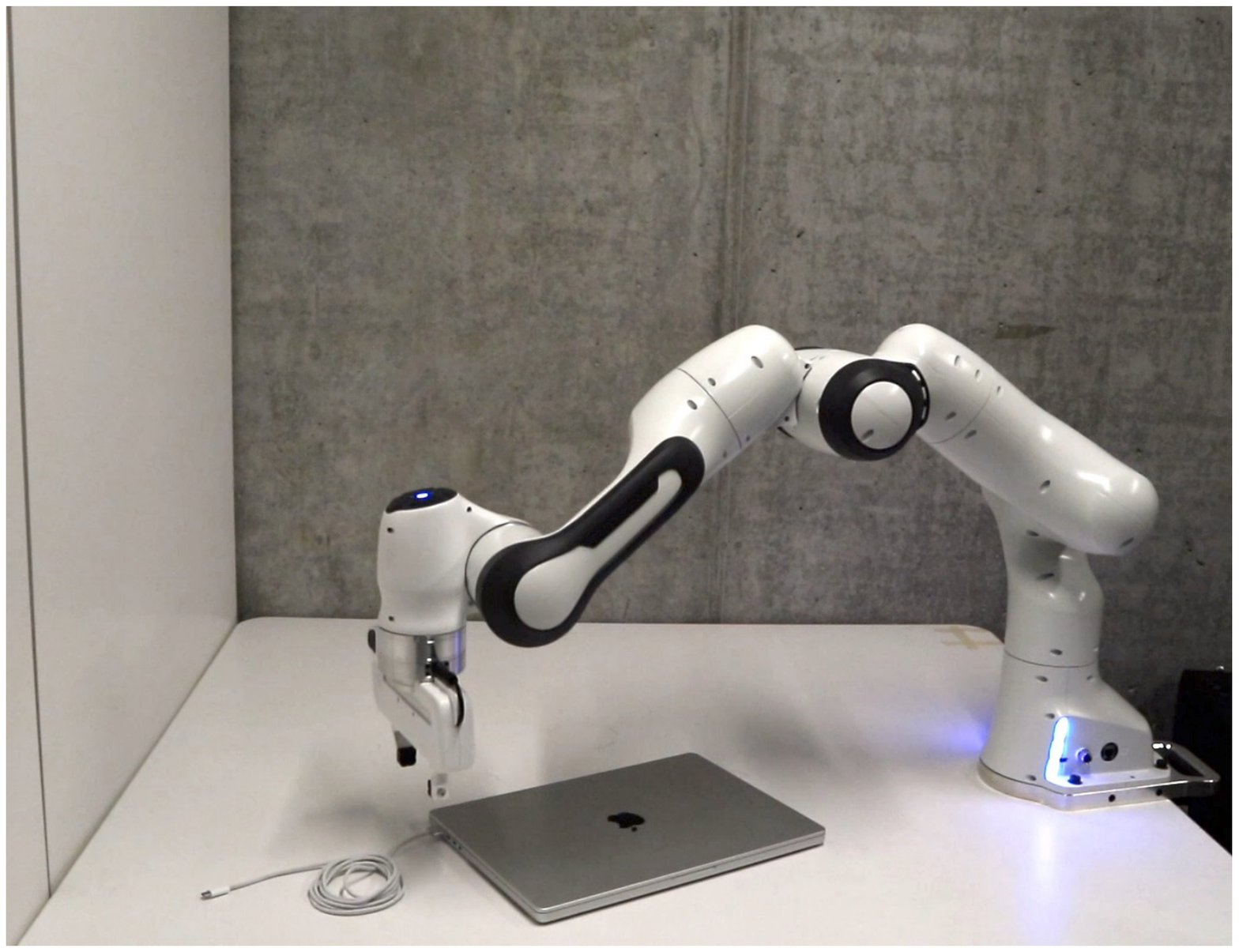} & 
  \includegraphics[width=.16\linewidth]{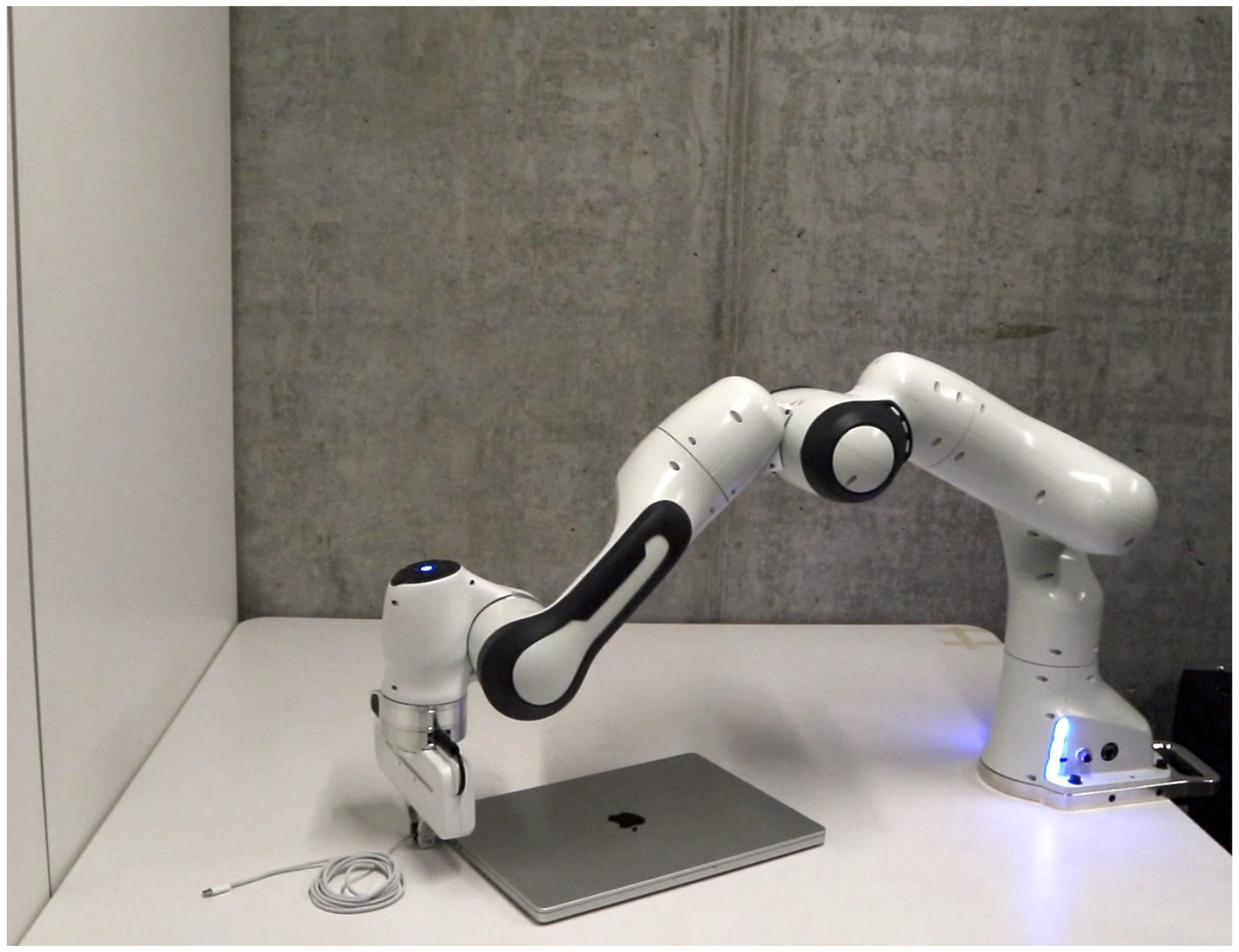} & 
  \includegraphics[width=.16\linewidth]{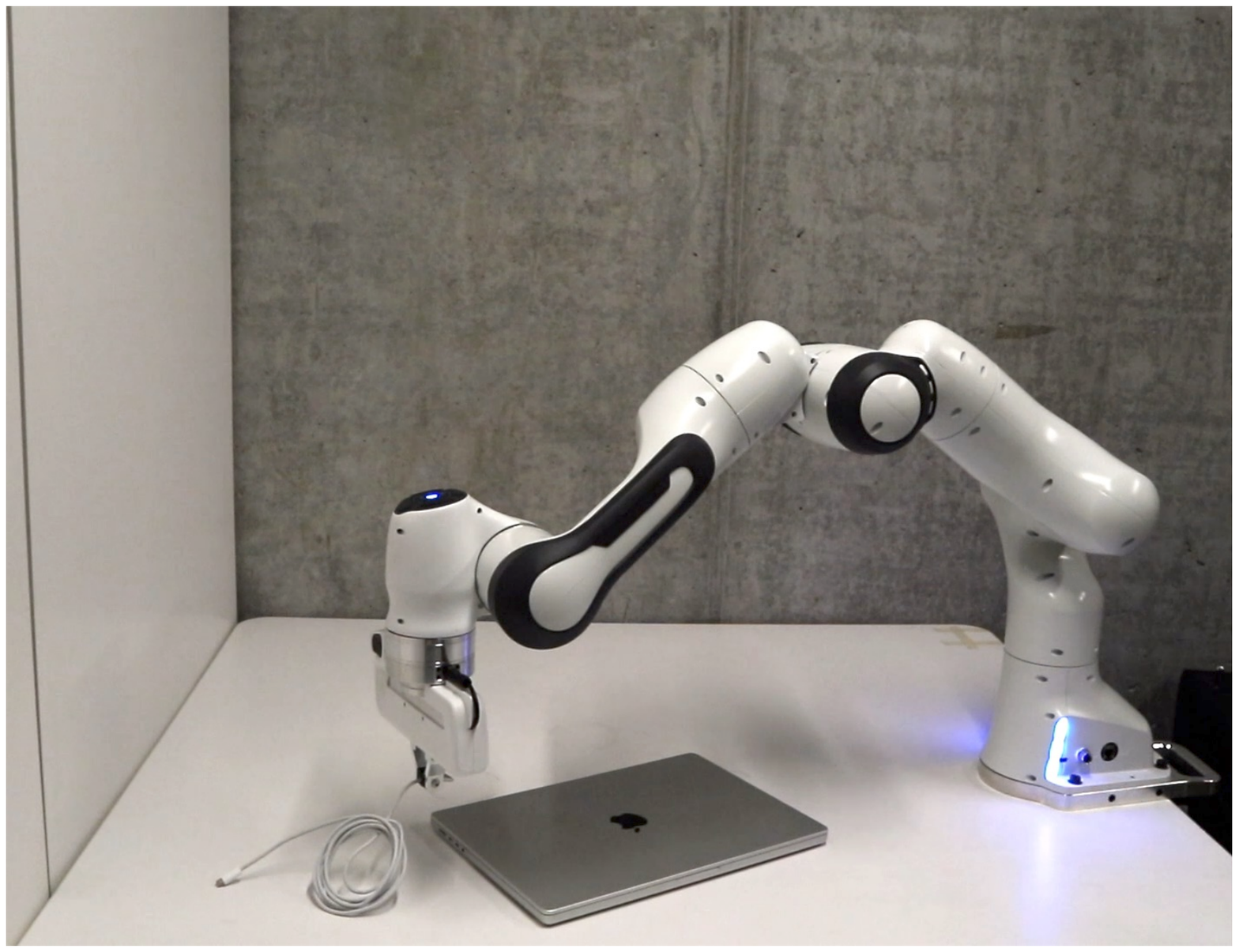} & 
  \includegraphics[width=.16\linewidth]{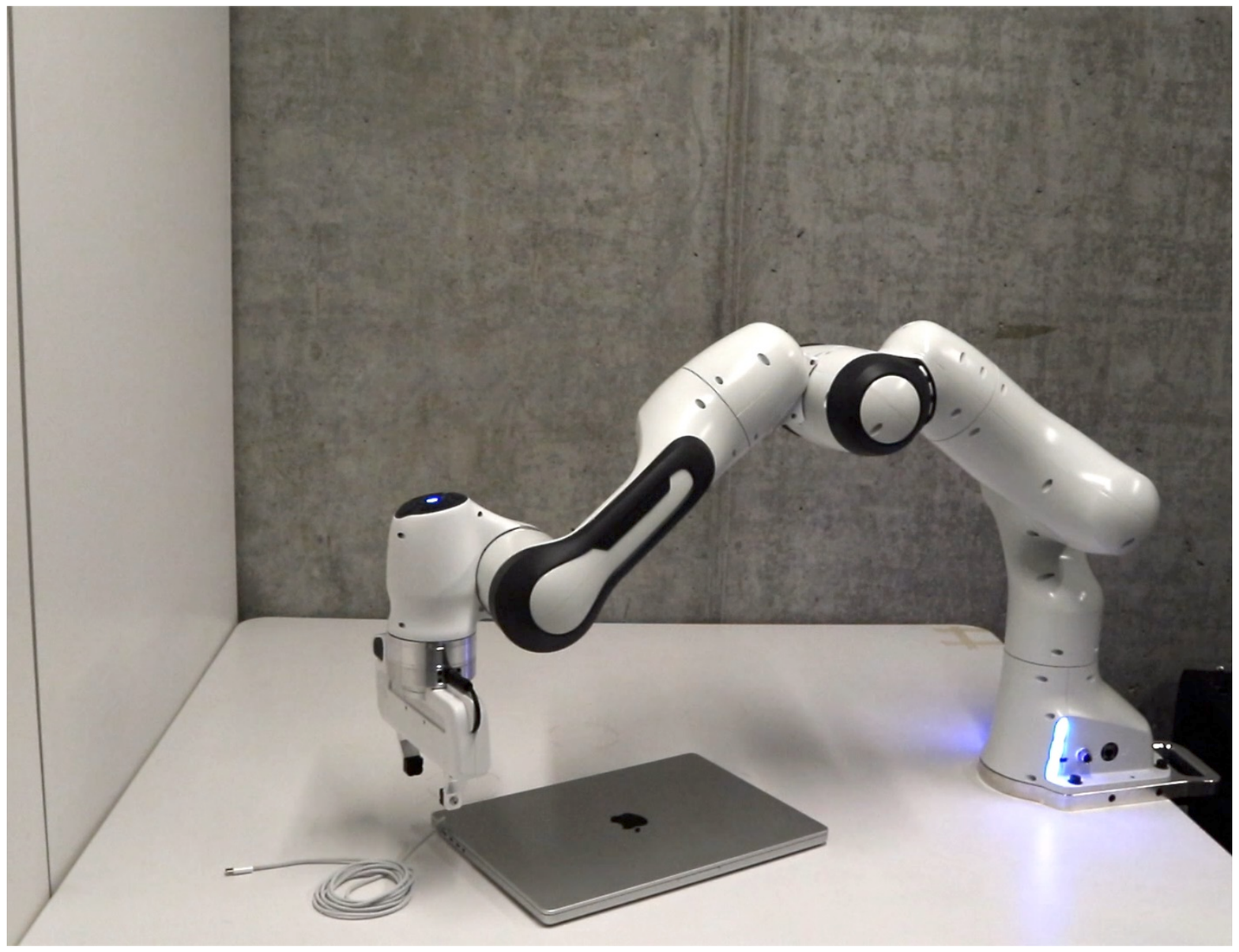} \\

  \multicolumn{6}{c}{(d) unplug laptop} \\

  \includegraphics[width=.16\linewidth]{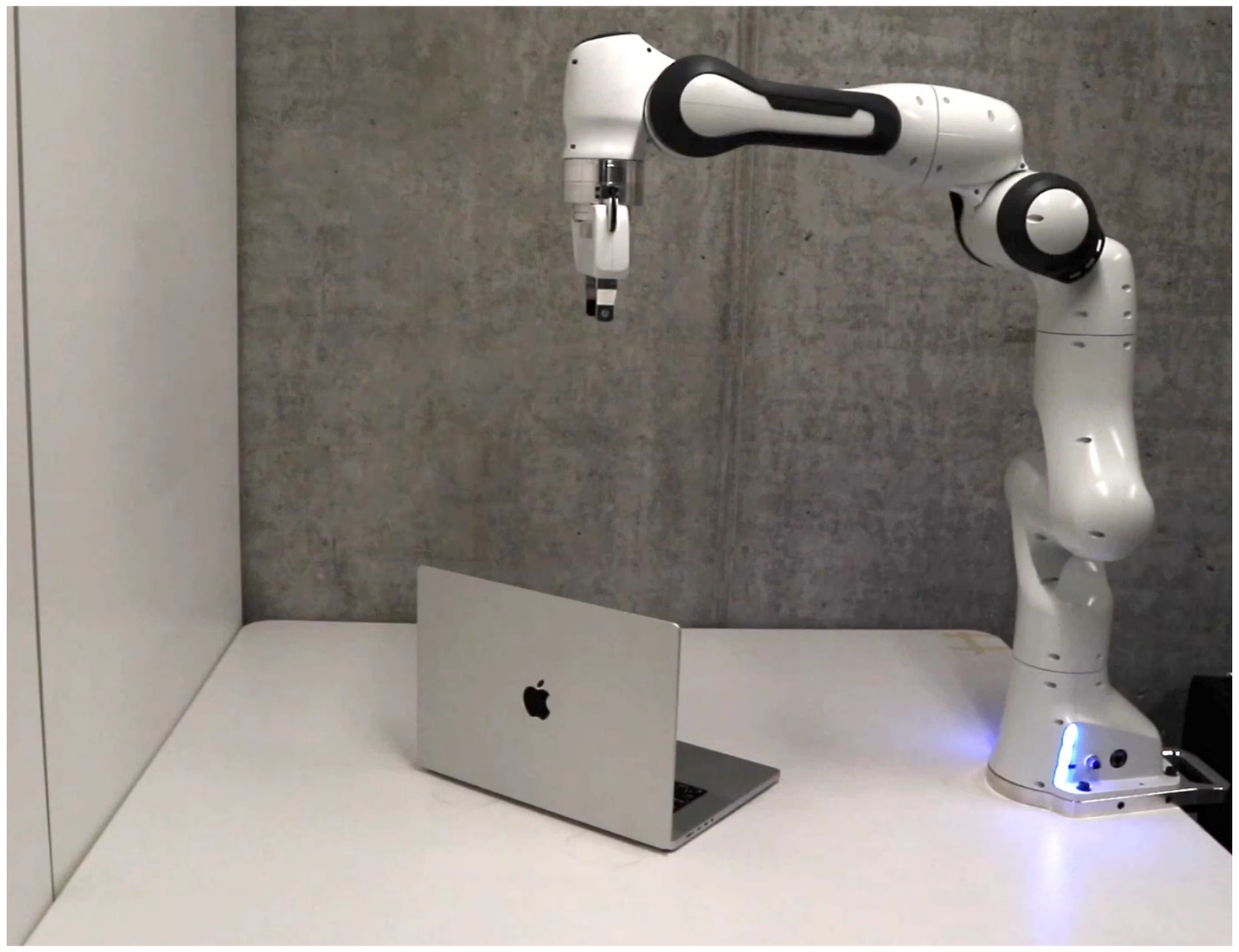} & 
  \includegraphics[width=.16\linewidth]{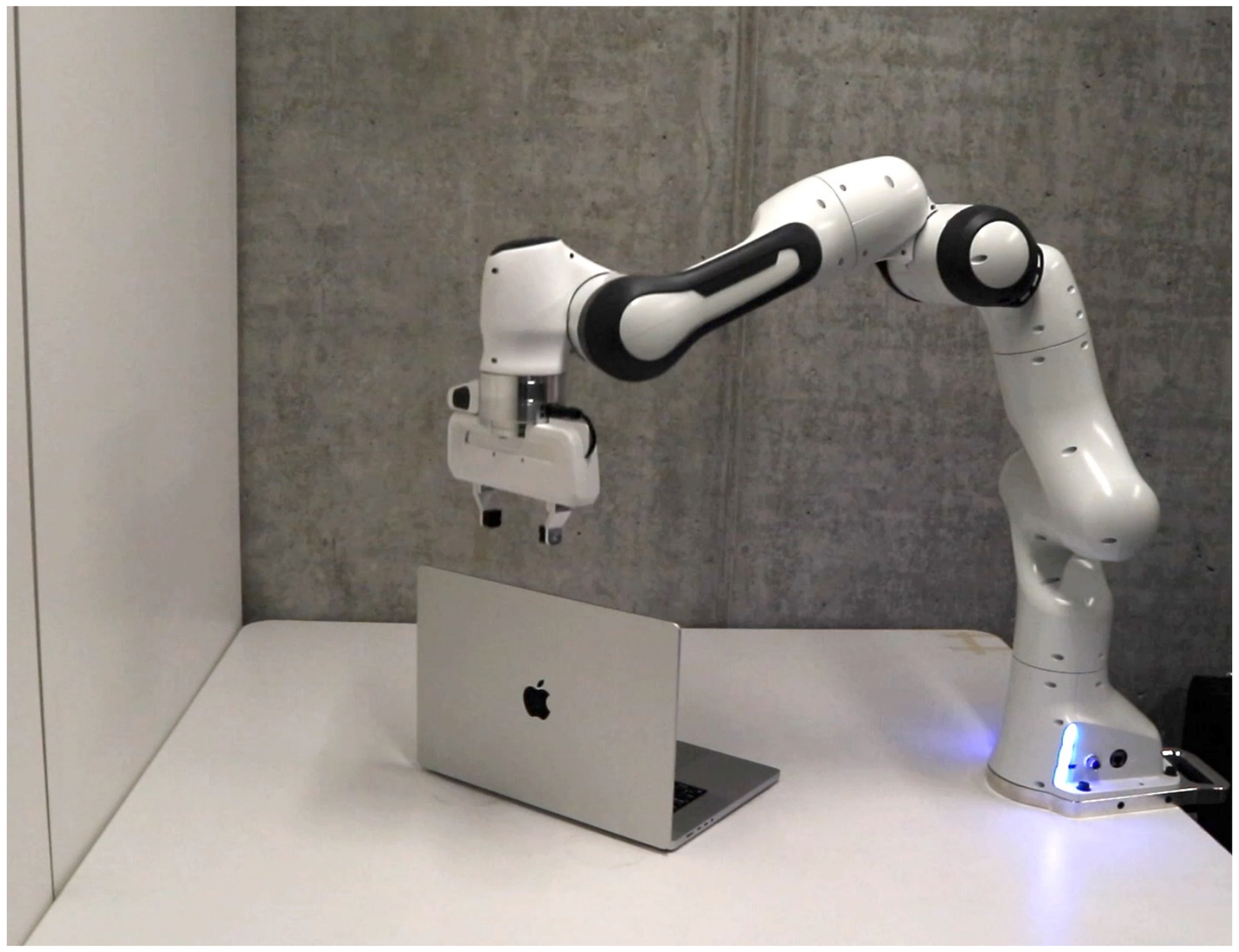} & 
  \includegraphics[width=.16\linewidth]{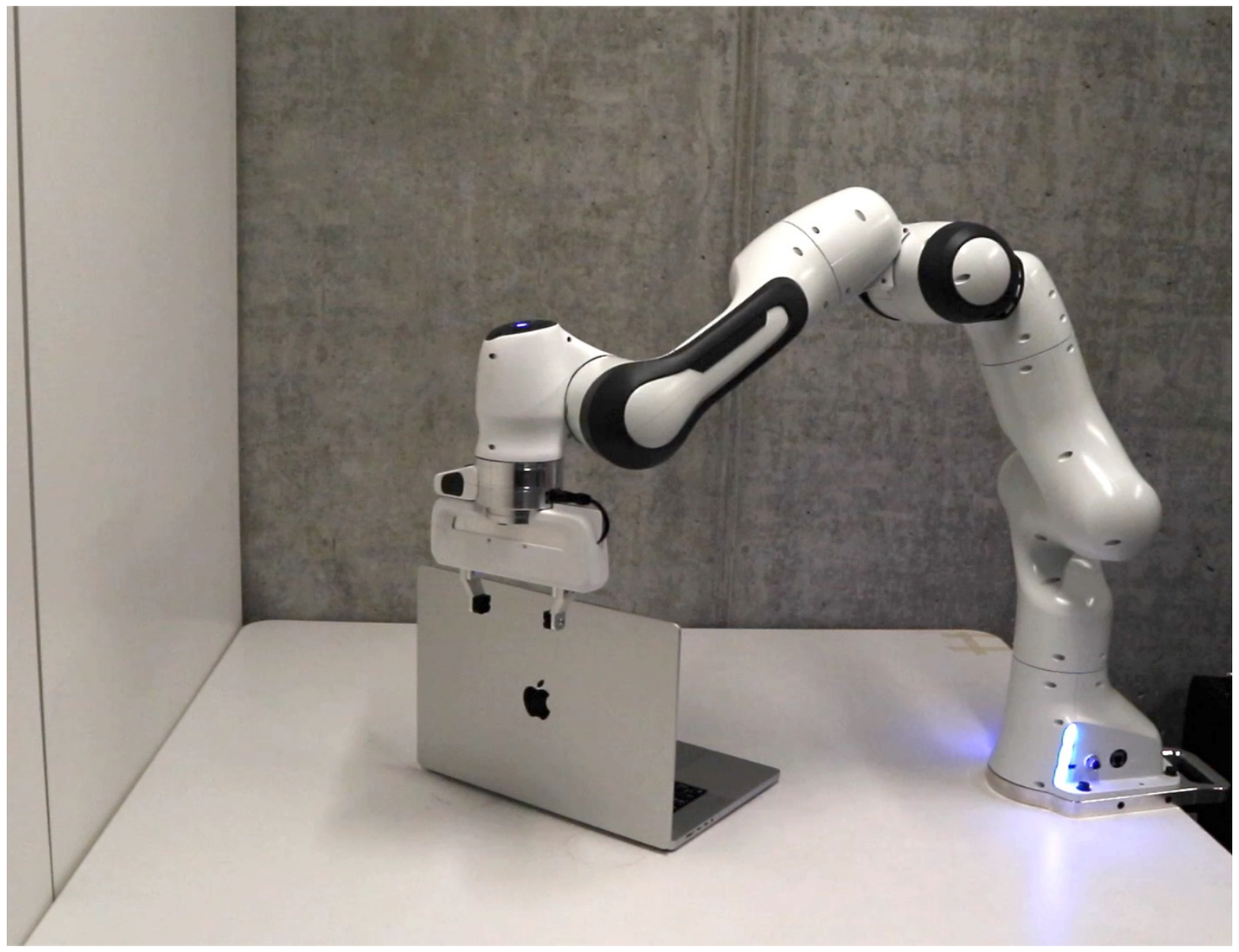} & 
  \includegraphics[width=.16\linewidth]{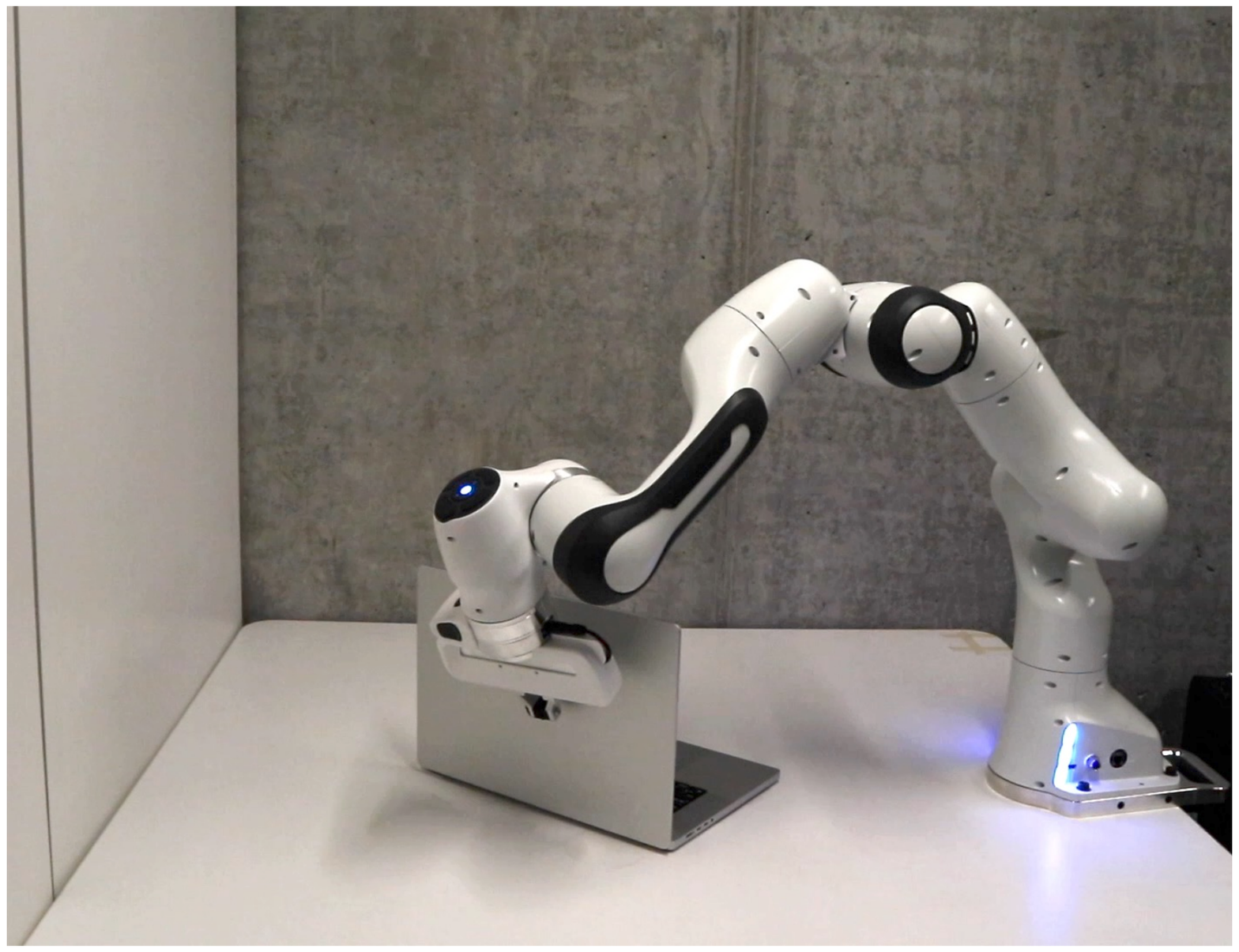} & 
  \includegraphics[width=.16\linewidth]{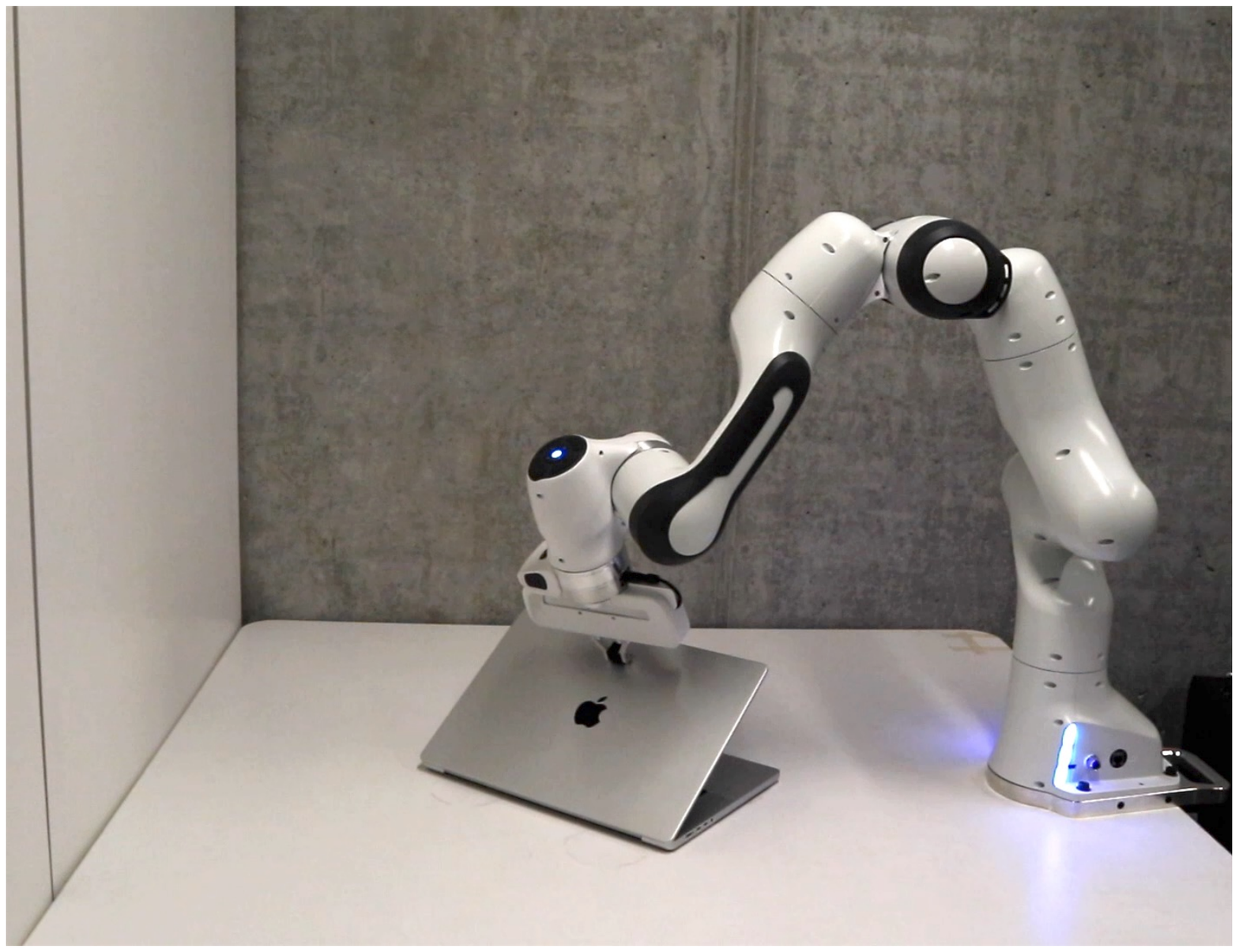} & 
  \includegraphics[width=.16\linewidth]{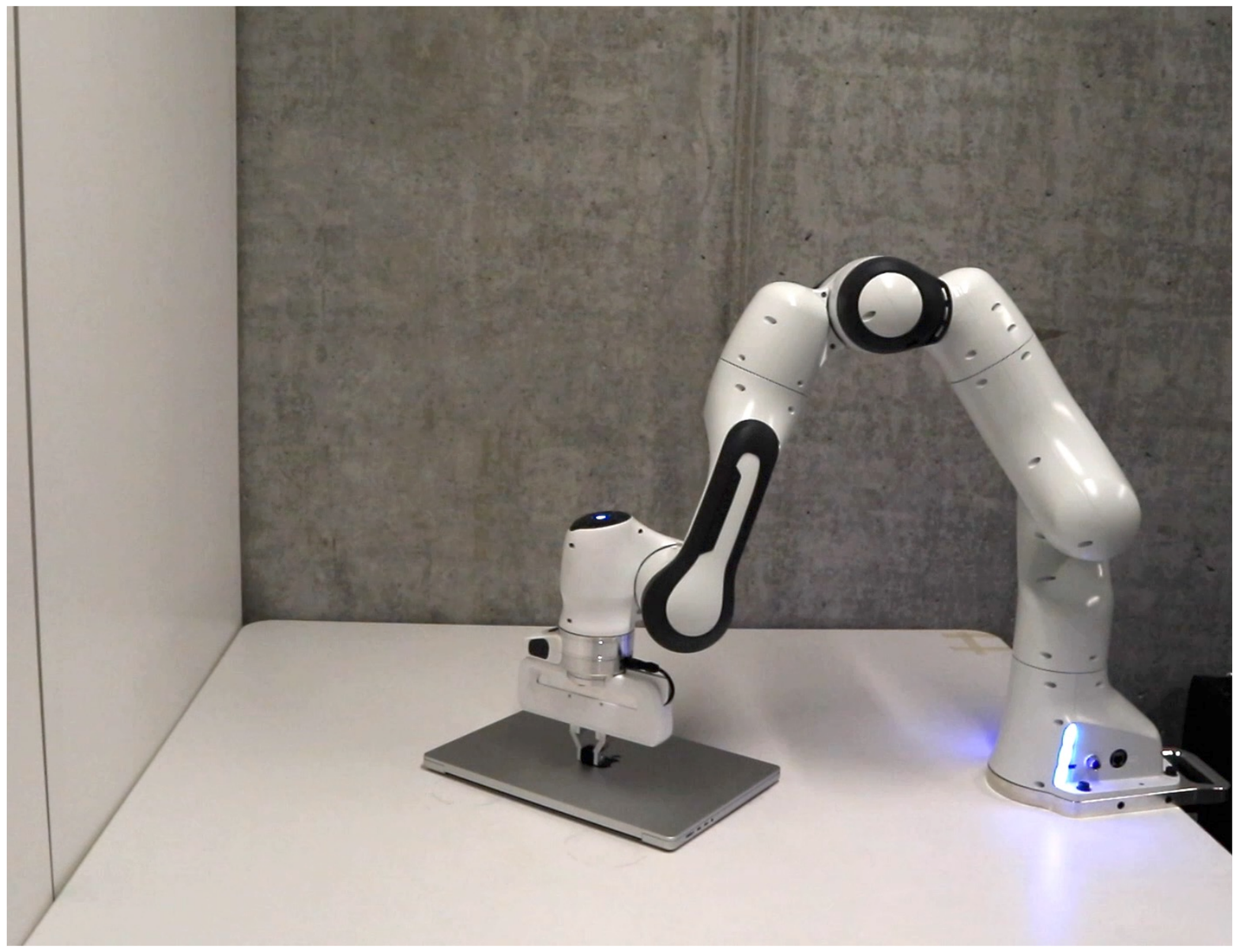} \\

  \multicolumn{6}{c}{(e) close laptop} \\
  
  \includegraphics[width=.16\linewidth]{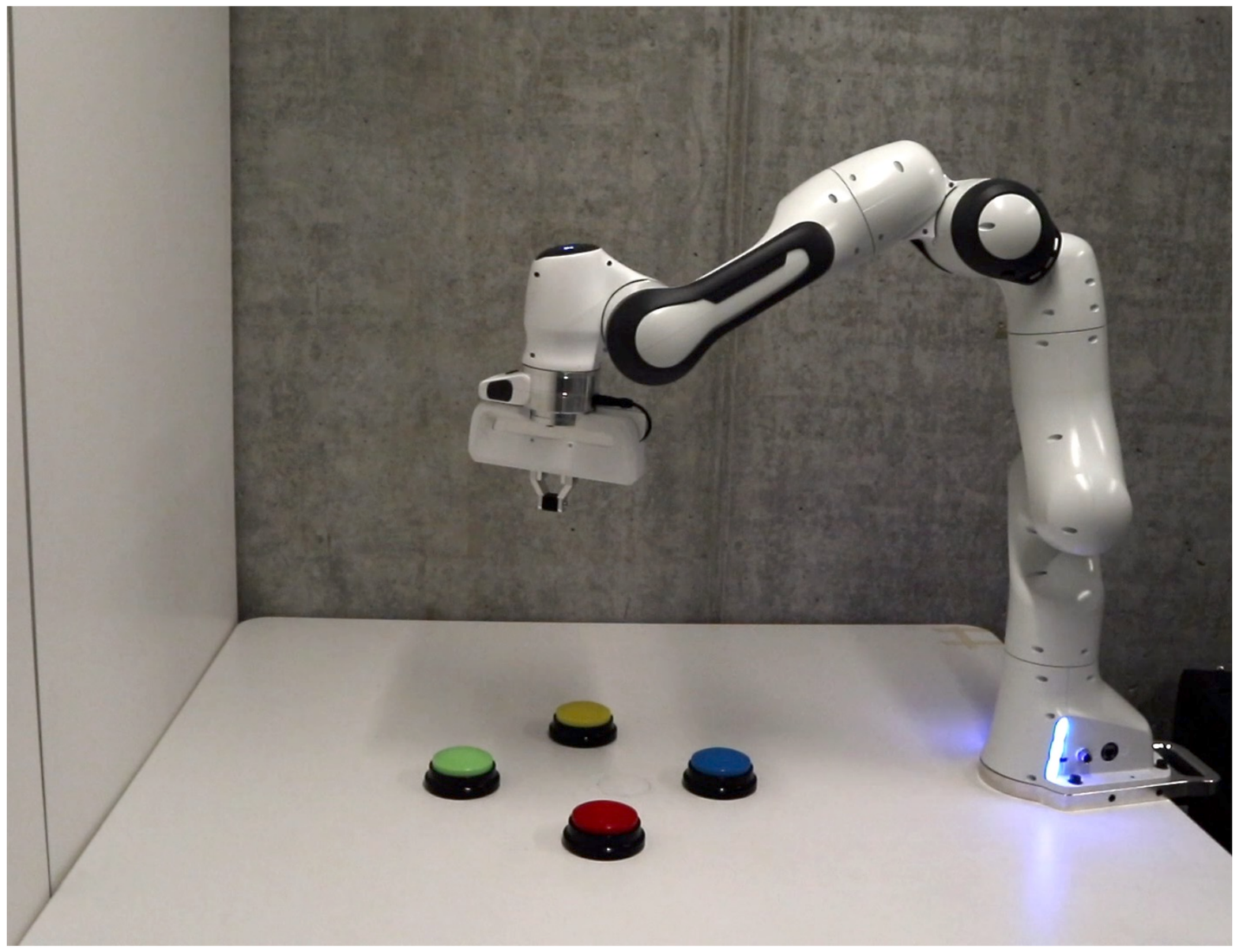} & 
  \includegraphics[width=.16\linewidth]{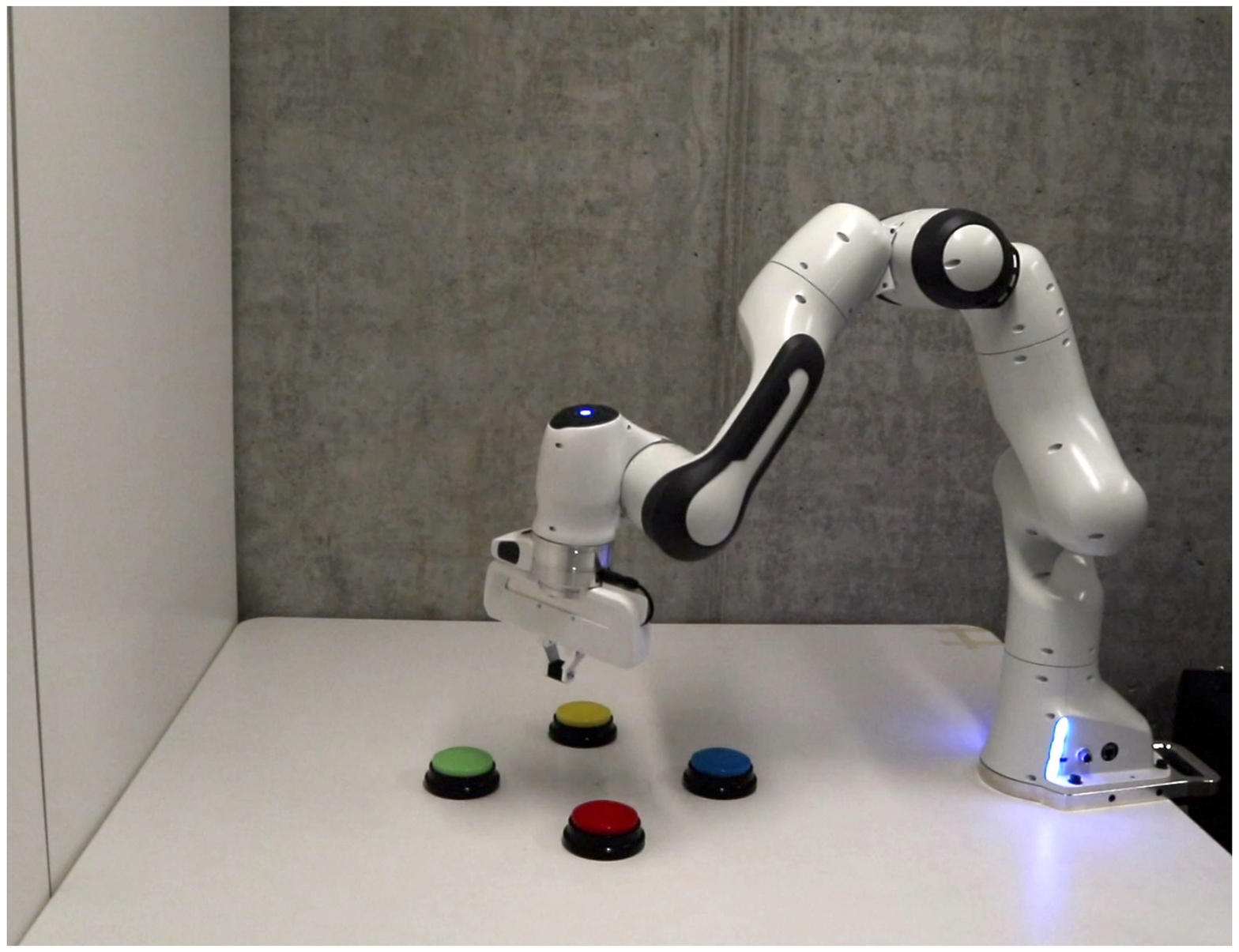} & 
  \includegraphics[width=.16\linewidth]{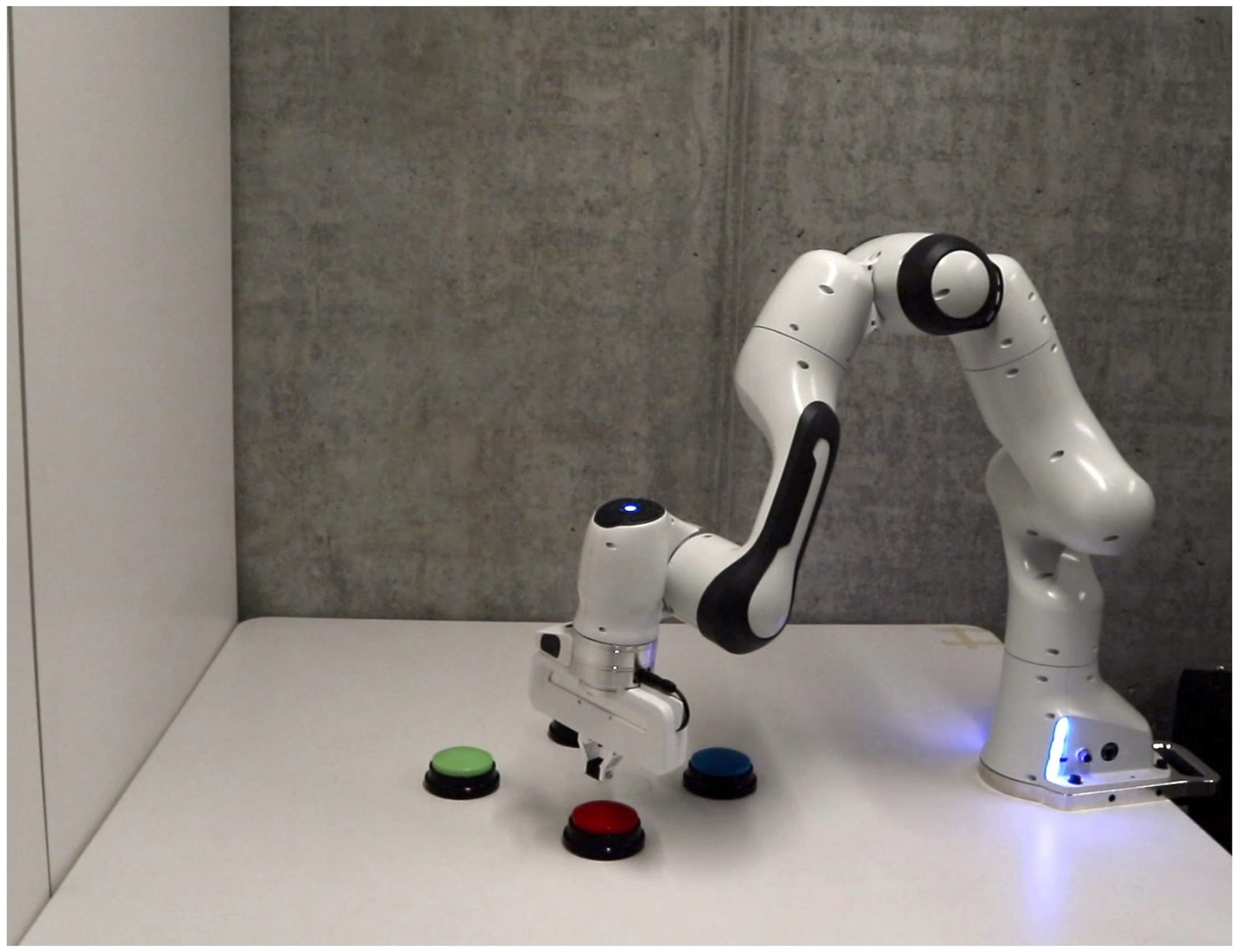} & 
  \includegraphics[width=.16\linewidth]{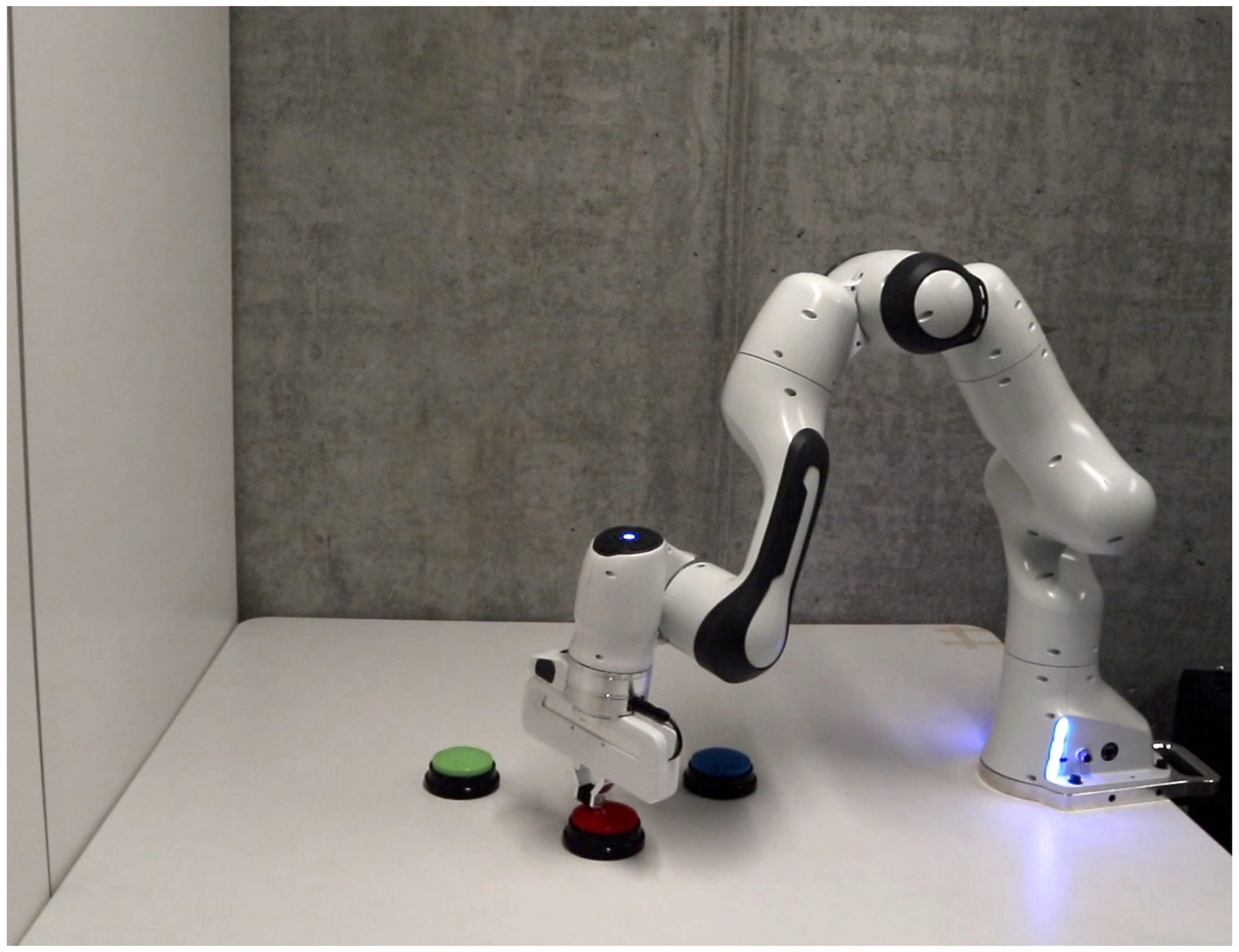} & 
  \includegraphics[width=.16\linewidth]{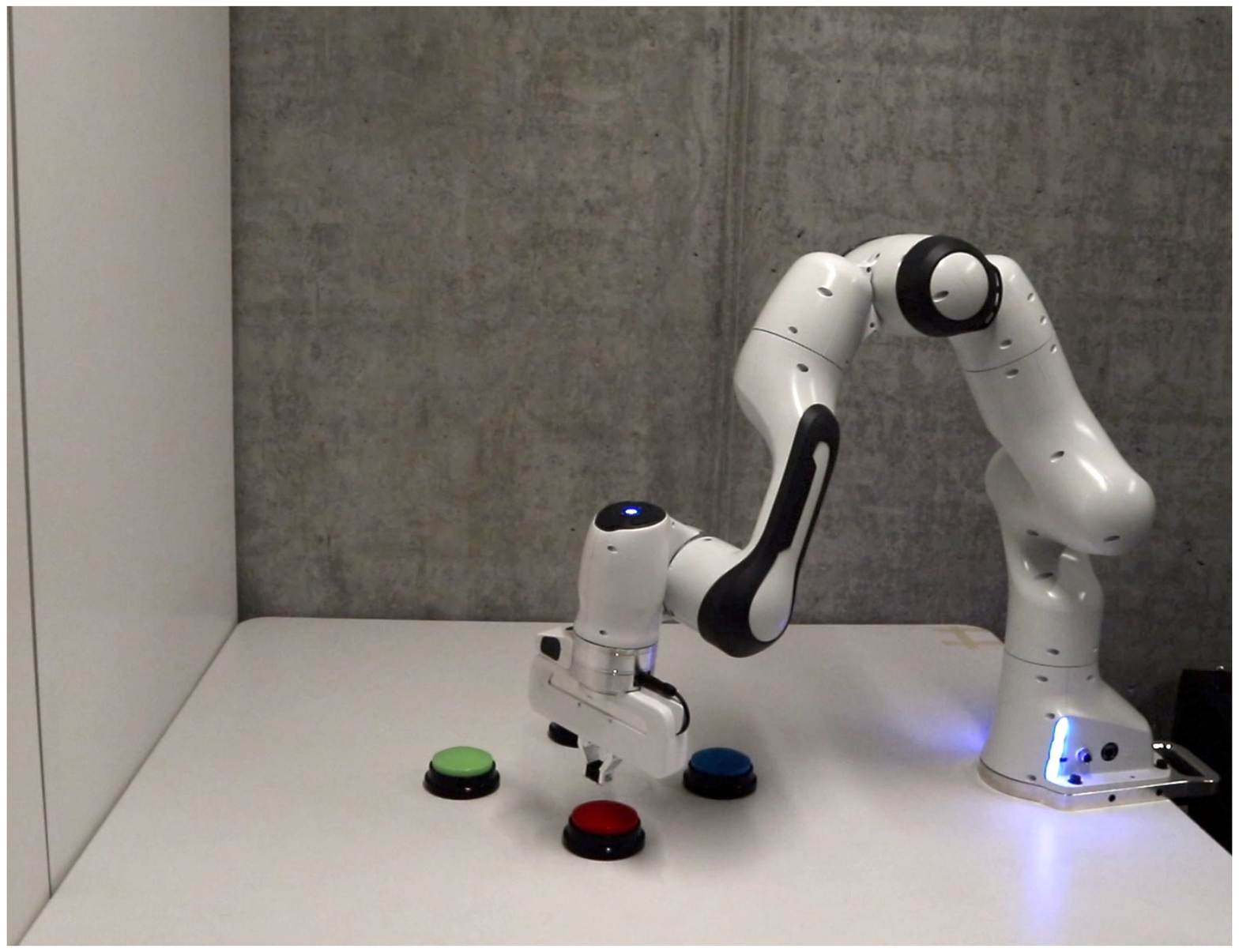} & 
  \includegraphics[width=.16\linewidth]{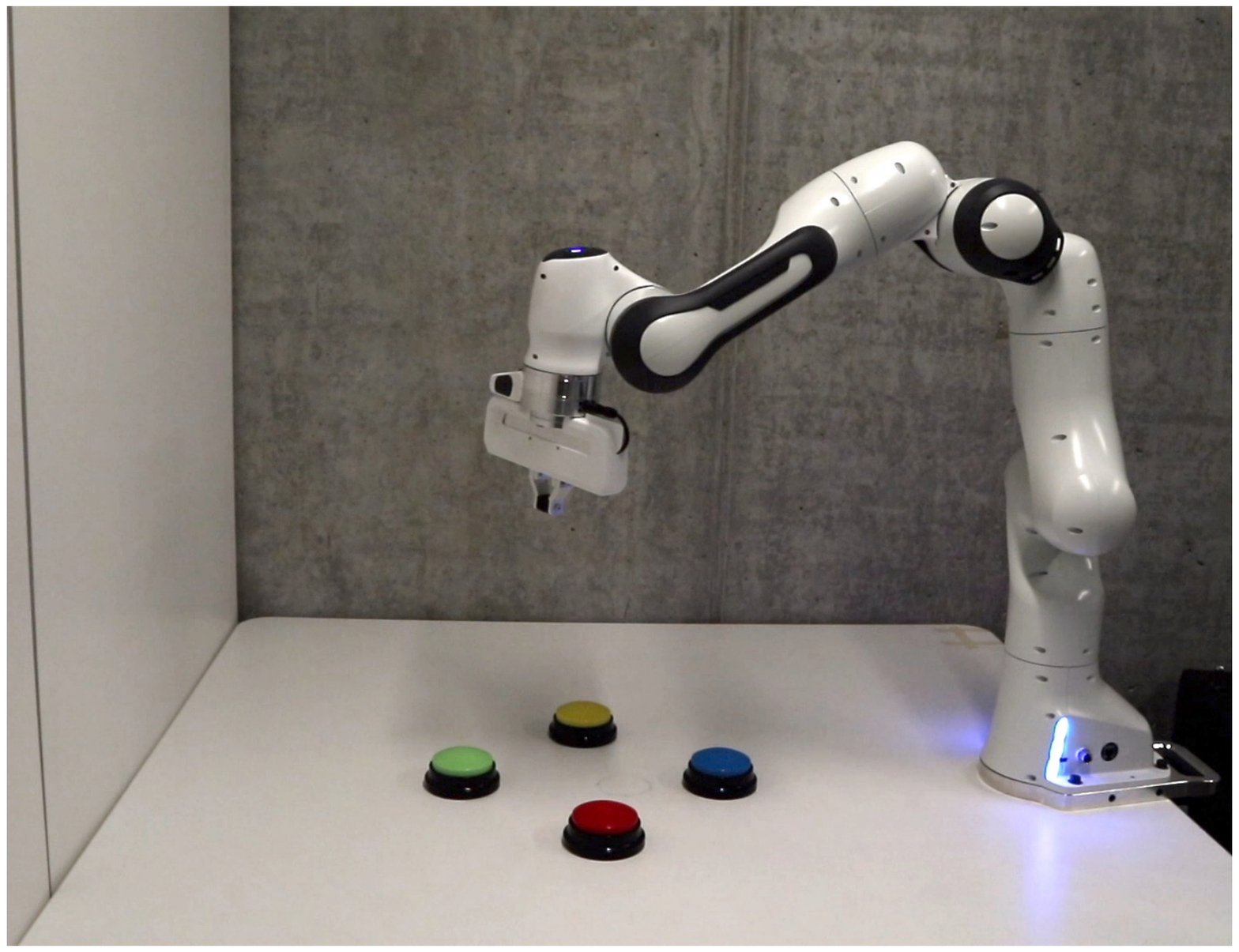} \\

  \multicolumn{6}{c}{(f) push multiple buttons} \\
    
  \end{tabular}
  \captionof{figure}{\textbf{Qualitative visualization of {\smodel} on 6 real-world tasks.}}
  \label{fig:rollout}
  \vspace{-0.1cm}
\end{figure*}

\minisection{Push multiple buttons} The task is to push multiple buttons in a specified order. The success criteria is each button being pressed correctly in a specified order. The names of objects we provide are ``red button'', ``yellow button'', ``green button'', and ``blue button''. To increase the difficult of this task, we only provide example episodes to \emph{push a single button} (\eg, the task instruction is ``push the red/yellow/green/blue button''). During the evaluation, we evaluate our method to \emph{press a sequence of buttons} (\eg, the task instruction is ``push the red/yellow/green/blue button, then push the red/yellow/green/blue button, ...''). The total number of buttons to be pressed during the evaluation is uniformly sampled from $\{1,\cdots,6\}$ and the pressing order is randomized.

\section{Qualitative Visualizations}
\label{supp:qual}
In Figure~\ref{fig:rollout}, we present qualitative visualizations for the 6 real-world tasks that {\smodel} is evaluated on.

\end{document}